\documentclass{article}

\usepackage[final,nonatbib]{neurips_2020}

\usepackage[utf8]{inputenc} 
\usepackage[T1]{fontenc}    
\usepackage{amsmath}
\usepackage{hyperref} 
\usepackage{cleveref}
\crefformat{footnote}{#2\footnotemark[#1]#3}

\usepackage{url}            
\usepackage{booktabs}       
\usepackage{amsfonts}       
\usepackage{nicefrac}       
\usepackage{microtype}      
\usepackage{xcolor}

\usepackage{bm}
\usepackage{amsthm}
\usepackage{amssymb}
\usepackage{amsfonts}
\usepackage{graphicx}
\usepackage{makecell}
\usepackage{multirow}

\theoremstyle{definition}

\newtheorem{proposition}{Proposition}

\newtheorem{remark}{Remark}
\newtheorem{example}{Example}

\newcommand{\todo}[1]{}

\newcommand{\mbf}[1]{\mathbf{#1}}
\newcommand{\mcl}[1]{\mathcal{#1}}
\newcommand{\mbb}[1]{\mathbb{#1}}

\newcommand{\da}{\mcl{D}}
\newcommand{\dc}{\mcl{D}_r}
\newcommand{\dr}{\mcl{D}_e}
\newcommand{\eubo}{\tilde{q}_u}
\newcommand{\elbo}{\tilde{q}_v}

\title{Variational Bayesian Unlearning}

\author{%
  Quoc Phong Nguyen, Bryan Kian Hsiang Low, and Patrick Jaillet\textsuperscript{\dag}\\
  Dept. of Computer Science, National University of Singapore, Republic of Singapore\\
  Dept. of Electrical Engineering and Computer Science, MIT, USA\textsuperscript{\dag}\\
  \texttt{\{qphong,lowkh\}@comp.nus.edu.sg}, \texttt{jaillet@mit.edu}\textsuperscript{\dag}
}

\begin{document}

\maketitle

\begin{abstract}
This paper studies the problem of approximately unlearning a Bayesian model from a small subset of the training data to be erased. We frame this problem as one of minimizing the Kullback-Leibler divergence between the approximate posterior belief of model parameters after directly unlearning from erased data vs.~the exact posterior belief from retraining with remaining data. Using the \emph{variational inference} (VI) framework, we show that it is equivalent to minimizing an evidence upper bound which trades off between fully unlearning from erased data vs.~not entirely forgetting the posterior belief given the full data (i.e., including the remaining data); the latter prevents catastrophic unlearning that can render the model useless. In model training with VI, only an approximate (instead of exact) posterior belief given the full data can be obtained, which makes unlearning even more challenging. We propose two novel tricks to tackle this challenge. We empirically demonstrate our unlearning methods on Bayesian models such as sparse Gaussian process and logistic regression using synthetic and real-world datasets.
\end{abstract}

\todo{measure -> criterion ->  the existence of the opposite -> leads to learn vs unlearn (ELBO vs EUBO)}

\section{Introduction}
\label{sec:intro}
Our interactions with \emph{machine learning} (ML) applications have surged in recent years 
such that large quantities of users' data are now deeply ingrained into the ML models being trained for these applications.
This greatly complicates the regulation of access to each user's data or implementation of  \emph{personal data ownership}, which are enforced by the General Data Protection Regulation in the European Union~\cite{mantelero2013eu}.
In particular, 
if a user would like to exercise her \emph{right to be forgotten}~\cite{mantelero2013eu} (e.g., when quitting an ML application),
then it would be desirable to have the trained ML model  ``unlearn'' from her data.
Such a problem of \emph{machine unlearning}~\cite{cao2015towards} extends to the practical scenario where a small subset of data previously used for training is later identified as malicious (e.g., anomalies)~\cite{cao2015towards,du2019lifelong} and the trained ML model can perform well once again if it can unlearn from the malicious data.

A naive alternative to machine unlearning is to simply retrain an ML model from scratch with the data \emph{remaining} after \emph{erasing} that to be unlearned from.
In practice, this is prohibitively expensive in terms of time and space costs since the remaining data is often large such as in the above scenarios.
How then can a trained ML model  \emph{directly} and efficiently unlearn from a small subset of data to be erased to become (a) exactly and if not, (b) approximately close to that from retraining with the large remaining data?
Unfortunately, (a) exact unlearning is only possible for selected ML models (e.g., naive Bayes classifier, linear regression, $k$-means clustering, and item-item collaborative filtering \cite{cao2015towards,ginart2019making,schelteramnesia}).
This motivates the need to consider (b) approximate unlearning as it is applicable to a broader family of ML models like neural networks  \cite{du2019lifelong,golatkar2019eternal} but, depending on its choice of loss function, may suffer from \emph{catastrophic unlearning}\footnote{\label{beef}A trained ML model is said to experience \emph{catastrophic unlearning} from the erased data when its resulting performance is considerably worse than that from retraining with the remaining data.} that can render the model useless.
For example, to mitigate this issue, 
%
%
the works of~\cite{du2019lifelong,golatkar2019eternal} have to ``patch up''
their loss functions by additionally bounding the loss incurred by erased data with a rectified linear unit and injecting a regularization term to retain information of the remaining data, respectively.
This begs the question whether there exists a loss function that can \emph{directly} quantify the approximation gap
and \emph{naturally} prevent catastrophic unlearning.

\nocite{bourtoule2019machine}

Our work here addresses the above question by focusing on the family of Bayesian models.
Specifically, our proposed loss function measures the \emph{Kullback-Leibler} (KL) divergence between the approximate posterior belief of model parameters by directly unlearning from erased data vs.~the exact posterior belief from retraining with remaining data. Using the \emph{variational inference} (VI) framework, we show that minimizing this KL divergence is equivalent to \emph{minimizing} (instead of maximizing) a counterpart of the evidence lower bound called the \emph{evidence upper bound} (EUBO) (Sec.~\ref{subsec:exactfull}). Interestingly, the EUBO lends itself to a natural interpretation of a trade-off between fully unlearning from erased data vs.~not entirely forgetting the posterior belief given the \emph{full} data (i.e., including the remaining data); the latter prevents catastrophic unlearning induced by the former.



Often, in model training, only an approximate (instead of exact) posterior belief of model parameters given the full data can be learned, say, also using VI. This makes unlearning even more challenging.
To tackle this challenge, we analyse two sources of inaccuracy in the approximate posterior belief learned using VI, which lay the groundwork for proposing our first trick of an \emph{adjusted likelihood} of erased data (Sec.~\ref{subsubsec:eubo}): 
Our key idea is to curb
unlearning in the region of model parameters with low  approximate posterior belief where both sources of inaccuracy primarily occur.
Additionally, to avoid the risk of incorrectly tuning the adjusted likelihood, we propose another trick of \emph{reverse KL}  (Sec.~\ref{subsubsec:rkl}) which is naturally more protected from such inaccuracy without needing the adjusted likelihood. Nonetheless, our adjusted likelihood is general enough to be applied to reverse KL. 


VI is a popular approximate Bayesian inference framework due to its scalability to massive datasets~\cite{hensman2013gaussian,hoang2015unifying} and its ability to model complex posterior beliefs using generative adversarial networks \cite{yu19} and normalizing flows \cite{kingma2016improved,rezende2015variational}. 
Our work in this paper exploits VI to broaden the family of ML models that can be unlearned, which we empirically demonstrate using synthetic and real-world datasets on several Bayesian models such as sparse Gaussian process and logistic regression with the approximate posterior belief modeled by a normalizing flow (Sec.~\ref{sec:experiment}).
\section{Variational Inference (VI)}
\label{sec:vi}
In this section, we revisit the VI framework \cite{blei2017variational} for learning an approximate posterior belief of the parameters $\bm{\theta}$ of a Bayesian model. Given a prior belief $p(\bm{\theta})$ of the unknown model parameters $\bm{\theta}$ and a set $\da$ of training data, an approximate posterior belief $q(\bm{\theta}|\da) \approx p(\bm{\theta}|\da)$ is being optimized by minimizing the KL divergence $\text{KL}[q(\bm{\theta}|\da)\ \Vert\  p(\bm{\theta}|\da)] \triangleq \int q(\bm{\theta}|\da)\ \log (q(\bm{\theta}|\da) / p(\bm{\theta}|\da))\ \text{d}\bm{\theta}$ or, equivalently, maximizing the \emph{evidence lower bound} (ELBO) $\mcl{L}$ \cite{blei2017variational}:
\begin{equation}
\mcl{L} \triangleq \int q(\bm{\theta}|\da)\ \log p(\da | \bm{\theta})\ \text{d}\bm{\theta} - \text{KL}[q(\bm{\theta}|\da)\ \Vert\ p(\bm{\theta})]\ .
\label{eq:elbo}
\end{equation}
%
Such an equivalence follows directly from $\mcl{L}=\log p(\da) - \text{KL}[q(\bm{\theta}|\da)\ \Vert\  p(\bm{\theta}|\da)]$ where the log-marginal likelihood $\log p(\da)$ is independent of $q(\bm{\theta}|\da)$. Since $\text{KL}[q(\bm{\theta}|\da)\ \Vert\ p(\bm{\theta}|\da)] \ge 0$, the ELBO $\mcl{L}$ is a lower bound of $\log p(\da)$.
The ELBO $\mcl{L}$ in~\eqref{eq:elbo} can be interpreted as a trade-off between attaining a higher likelihood of $\da$ (first term) vs.~not entirely forgetting the prior belief $p(\bm{\theta})$ (second term).

When the ELBO $\mcl{L}$~\eqref{eq:elbo} cannot be evaluated in closed form, it can be maximized using \emph{stochastic gradient ascent} (SGA) by approximating the expectation in 
\[
\mcl{L} = \mbb{E}_{q(\bm{\theta}|\da)} [\log p(\da|\bm{\theta}) + \log (p(\bm{\theta}) / q(\bm{\theta}|\da))] = \int q(\bm{\theta}|\da)\ (\log p(\da|\bm{\theta}) + \log (p(\bm{\theta}) / q(\bm{\theta}|\da)))\ \text{d}\bm{\theta}
\]
with stochastic sampling in each iteration of SGA. The approximate posterior belief $q(\bm{\theta}|\da)$ can be represented by a simple distribution (e.g., in the exponential family) for computational ease or a complex distribution (e.g., using generative neural networks) for expressive power.
Note that when the distribution of $q(\bm{\theta}|\da)$ is modeled by a generative neural network whose density cannot be evaluated, the ELBO can be maximized with adversarial training by alternating between estimating the log-density ratio $\log (p(\bm{\theta}) / q(\bm{\theta}|\da))$ and maximizing the ELBO \cite{yu19}.
On the other hand, when the distribution of $q(\bm{\theta}|\da)$ is modeled by a normalizing flow (e.g., \emph{inverse autoregressive flow} (IAF) \cite{kingma2016improved}) whose density can be computed, the ELBO can be maximized with SGA. 
%
\todo{mentioning adversarial method with generative model without known density}
\todo{mentioning that using IAF to achieve better approximation around the region of high posterior probability}
\section{Bayesian Unlearning}
\label{sec:bayesunlearn}
\subsection{Exact Bayesian Unlearning}
\label{subsec:exactbayesunlearn}
Let the (\emph{full}) training data $\da$ be partitioned into a small subset $\dr$ of data to be \emph{erased} and a (large) set $\dc$ of \emph{remaining} data, i.e., $\da = \dc\ \cup\ \dr$ and $\dc\ \cap\ \dr = \emptyset$. 
The problem of \emph{exact Bayesian unlearning} involves
recovering the exact posterior belief $p(\bm{\theta}|\dc)$ of model parameters $\bm{\theta}$ given remaining data $\dc$ from that given full data $\da$ (i.e., $p(\bm{\theta}|\da)$ assumed to be available)
by directly unlearning from erased data $\dr$.
Note that $p(\bm{\theta}|\dc)$ can also be obtained from retraining with remaining data $\dc$, which is computationally costly, as discussed in Sec.~\ref{sec:intro}.
By using Bayes' rule and assuming conditional independence between $\dc$ and $\dr$ given $\bm{\theta}$, 
%
\begin{equation}
p(\bm{\theta}|\dc)\ \ =\ \  {p(\bm{\theta}|\da) \ p(\dr|\dc)}/{p(\dr| \bm{\theta})}
	\ \ \propto\ \  {p(\bm{\theta}|\da)}/{p(\dr| \bm{\theta})}\ .
\label{eq:suffdata}
\end{equation}
%
%
%
When the model parameters $\bm{\theta}$ are discrete-valued, $p(\bm{\theta}|\dc)$ can be  obtained from \eqref{eq:suffdata} directly. 
The use of a conjugate prior also makes unlearning relatively simple.
We will investigate the more interesting case of a non-conjugate prior in the rest of Sec.~\ref{sec:bayesunlearn}.
%
%
\subsection{Approximate Bayesian Unlearning with Exact Posterior Belief
$p($\texorpdfstring{$\bm{\theta}$}{theta}$|\da)$}
\label{subsec:exactfull}
\begin{figure}
\begin{tabular}{c|c}
     \includegraphics[height=0.17\textwidth]{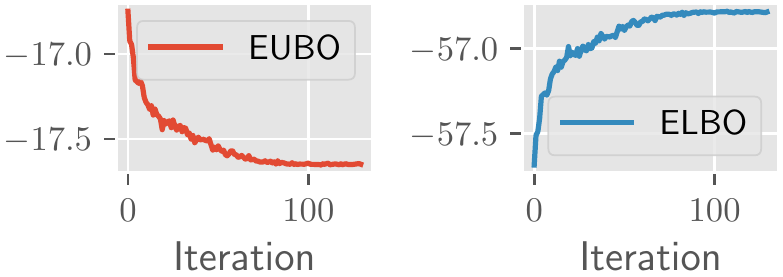}
     &
     \includegraphics[height=0.17\textwidth]{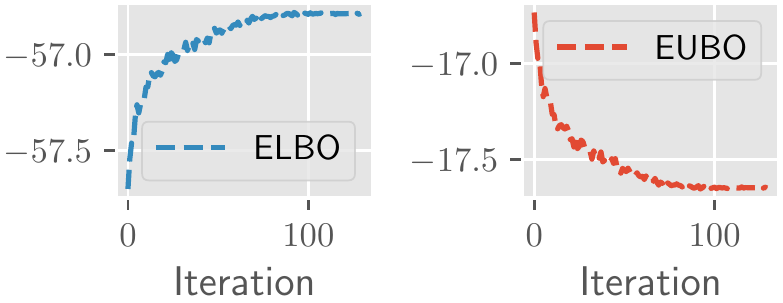}\\
     (a) Unlearning from $\dr$ by minimizing EUBO
     & 
     (b) Retraining with $\dc$ by maximizing ELBO
\end{tabular}
\caption{Plots of EUBO and ELBO when (a) unlearning from $\dr$ and (b) retraining with $\dc$.}
\label{fig:eubovselbo}
\end{figure}

The problem of \emph{approximate Bayesian unlearning} differs from that of exact Bayesian unlearning (Sec.~\ref{subsec:exactbayesunlearn}) in that only the approximate posterior belief $q_u(\bm{\theta}|\dc)$ (instead of the exact one $p(\bm{\theta}|\dc)$) can be recovered by directly unlearning from erased data $\dr$.
Since existing unlearning methods often use their model predictions to construct their loss functions \cite{bourtoule2019machine,cao2015towards,ginart2019making,guo2019certified}, we have initially considered doing likewise (albeit in the Bayesian context) by defining the loss function
as the KL divergence between the approximate predictive distribution $q_u(y|\dc) \triangleq 
\int p(y|\bm{\theta})\ q_u(\bm{\theta}|\dc)\  \text{d}\bm{\theta}$ vs.~the exact predictive distribution
$p(y|\dc) = 
\int p(y|\bm{\theta})\ p(\bm{\theta}|\dc)\  \text{d}\bm{\theta}$ where the observation $y$ (i.e., drawn from a model with parameters $\bm{\theta}$) is conditionally independent of $\dc$ given $\bm{\theta}$.
%
%
However, it may not be possible to evaluate these predictive distributions in closed form,
hence making the optimization of this loss function computationally difficult.
Fortunately, such a loss function can be bounded from above by the KL divergence between posterior beliefs $q_u(\bm{\theta}|\dc)$ vs.~$p(\bm{\theta}|\dc)$, as proven in Appendix~\ref{app:klmarginal}:\vspace{1mm}
\begin{proposition}
\label{rmk:klmarginal}
%
$\text{KL}[q_u(y|\dc)\ \Vert\ p(y|\dc)] \le \text{KL}[q_u(\bm{\theta}|\dc)\ \Vert\ p(\bm{\theta}|\dc)]\ .$\footnote{Similarly, $\text{KL}[p(y|\dc)\ \Vert\ q_u(y|\dc)] \le \text{KL}[p(\bm{\theta}|\dc)\ \Vert\ q_u(\bm{\theta}|\dc)]$  holds.}
%
\end{proposition}
Proposition~\ref{rmk:klmarginal} reveals that reducing $\text{KL}[q_u(\bm{\theta}|\dc)\ \Vert\ p(\bm{\theta}|\dc)]$ decreases $\text{KL}[q_u(y|\dc)\ \Vert\ p(y|\dc)]$, thus motivating its use as the loss function instead. 
%
In particular, it follows immediately from our result below (i.e., proven in Appendix~\ref{app:eubo}) that
minimizing $\text{KL}[q_u(\bm{\theta}|\dc)\ \Vert\ p(\bm{\theta}|\dc)]$ is equivalent to \emph{minimizing} a counterpart of the ELBO called the \emph{evidence upper bound} (EUBO) $\mcl{U}$:\vspace{1mm}
\begin{proposition}
\label{theo:eubo}
Define the EUBO $\mcl{U}$ as
\begin{equation}
\mcl{U} \triangleq \int q_u(\bm{\theta}|\dc)\ \log p(\dr|\bm{\theta})\ \text{d}\bm{\theta} + \text{KL}[q_u(\bm{\theta}|\dc)\ \Vert\ p(\bm{\theta}|\da)]\ .
\label{eq:eubo}
\end{equation}
%
Then, $\mcl{U} = \log p(\dr|\dc) + \text{KL}[q_u(\bm{\theta}|\dc)\ \Vert\ p(\bm{\theta}|\dc)] \ge \log p(\dr|\dc)$ such that $p(\dr|\dc)$ is independent of $q_u(\bm{\theta}|\dc)$. 
\end{proposition}
From Proposition~\ref{theo:eubo}, minimizing EUBO~\eqref{eq:eubo} is equivalent to minimizing $\text{KL}[q_u(\bm{\theta}|\dc)\ \Vert\ p(\bm{\theta}|\dc)]$
which is precisely achieved using VI (i.e., by maximizing ELBO~\eqref{eq:elbo}) from retraining with remaining data $\dc$.
This is illustrated in 
Fig.~\ref{fig:eubovselbo}a where unlearning from $\dr$ by minimizing EUBO maximizes ELBO w.r.t.~$\dc$; 
in Fig.~\ref{fig:eubovselbo}b, retraining with $\dc$ by maximizing ELBO minimizes EUBO w.r.t.~$\dr$.

\todo{mention another form of evidence upper bound but only useful if exact posterior can be computed}
\todo{talk about stochastic optimization version of EUBO}
The EUBO $\mcl{U}$~\eqref{eq:eubo} can be interpreted as a trade-off between fully unlearning from erased data $\dr$ (first term) vs.~not entirely forgetting the exact posterior belief $p(\bm{\theta}|\da)$ given the full data $\da$ (i.e., including the remaining data $\dc$) (second term). The latter can be viewed as a regularization term to prevent catastrophic unlearning\cref{beef} (i.e., potentially induced by the former) 
that \emph{naturally} results from minimizing our loss function $\text{KL}[q_u(\bm{\theta}|\dc)\ \Vert\ p(\bm{\theta}|\dc)]$, which 
differs from the works of \cite{du2019lifelong,golatkar2019eternal} needing to ``patch up'' their loss functions (Sec.~\ref{sec:intro}).
Generative models can be used to model the approximate posterior belief $q_u(\bm{\theta}|\dc)$ in the EUBO $\mcl{U}$~\eqref{eq:eubo} in the same way as that in the ELBO $\mcl{L}$~\eqref{eq:elbo}.
%
%
%
\subsection{Approximate Bayesian Unlearning with Approximate Posterior Belief
$q($\texorpdfstring{$\bm{\theta}$}{theta}$|\da)$}
\label{subsec:apprfull}
Often, in model training, only an approximate posterior belief\footnote{With a slight abuse of notation, we let  $q(\bm{\theta}|\da)$
be the approximate posterior belief that maximizes the ELBO $\mcl{L}$~\eqref{eq:elbo} (Sec.~\ref{sec:vi}) from Sec.~\ref{subsec:apprfull} onwards.} $q(\bm{\theta}|\da)$ (instead of the exact $p(\bm{\theta}|\da)$ in Sec.~\ref{subsec:exactfull}) of model parameters $\bm{\theta}$ given full data $\da$ can be learned, say, using 
VI by maximizing the ELBO (Sec.~\ref{sec:vi}).
Our proposed unlearning methods are parsimonious in requiring only $q(\bm{\theta}|\da)$ and erased data $\dr$ to be available,
which makes unlearning even more challenging since there is no further 
information about $p(\bm{\theta}|\da)$ nor the difference between $p(\bm{\theta}|\da)$ vs.~$q(\bm{\theta}|\da)$. 
%
So, we estimate the unknown $p(\bm{\theta}|\dc)$~\eqref{eq:suffdata} with
%
\begin{equation}
\tilde{p}(\bm{\theta}|\dc)\ \  \propto\ \ {q(\bm{\theta}|\da)}/{p(\dr| \bm{\theta})}
\label{eq:postapprx}
\end{equation}
%
and minimize the KL divergence between the approximate posterior belief recovered by directly unlearning from erased data $\dr$
vs.~$\tilde{p}(\bm{\theta}|\dc)$~\eqref{eq:postapprx} instead. We will discuss  two novel tricks below to alleviate the undesirable consequence of using $\tilde{p}(\bm{\theta}|\dc)$ instead of the unknown $p(\bm{\theta}|\dc)$~\eqref{eq:suffdata}.
\subsubsection{EUBO with Adjusted Likelihood}
\label{subsubsec:eubo}
\begin{figure}
\centering
\includegraphics[width=0.63\textwidth]{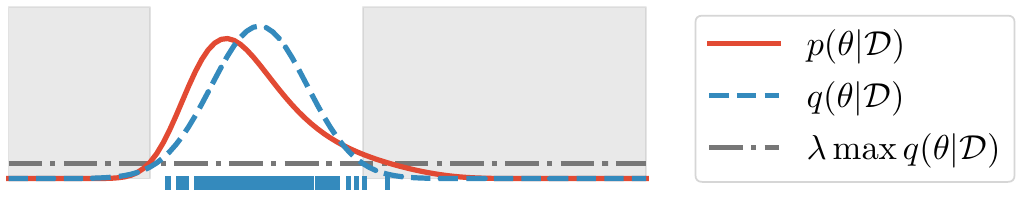}
\caption{Plot of  $q(\bm{\theta}|\da)$ learned using VI. Gray shaded region corresponds to values of $\bm{\theta}$ where $q(\bm{\theta}|\da) \le \lambda \max_{\bm{\theta}'} q(\bm{\theta}'|\da)$. Vertical blue strips on  horizontal axis show $100$ samples of $\bm{\theta} \sim q(\bm{\theta}|\da)$.}
\label{fig:adjustedvi}
\end{figure}

\begin{figure}
\centering
\includegraphics[width=\textwidth]{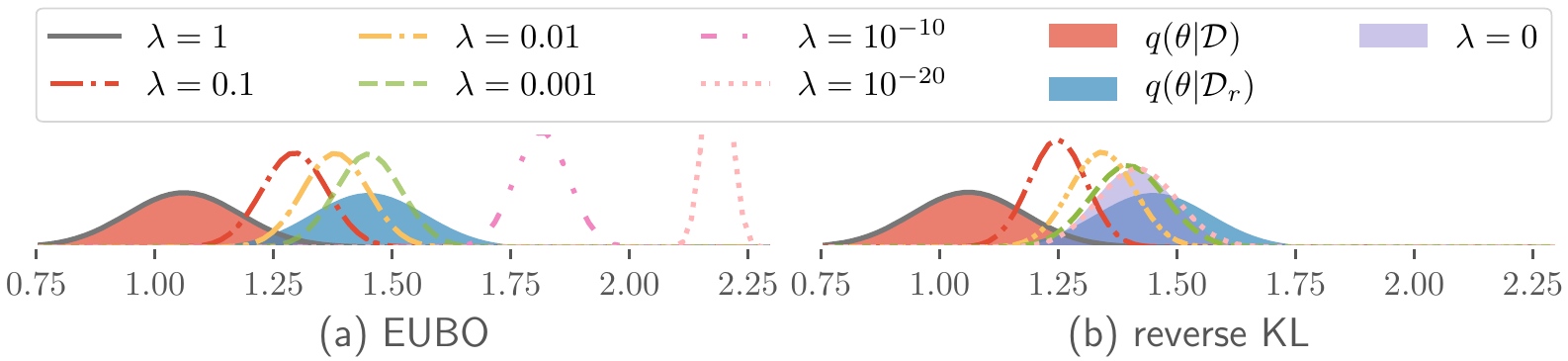}
\caption{Plot of approximate posterior beliefs with varying $\lambda$ obtained by minimizing (a) EUBO (i.e., $\eubo(\bm{\theta}|\dc;\lambda)$) and (b) reverse KL (i.e., $\elbo(\bm{\theta}|\dc;\lambda)$); horizontal axis denotes $\theta = \alpha$.
In (a), a huge probability
mass of $\eubo(\bm{\theta}|\dc, \lambda=0)$ is at large values of $\alpha$ beyond the plotting area and the top of the plot of $\eubo(\bm{\theta}|\dc, \lambda=10^{-20})$ is cut off due to lack of space.
}
\label{fig:expunlearn}
\end{figure}

Let the loss function $\text{KL}[\eubo(\bm{\theta}|\dc)\ \Vert\ \tilde{p}(\bm{\theta}|\dc)]$ be minimized w.r.t.~the approximate posterior belief $\eubo(\bm{\theta}|\dc)$ that is recovered by directly unlearning from erased data $\dr$. 
Similar to Proposition~\ref{theo:eubo}, $\eubo(\bm{\theta}|\dc)$ can be optimized by minimizing the following EUBO:
\begin{equation}
\widetilde{\mcl{U}} \triangleq \int \eubo(\bm{\theta}|\dc)\ \log p(\dr|\bm{\theta})\ \text{d}\bm{\theta} + \text{KL}[\eubo(\bm{\theta}|\dc)\ \Vert\ q(\bm{\theta}|\da)]
\label{eq:euboapprx}
\end{equation}
%
which follows from simply replacing the unknown $p(\bm{\theta}|\da)$ in $\mcl{U}$~\eqref{eq:eubo} with $q(\bm{\theta}|\da)$. 
We discuss the difference between $p(\bm{\theta}|\da)$ vs.~$q(\bm{\theta}|\da)$ in the remark below:\vspace{1mm}
\begin{remark}
\label{rmk:inacc}
We analyze two possible sources of inaccuracy in $q(\bm{\theta}|\da)$ that is learned using VI by minimizing the loss function $\text{KL}[q(\bm{\theta}|\da)\ \Vert\ p(\bm{\theta}|\da)]$ (Sec.~\ref{sec:vi}).
Firstly, $q(\bm{\theta}|\da)$ often underestimates the variance of $p(\bm{\theta}|\da)$: Though $q(\bm{\theta}|\da)$ tends to be close to $0$ at values of $\bm{\theta}$ where $p(\bm{\theta}|\da)$ is close to $0$, the reverse is not enforced \cite{bishop2006pattern} (see, for example, Fig.~\ref{fig:adjustedvi}).
So, $q(\bm{\theta}|\da)$ can differ from $p(\bm{\theta}|\da)$ at values of $\bm{\theta}$ where $q(\bm{\theta}|\da)$ is close to $0$.
Secondly, if $q(\bm{\theta}|\da)$ is learned through stochastic optimization of the ELBO (i.e., with stochastic samples of $\bm{\theta} \sim q(\bm{\theta}|\da)$ in each iteration of SGA), then it is unlikely that the ELBO is maximized using samples of $\bm{\theta}$ with small $q(\bm{\theta}|\da)$ (Fig.~\ref{fig:adjustedvi}). 
Thus, both sources of inaccuracy primarily occur at values of $\bm{\theta}$ with small $q(\bm{\theta}|\da)$. Though it can also be inaccurate at values of $\bm{\theta}$ with large $q(\bm{\theta}|\da)$, such an inaccuracy can be reduced by representing $q(\bm{\theta}|\da)$ with a complex distribution (Sec.~\ref{sec:vi}).
\end{remark}
Remark~\ref{rmk:inacc} motivates us to curb unlearning at values of $\bm{\theta}$ with small $q(\bm{\theta}|\da)$ by proposing our first novel trick of an adjusted likelihood of the erased data:
%
\begin{equation}
p_{\text{adj}}(\dr|\bm{\theta}; \lambda) \triangleq 
	\begin{cases}
		p(\dr|\bm{\theta}) &\text{ if } q(\bm{\theta}|\da) > \lambda \max_{\bm{\theta}'} q(\bm{\theta}'|\da)\ ,\\
		1 &\text{ otherwise (i.e., shaded area in Fig.~\ref{fig:adjustedvi})}\ ;
	\end{cases}
\label{eq:likelihoodadj}
\end{equation}
for any $\bm{\theta}$ where $\lambda \in [0,1]$ controls the magnitude of a threshold under which  $q(\bm{\theta}|\da)$ is considered small. To understand the effect of $\lambda$, let $\tilde{p}_{\text{adj}}(\bm{\theta}|\dc; \lambda) \propto q(\bm{\theta}|\da) / p_{\text{adj}}(\dr| \bm{\theta}; \lambda)$, i.e., by replacing $p(\dr| \bm{\theta})$ in \eqref{eq:postapprx} with $p_{\text{adj}}(\dr| \bm{\theta}; \lambda)$. Then, using~\eqref{eq:likelihoodadj},
\begin{equation}
\tilde{p}_{\text{adj}}(\bm{\theta}|\dc; \lambda) \propto 
	\begin{cases}
		q(\bm{\theta}|\da) / p(\dr| \bm{\theta}) &\text{ if } q(\bm{\theta}|\da) > \lambda \max_{\bm{\theta}'} q(\bm{\theta}'|\da)\ ,\\
		q(\bm{\theta}|\da) &\text{ otherwise (i.e., shaded area in Fig.~\ref{fig:adjustedvi})\ .}
	\end{cases}
\label{eq:postappradj}
\end{equation}
According to~\eqref{eq:postappradj}, unlearning is therefore  focused at values of $\bm{\theta}$ with sufficiently large $q(\bm{\theta}|\da)$ (i.e.,
$q(\bm{\theta}|\da) > \lambda \max_{\bm{\theta}'} q(\bm{\theta}'|\da)$).
Let the loss function $\text{KL}[\eubo(\bm{\theta}|\dc; \lambda)\ \Vert\ \tilde{p}_{\text{adj}}(\bm{\theta}|\dc; \lambda)]$ be minimized w.r.t.~the approximate posterior belief $\eubo(\bm{\theta}|\dc; \lambda)$ that is recovered by directly unlearning from erased data $\dr$.
Similar to~\eqref{eq:euboapprx},  
$\eubo(\bm{\theta}|\dc; \lambda)$ can be optimized by minimizing the following EUBO:
\begin{equation}
\widetilde{\mcl{U}}_{\text{adj}}(\lambda) \triangleq \int \eubo(\bm{\theta}|\dc; \lambda)\ \log p_{\text{adj}}(\dr|\bm{\theta}; \lambda)\ \text{d}\bm{\theta} + \text{KL}[\eubo(\bm{\theta}|\dc; \lambda)\ \Vert\ q(\bm{\theta}|\da)]
\label{eq:euboapprxadj}
\end{equation}
which follows from replacing $p(\dr|\bm{\theta})$ in \eqref{eq:euboapprx} with $p_{\text{adj}}(\dr|\bm{\theta}; \lambda)$.
Note that $\eubo(\bm{\theta}|\dc; \lambda)$ can be represented by a simple distribution (e.g., Gaussian) or a complex one (e.g., generative neural network, IAF). 
We initialize $\eubo(\bm{\theta}|\dc; \lambda)$ at $q(\bm{\theta}|\da)$ for achieving empirically faster convergence.
When $\lambda=0$, $\widetilde{\mcl{U}}_{\text{adj}}(\lambda=0)$ reduces to $\widetilde{\mcl{U}}$~\eqref{eq:euboapprx}, i.e., EUBO is minimized without the adjusted likelihood. As a result,  $\eubo(\bm{\theta}|\dc; \lambda=0)=\eubo(\bm{\theta}|\dc)$.
As $\lambda$ increases, unlearning is focused on a smaller and smaller region of $\bm{\theta}$ with sufficiently large $q(\bm{\theta}|\da)$.
When $\lambda$ reaches $1$, no unlearning is performed since 
$\tilde{p}_{\text{adj}}(\bm{\theta}|\dc; \lambda=1) = q(\bm{\theta}|\da)$, which results in $\eubo(\bm{\theta}|\dc; \lambda=1) = q(\bm{\theta}|\da)$ minimizing the loss function $\text{KL}[\eubo(\bm{\theta}|\dc; \lambda=1)\ \Vert\ \tilde{p}_{\text{adj}}(\bm{\theta}|\dc; \lambda=1)]$.\vspace{1mm} 
\todo{should we mention how to choose $\lambda$?}
\begin{example}
\label{exp:gamma}
To visualize the effect of varying $\lambda$ on $\eubo(\bm{\theta}|\dc;\lambda)$, we consider learning the shape $\alpha$ of a Gamma distribution with a known rate 
(i.e., $\bm{\theta} = \alpha$): $\da$ are $20$ samples of the Gamma distribution, $\dr$ are the smallest $5$ samples in $\da$, and the (non-conjugate) prior belief and approximate 
posterior beliefs of $\alpha$ are all Gaussians.
Fig.~\ref{fig:expunlearn}a shows the approximate posterior beliefs  $\eubo(\bm{\theta}|\dc; \lambda)$ with varying $\lambda$ as well as  $q(\bm{\theta}|\da)$ and 
$q(\bm{\theta}|\dc)$ learned using VI. 
As explained above, $\eubo(\bm{\theta}|\dc, \lambda=1) = q(\bm{\theta}|\da)$. 
When $\lambda = 0.001$, $\eubo(\bm{\theta}|\dc, \lambda=0.001)$ is close to 
$q(\bm{\theta}|\dc)$. 
However, as $\lambda$ decreases to $0$, $\eubo(\bm{\theta}|\dc, \lambda)$ moves away from $q(\bm{\theta}|\dc)$.
\end{example}
%
%
\todo{need details about this example}
%
The optimized $\eubo(\bm{\theta}|\dc; \lambda)$
suffers from the same issue of underestimating the variance as $q(\bm{\theta}|\da)$ learned using VI (see Remark~\ref{rmk:inacc}), 
especially when $\lambda$ tends to $0$ (e.g., see $\eubo(\bm{\theta}|\dc; \lambda=10^{-20})$ in Fig.~\ref{fig:expunlearn}a).
Though this issue can be mitigated by tuning $\lambda$ in the adjusted likelihood~\eqref{eq:likelihoodadj}, we may not want to risk facing the consequence of picking a value of $\lambda$ that is too small.
So, in Sec.~\ref{subsubsec:rkl}, we will propose another novel trick that is not inconvenienced by this issue.
%
\subsubsection{Reverse KL}
\label{subsubsec:rkl}
%
Let the loss function be the reverse KL divergence $\text{KL}[\tilde{p}(\bm{\theta}|\dc)\ \Vert\ \elbo(\bm{\theta}|\dc)]$ that is minimized w.r.t.~the approximate posterior belief $\elbo(\bm{\theta}|\dc)$ recovered by directly unlearning from erased data $\dr$. 
In contrast to the optimized $\eubo(\bm{\theta}|\dc; \lambda)$
from minimizing EUBO~\eqref{eq:euboapprxadj}, 
the optimized $\elbo(\bm{\theta}|\dc)$ from minimizing the reverse KL divergence 
overestimates (instead of underestimates) the variance of $\tilde{p}(\bm{\theta}|\dc)$ \cite{bishop2006pattern}: If $\tilde{p}(\bm{\theta}|\dc)$ is close to $0$, then $\elbo(\bm{\theta}|\dc)$ is not necessarily close to $0$.
From~\eqref{eq:postapprx}, the reverse KL divergence can be expressed as follows:
\begin{equation}
\text{KL}[\tilde{p}(\bm{\theta}|\dc)\ \Vert\ \elbo(\bm{\theta}|\dc)] 
	= C_0 - C_1\  \mbb{E}_{q(\bm{\theta}|\da)} \left[(\log \elbo(\bm{\theta}|\dc))/{p(\dr|\bm{\theta})}  \right]
\label{eq:reversekl}
\end{equation}
where $C_0$ and $C_1$ are constants independent of $\elbo(\bm{\theta}|\dc)$. So, the reverse KL divergence~\eqref{eq:reversekl} can be minimized by maximizing $\mbb{E}_{q(\bm{\theta}|\da)} [ (\log \elbo(\bm{\theta}|\dc)) / p(\dr|\bm{\theta}) ]$ with \emph{stochastic gradient ascent} (SGA).
We also initialize $\elbo(\bm{\theta}|\dc)$ at $q(\bm{\theta}|\da)$ for achieving empirically faster convergence.
Since stochastic optimization is performed with samples of $\bm{\theta} \sim q(\bm{\theta}|\da)$ in each iteration of SGA, it is unlikely that the reverse KL divergence~\eqref{eq:reversekl} is minimized using samples of $\bm{\theta}$ with small $q(\bm{\theta}|\da)$.
This naturally curbs unlearning at values of $\bm{\theta}$ with small $q(\bm{\theta}|\da)$, as motivated by Remark~\ref{rmk:inacc}.
%
%
%
On the other hand, it is still possible to employ the same trick of adjusted likelihood (Sec.~\ref{subsubsec:eubo}) by minimizing the reverse KL divergence  $\text{KL}[\tilde{p}_{\text{adj}}(\bm{\theta}|\dc; \lambda)\ \Vert\ \elbo(\bm{\theta}|\dc; \lambda)]$ as the loss function or, equivalently,  maximizing 
$\mbb{E}_{q(\bm{\theta}|\da)} [(\log \elbo(\bm{\theta}|\dc; \lambda))/p_{\text{adj}}(\dr|\bm{\theta}; \lambda) ]$
%
%
%
where $p_{\text{adj}}(\dr|\bm{\theta}; \lambda)$ and $\tilde{p}_{\text{adj}}(\bm{\theta}|\dc; \lambda)$ are previously defined in \eqref{eq:likelihoodadj} and \eqref{eq:postappradj}, respectively.
The role of $\lambda$ here is the same as that in \eqref{eq:euboapprxadj}. 

To illustrate the difference between minimizing the reverse KL divergence~\eqref{eq:reversekl} and EUBO~\eqref{eq:euboapprxadj}, Fig.~\ref{fig:expunlearn}b shows the approximate  posterior beliefs $\elbo(\bm{\theta}|\dc; \lambda)$ with varying $\lambda$. It can be observed that $\elbo(\bm{\theta}|\dc; \lambda=1) = q(\bm{\theta}|\da)$ (i.e., no unlearning). In contrast to minimizing EUBO (Fig.~\ref{fig:expunlearn}a), as $\lambda$ decreases to $0$, $\elbo(\bm{\theta}|\dc; \lambda)$ does not deviate  that much from  $q(\bm{\theta}|\dc)$, even when $\lambda = 0$ (i.e.,  the reverse KL divergence is minimized without the adjusted likelihood). 
This is because
the optimized $\elbo(\bm{\theta}|\dc; \lambda)$ is naturally more protected from both sources of inaccuracy (Remark~\ref{rmk:inacc}), as explained above.
Hence, we do not have to be as concerned about picking a small value of $\lambda$, which is also consistently observed in our experiments  (Sec.~\ref{sec:experiment}).
%
\section{Experiments and Discussion}
\label{sec:experiment}
This section empirically demonstrates our unlearning methods on Bayesian models such as sparse Gaussian process and logistic regression using synthetic and real-world datasets.
Further experimental results on Bayesian linear regression and with a bimodal posterior belief are reported in Appendices~\ref{app:lr} and~\ref{app:multimode}, respectively.
In our experiments, each dataset comprises pairs of input $\mbf{x}$ and its corresponding output/observation $y_{\mbf{x}}$.
We use RMSProp as the SGA algorithm with a learning rate of $10^{-4}$.
To assess the performance of our unlearning methods (i.e., by directly unlearning from erased data $\dr$),
we consider the difference between their induced predictive distributions vs.~that obtained using VI from retraining with remaining data $\dc$,
as motivated from Sec.~\ref{subsec:exactfull}.
To do this, we use a {\bf performance metric} that measures the KL divergence between the approximate predictive distributions
\[
\eubo(y_{\mbf{x}}|\dc) \triangleq \int p(y_{\mbf{x}}|\bm{\theta})\  \eubo(\bm{\theta}|\dc;\lambda)\ \text{d}\bm{\theta}
\quad\text{or}\quad
\elbo(y_{\mbf{x}}|\dc) \triangleq \int 
p(y_{\mbf{x}}|\bm{\theta})\ \elbo(\bm{\theta}|\dc;\lambda)\ \text{d}\bm{\theta}
\]
vs.~$q(y_{\mbf{x}}|\dc) \triangleq \int  p(y_{\mbf{x}}|\bm{\theta})\ q(\bm{\theta}|\dc)\  \text{d}\bm{\theta}$ where 
$\eubo(\bm{\theta}|\dc; \lambda)$ and $\elbo(\bm{\theta}|\dc; \lambda)$
are optimized
by, respectively, minimizing  EUBO~\eqref{eq:euboapprxadj} and \emph{reverse KL} (rKL) divergence~\eqref{eq:reversekl} while requiring only $q(\bm{\theta}|\da)$ and erased data $\dr$ (Sec.~\ref{subsec:apprfull}), and $q(\bm{\theta}|\dc)$ is learned using VI (Sec.~\ref{sec:vi}).
The above predictive distributions are computed via sampling with $100$ samples of $\bm{\theta}$.
For tractability reasons,
we evaluate the above performance metric over  finite input domains, specifically, over that in $\dr$ and $\dc$, which allows us to assess the 
performance of our unlearning methods on both the erased and remaining data, i.e., whether they can fully unlearn from $\dr$ yet not forget nor catastrophically unlearn from $\dc$, respectively.
For example, the performance of our EUBO-based unlearning method over $\dr$ is shown as an average (with standard deviation) of the KL divergences between $\eubo(y_{\mbf{x}}|\dc)$ vs.~$q(y_{\mbf{x}}|\dc)$ over all $(\mbf{x},y_{\mbf{x}}) \in \dr$.
%
We also plot an average (with standard deviation) of the KL divergences between $q(y_{\mbf{x}}|\da)$ vs.~$q(y_{\mbf{x}}|\dc)$ over $\dc$ and $\dr$ as \emph{baselines} (i.e., representing no unlearning), which is expected to be larger than that of our unlearning methods (i.e., if performing well)
and labeled as \emph{full} in the plots below.
\subsection{Sparse Gaussian Process (GP) Classification with Synthetic Moon Dataset}
\label{subsec:expmoon}
\begin{figure}
\hspace{-2mm}
\begin{tabular}{@{}c@{}}
    \begin{tabular}{@{}c@{}c@{}c@{}c@{}c@{}}
    \includegraphics[trim={7mm 8mm 3mm 3mm}, clip,height=0.16\textwidth]{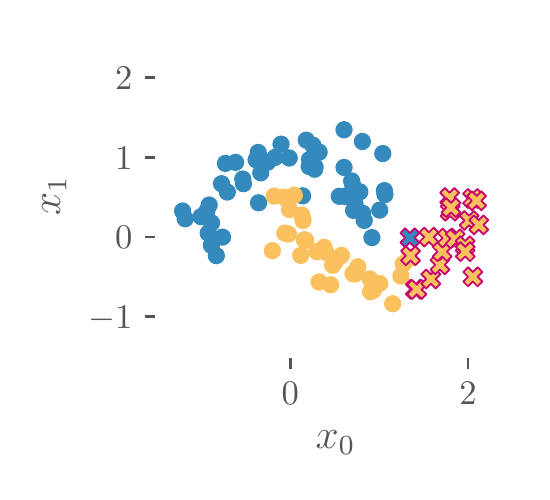}
    &
    \includegraphics[height=0.15\textwidth]{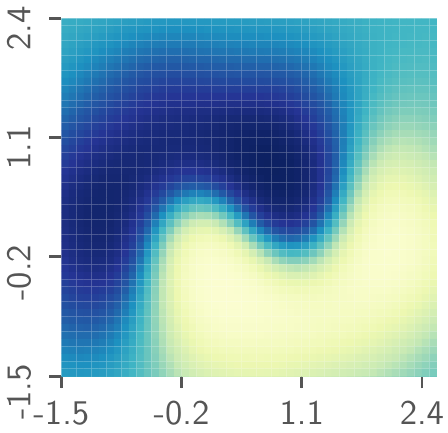}
    &
    \includegraphics[height=0.15\textwidth]{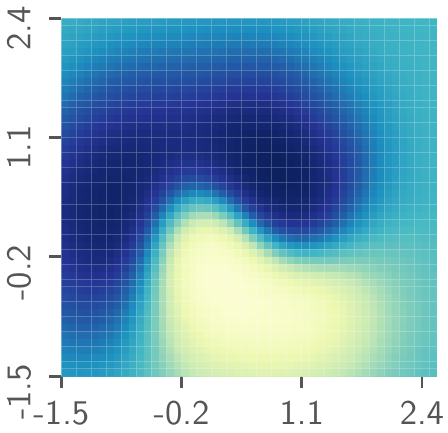}
    & \hspace{-2mm}
    \includegraphics[height=0.15\textwidth]{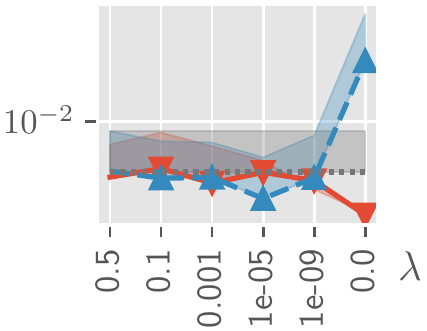}
    &
    \includegraphics[height=0.15\textwidth]{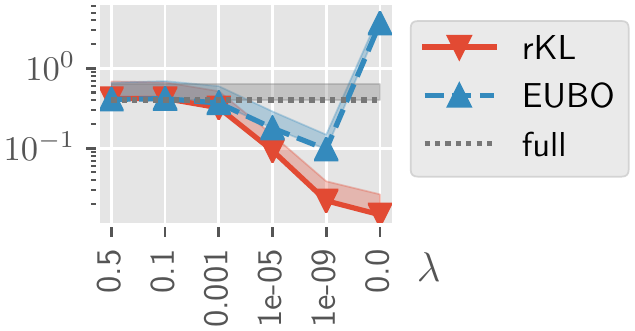}
    \\
    (a) Dataset
    &
    (b) $q(y_{\mbf{x}}=1| \da)$ 
    & \hspace{1mm}
    (c) $q(y_{\mbf{x}}=1| \dc)$ 
    &
    (d) $\dc$
    &
    (e) $\dr$
    \end{tabular}
\vspace{2mm}\\
    \begin{tabular}{@{}c@{}c@{}c@{}|@{}c@{}c@{}c@{}}
    \includegraphics[height=0.135\textwidth]{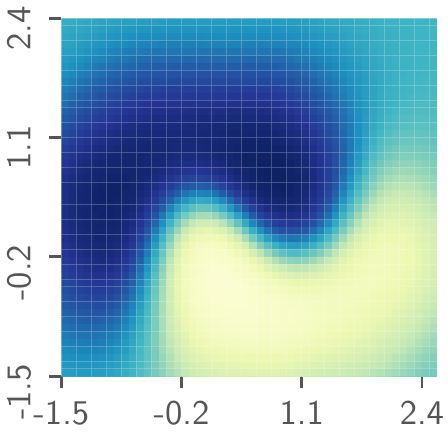}
    &\hspace{2.5mm}
    \includegraphics[height=0.135\textwidth]{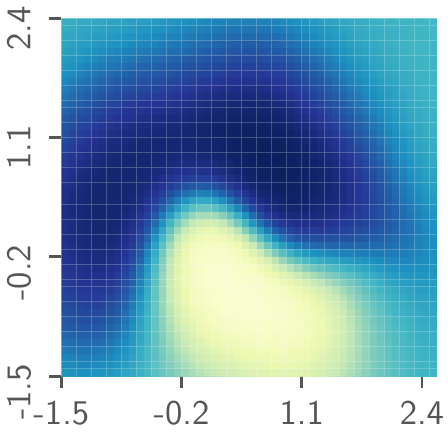}
    &\hspace{2.5mm}
    \includegraphics[height=0.135\textwidth]{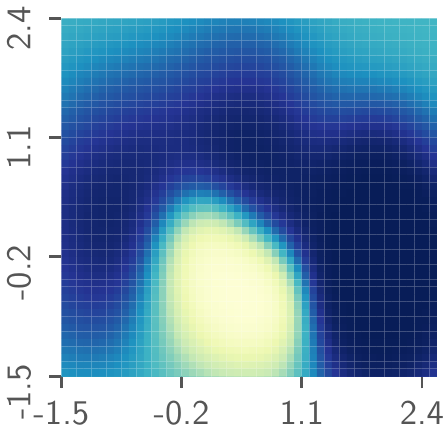}\hspace{2mm}
    &
    \hspace{2mm}
    \includegraphics[height=0.135\textwidth]{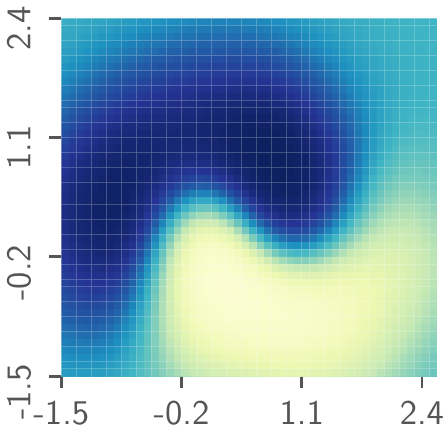}
    &\hspace{2.5mm}
    \includegraphics[height=0.135\textwidth]{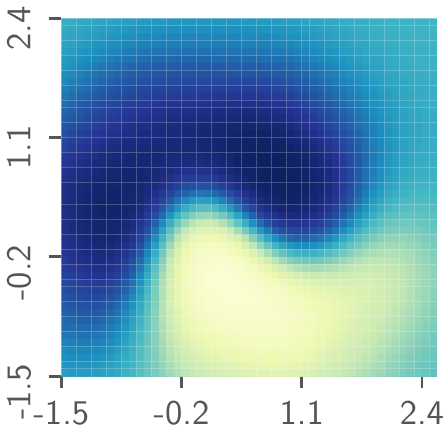}
    &\hspace{2.5mm}
    \includegraphics[height=0.135\textwidth]{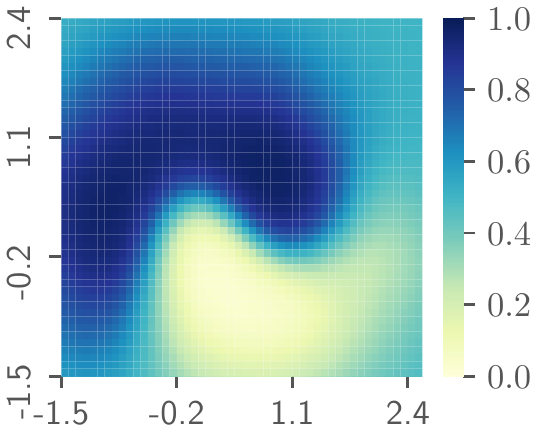}
    \\
    (f) $\lambda = 10^{-5}$
    &
    (g) $\lambda = 10^{-9}$
    &
    (h) $\lambda = 0$
    &
    (i) $\lambda = 10^{-5}$
    &
    (j) $\lambda = 10^{-9}$
    &
    (k) $\lambda = 0$
    \end{tabular}
\end{tabular}\vspace{-0.7mm}
\caption{Plots of (a) synthetic moon dataset with erased data $\dr$ (crosses) and remaining data $\dc$ (dots), and of predictive distributions 
obtained using VI from (b) training with full data $\da$ and (c) retraining with $\dc$.
Graphs of averaged KL divergence vs.~$\lambda$ achieved by EUBO, rKL, and $q(\bm{\theta}|\da)$ (i.e., baseline labeled as \emph{full}) over (d) $\dc$ and (e) $\dr$. Plots of predictive distributions (f-h) $\eubo(y_{\mbf{x}}=1|\dc)$ and (i-k) $\elbo(y_{\mbf{x}}=1|\dc)$ induced, respectively, by EUBO and rKL for varying $\lambda$.
}
\label{fig:moon}\vspace{-0.6mm}
\end{figure}

This experiment is about unlearning a binary classifier that is previously trained with the synthetic moon dataset (Fig.~\ref{fig:moon}a). 
The probability of input $\mbf{x} \in \mbb{R}^2$ being in the `blue' class (i.e., $y_{\mbf{x}} = 1$ and denoted by blue dots in Fig.~\ref{fig:moon}a) is defined as $1/(1 + \exp(f_{\mbf{x}}))$ where $f_{\mbf{x}}$ is a latent function modeled by a sparse GP \cite{quinonero2005unifying}, which is elaborated in Appendix~\ref{app:moon}. 
The parameters $\bm{\theta}$ of the sparse GP consist of $20$ inducing variables;
 the approximate
posterior beliefs of $\bm{\theta}$ are thus 
multivariate Gaussians (with full covariance matrices), as shown in Appendix~\ref{app:moon}.
By comparing Figs.~\ref{fig:moon}b and~\ref{fig:moon}c, it can be observed that after erasing $\dr$ (i.e., mainly in `yellow' class), $q(y_{\mbf{x}} = 1|\dc)$ increases at $\mbf{x} \in \dr$.
Figs.~\ref{fig:moon}d~and~\ref{fig:moon}e show results of the performance of our EUBO- and rKL-based unlearning methods over $\dc$ and $\dr$ with varying $\lambda$, respectively.\footnote{\label{log}Note that the log plots can only properly display the upper confidence interval of $1$ standard deviation (shaded area) and hence do not show the lower confidence interval.
}
When $\lambda = 10^{-9}$, EUBO performs reasonably well (compare Figs.~\ref{fig:moon}g~vs.~\ref{fig:moon}c) as its averaged KL divergence is smaller than that of $q(\bm{\theta}|\da)$ (i.e., baseline labeled as \emph{full}).
When $\lambda = 0$, EUBO performs poorly (compare Figs.~\ref{fig:moon}h~vs.~\ref{fig:moon}c) as its averaged KL divergence is much larger than that of $q(\bm{\theta}|\da)$, as shown in Figs.~\ref{fig:moon}d and~\ref{fig:moon}e.
This agrees with our discussion of the issue with picking too small a value of $\lambda$ for EUBO at the end of Sec.~\ref{subsubsec:eubo}.
In particular, catastrophic unlearning is observed as the input region containing $\dr$ (i.e., mainly in `yellow' class) has a high probability in `blue' class after unlearning in Fig.~\ref{fig:moon}h. 
On the other hand, when $\lambda = 0$, 
rKL performs well (compare Figs.~\ref{fig:moon}k~vs.~\ref{fig:moon}c) as its
KL divergence is much smaller than that of $q(\bm{\theta}|\da)$, as seen in  Figs.~\ref{fig:moon}d and~\ref{fig:moon}e. This agrees with our discussion at the end of Sec.~\ref{subsubsec:rkl} that rKL can work well without needing the adjusted likelihood. 

One may question how the performance of our unlearning methods would vary when erasing a large quantity of data or with different distributions of erased data (e.g., erasing the data randomly vs.~deliberately erasing all data in a given class). 
To address this question, 
we have discovered that a key factor influencing their unlearning performance in these scenarios is the difference between the posterior beliefs of model parameters $\bm{\theta}$ given remaining data $\dc$ vs.~that given full data $\da$, especially at values of $\bm{\theta}$ with small $q(\bm{\theta}|\da)$ since unlearning in such a region is curbed by the adjusted likelihood and reverse KL.
In practice, we expect such a difference not to be large due to the small quantity of erased data and redundancy in real-world datasets.
%
%
We will present the details of this study in Appendix~\ref{app:information} due to lack of space by considering how much $\dr$ reduces the entropy of $\bm{\theta}$ given $\dc$.
\subsection{Logistic Regression with Banknote Authentication Dataset}
\label{subsec:experimentbanknote}
\todo{structure of the IAF}
The banknote authentication dataset~\cite{Dua:2019} of size $|\da|=1372$
is partitioned into erased data of size $|\dr|=412$ and remaining data of size $|\dc|=960$.
Each input $\mbf{x}$ comprises $4$ features extracted from an image of a banknote and its corresponding binary label $y_\mbf{x}$ indicates whether the banknote is genuine or forged. 
We use a logistic regression model with $5$ parameters that is trained with this dataset. The prior beliefs of the model parameters are independent Gaussians $\mcl{N}(0,100)$. 

Unlike the previous experiment, the erased data $\dr$ here is randomly selected and hence does not reduce the entropy of model parameters $\bm{\theta}$ given $\dc$ much, as explained in Appendix~\ref{app:information};
a discussion on erasing informative data (such as that in Sec.~\ref{subsec:expmoon}) is in Appendix~\ref{app:information}.
As a result, Figs.~\ref{fig:authenresult}a and~\ref{fig:authenresult}b show a very small averaged KL divergence of about $10^{-3}$ between $q(y_{\mbf{x}}|\da)$ vs.~$q(y_{\mbf{x}}|\dc)$ 
(i.e., baselines) over $\dc$ and $\dr$.\cref{log}
Figs.~\ref{fig:authenresult}a and~\ref{fig:authenresult}b
also show that our unlearning methods do not perform well 
when using multivariate Gaussians to model the approximate posterior beliefs of $\bm{\theta}$:
While rKL still gives a useful $\elbo(\bm{\theta}|\dc; \lambda)$ achieving an averaged KL divergence close to that of $q(\bm{\theta}|\da)$,
EUBO gives a useless $\eubo(\bm{\theta}|\dc; \lambda)$ incurring a large averaged KL divergence 
when $\lambda$ is small. 
On the other hand, when 
more complex models like  normalizing flows with the MADE architecture~\cite{papamakarios2017masked} are used to represent the approximate posterior beliefs, EUBO and rKL can unlearn well (Figs.~\ref{fig:authenresult}c and~\ref{fig:authenresult}d).
\begin{figure}
\centering
\begin{tabular}{@{}c@{}c@{}c@{}c@{}}
\includegraphics[height=0.18\textwidth]{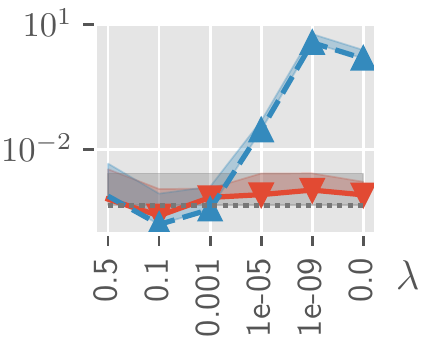}
&
\includegraphics[height=0.17\textwidth]{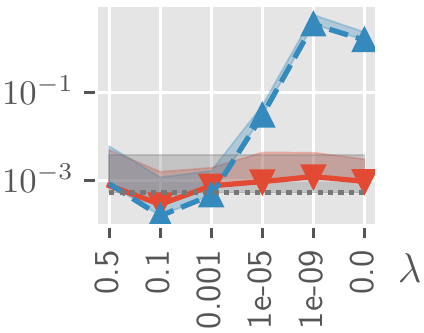}
&
\includegraphics[height=0.17\textwidth]{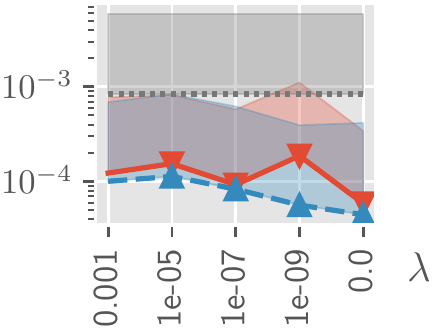}
&
\includegraphics[height=0.17\textwidth]{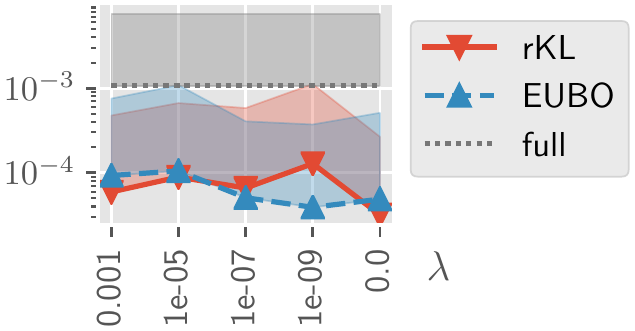}
\\
(a) $\dc$
&
(b) $\dr$
&
(c) $\dc$
&
(d) $\dr$
\end{tabular}\vspace{-2mm}
\caption{Graphs of averaged KL divergence vs.~$\lambda$ achieved by EUBO, rKL, and $q(\bm{\theta}|\da)$ (i.e., baseline labeled as \emph{full}) over $\dc$ and $\dr$ 
for the banknote authentication dataset with the approximate posterior beliefs of model parameters represented by (a-b) multivariate Gaussians and (c-d) normalizing flows.}
\label{fig:authenresult}\vspace{-2.8mm}
\end{figure}
%
\subsection{Logistic Regression with Fashion MNIST Dataset}
The fashion MNIST dataset of size $|\da| = 60000$ ($28\times28$ images of fashion items in $10$ classes) is partitioned into erased data of size $|\dr| = 10000$ and remaining data of size $|\dc| = 50000$.
The classification model is a neural network with $3$ fully-connected hidden layers of $128$, $128$, $64$ hidden neurons and a softmax layer to output the $10$-class probabilities.
The model can be interpreted as one of logistic regression on $64$ features generated from the hidden layer of $64$ neurons.
Since modeling all weights of the neural network as random variables can be costly, we model only $650$ weights in the transformation of the $64$ features to the inputs of the softmax layer.
The other weights remain constant during unlearning and retraining.
The prior beliefs of the network weights are $\mcl{N}(0, 10)$. The approximate posterior beliefs are modeled with independent Gaussians.
Though a large part of the network is fixed and we use simple models to represent the approximate posterior beliefs, we show that unlearning is still fairly effective.

\begin{figure}
\centering
\begin{tabular}{@{}c|c@{}c@{}}
\includegraphics[width=0.3\textwidth]{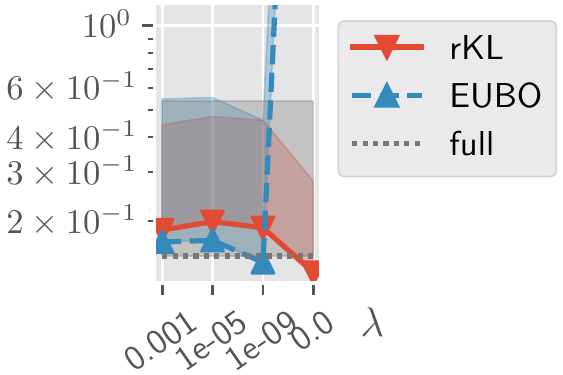}
&
\makecell[t]{\includegraphics[width=0.07\textwidth]{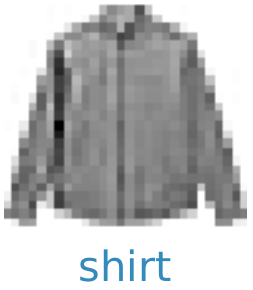}} 
&
\includegraphics[width=0.55\textwidth]{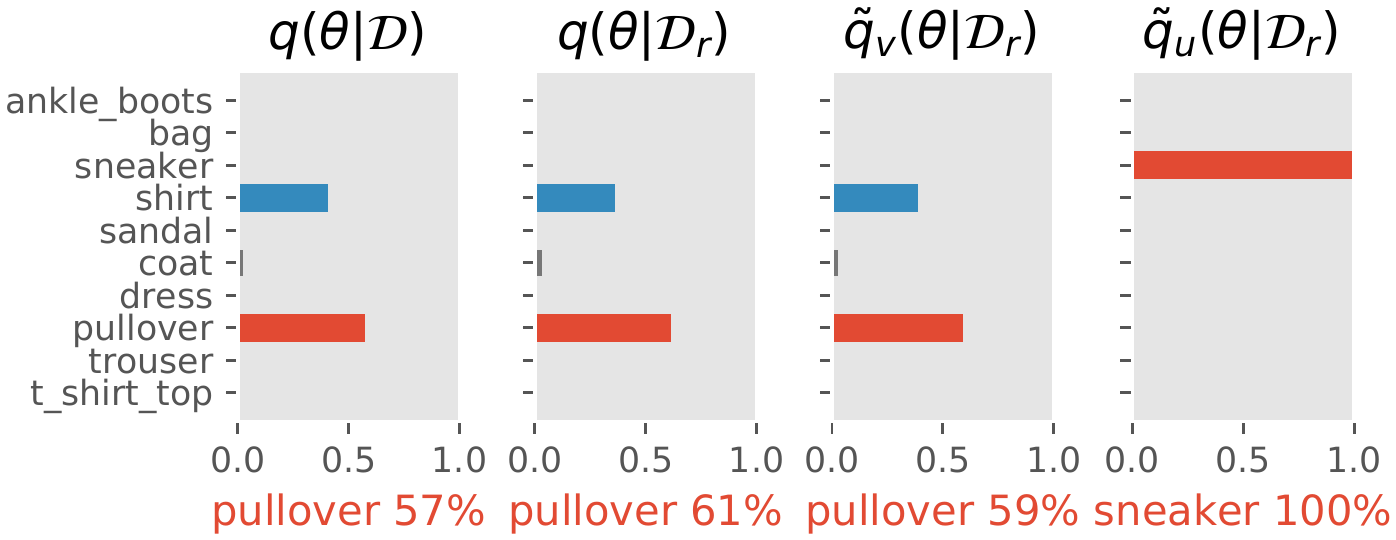} 
\\
(a) $\dc$ & \multicolumn{2}{c}{(c) Class probabilities for a `shirt' image}
\vspace{1mm}\\
\includegraphics[width=0.3\textwidth]{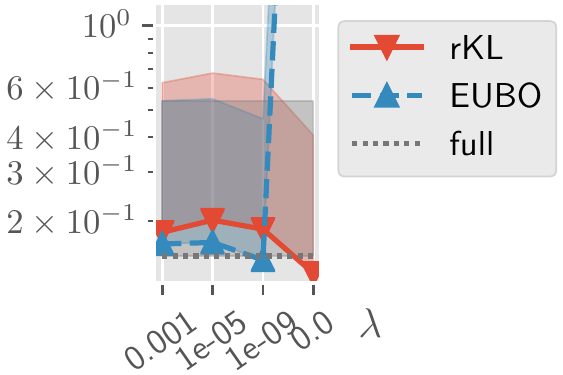}
&
\makecell[t]{\includegraphics[width=0.07\textwidth]{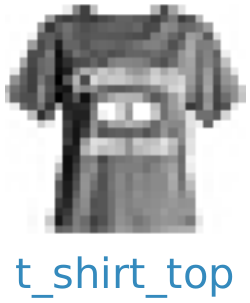}} 
&
\includegraphics[width=0.55\textwidth]{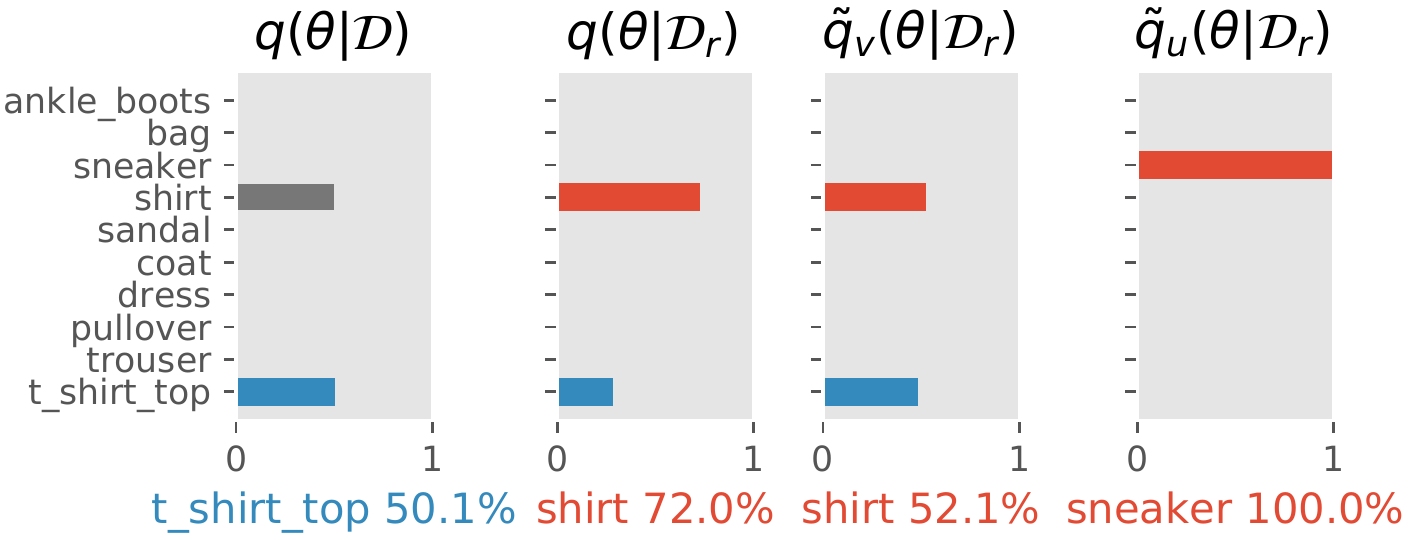} 
\\
(b) $\dr$ & \multicolumn{2}{c}{(d) Class probabilities for a `T-shirt' image}
\end{tabular}
\caption{Graphs of averaged KL divergence vs.~$\lambda$ achieved by EUBO, rKL, and $q(\bm{\theta}|\da)$ (i.e., baseline labeled as \emph{full}) over (a) $\dc$ and (b) $\dr$. (c-d) Plots of class probabilities for two images in the fashion MNIST dataset obtained using $q(\bm{\theta}|\da)$, $q(\bm{\theta}|\dc)$, optimized $\elbo(\bm{\theta}|\dc; \lambda = 0)$ and $\eubo(\bm{\theta}|\dc; \lambda = 0)$.}
\label{fig:mnistresult}
\end{figure}

As discussed in Sec.~\ref{subsec:expmoon},~\ref{subsec:experimentbanknote}, and Appendix~\ref{app:information}, the random selection of erased data $\dr$ and  redundancy in $\da$ lead to a small averaged KL divergence of about $0.1$ between $q(y_{\mbf{x}}|\da)$ vs.~$q(y_{\mbf{x}}|\dc)$ (i.e., baselines) over $\dc$ and $\dr$ (Figs.~\ref{fig:mnistresult}a and~\ref{fig:mnistresult}b) despite choosing a relatively large $|\dr|$.
Figs.~\ref{fig:mnistresult}a and~\ref{fig:mnistresult}b  show that when $\lambda \ge 10^{-9}$, EUBO and rKL achieve averaged KL divergences comparable to that of $q(\bm{\theta}|\da)$ (i.e., baseline labeled as \emph{full}), hence making their unlearning insignificant.\cref{log} 
However, at $\lambda = 0$, the unlearning performance of rKL improves by achieving a smaller averaged KL divergence than that of $q(\bm{\theta}|\da)$, while EUBO's performance deteriorates.
Their performance can be further improved by using more complex models to represent their approximate posterior beliefs like that in Sec.~\ref{subsec:experimentbanknote}, albeit high-dimensional.
Figs.~\ref{fig:mnistresult}c and~\ref{fig:mnistresult}d show the class probabilities for two images evaluated at the mean of the approximate posterior beliefs with $\lambda = 0$. 
We observe that rKL induces the highest class probability for the same class as that of
$q(\bm{\theta}|\dc)$.
The class probabilities for other images 
are shown in Appendix~\ref{app:fashionmnist}.
The two images 
are taken from a separate  set of $10000$ test images (i.e., different from $\da$) where rKL with $\lambda = 0$ yields the same predictions as $q(\bm{\theta}|\dc)$ and $q(\bm{\theta}|\da)$ in, respectively, $99.34\%$ and $99.22\%$ of the test images, the latter of which are contained in the former.
%
%
%
%
%
%
%
%
\subsection{Sparse Gaussian Process (GP) Regression with Airline Dataset}
\label{subsec:airline}
This section illustrates the scalability of unlearning to the massive airline dataset of $\sim 2$ million flights~\cite{hensman2013gaussian}. 
Training a sparse GP model with this massive dataset is made possible through stochastic VI~\cite{hensman2013gaussian}. 
Let $\mcl{X}_\mbf{u}$ denote the set of $50$ inducing inputs in the sparse GP model and $\mbf{f}_{\mcl{X}_\mbf{u}}$ be a vector of corresponding latent function values (i.e., inducing variables). The posterior belief $p(\mbf{f}_{\mcl{D}},\mbf{f}_{\mcl{X}_\mbf{u}}|\mcl{D})$ is approximated as $q(\mbf{f}_{\mcl{D}},\mbf{f}_{\mcl{X}_\mbf{u}}|\mcl{D}) \triangleq q(\mbf{f}_{\mcl{X}_\mbf{u}}|\mcl{D})\ p(\mbf{f}_{\mcl{D}}| \mbf{f}_{\mcl{X}_\mbf{u}})$ where $\mbf{f}_{\mcl{D}} \triangleq (f_{\mbf{x}})_{\mbf{x} \in \mcl{D}}$.
Let the sets $\mcl{X}_{\mcl{D}}$ and $\mcl{X}_{\dr}$ denote inputs in the full and erased data, respectively.
Then, the ELBO can be decomposed to
%
%
\begin{equation}
\mcl{L} = \sum_{\mbf{x} \in \mcl{X}_{\mcl{D}}} \int q(\mbf{f}_{\mcl{X}_\mbf{u}}|\mcl{D})\  p(f_{\mbf{x}}| \mbf{f}_{\mcl{X}_\mbf{u}}) \log p(y_{\mbf{x}}|f_{\mbf{x}})\ \text{d}f_{\mbf{x}}\ \text{d}\mbf{f}_{\mcl{X}_\mbf{u}} 
	- \text{KL}[q(\mbf{f}_{\mcl{X}_\mbf{u}}|\mcl{D})\ \Vert\ p(\mbf{f}_{\mcl{X}_\mbf{u}})]
\label{eq:elbosvi}
\end{equation}
where $\int p(f_{\mbf{x}}| \mbf{f}_{\mcl{X}_\mbf{u}}) \log p(y_{\mbf{x}}|f_{\mbf{x}})\ \text{d}f_{\mbf{x}}$ can be evaluated in closed form~\cite{gal2014variational}.
To unlearn such a trained model from $\dr$ ($|\dr|=100$K here), the  EUBO~\eqref{eq:euboapprxadj} can be expressed in a similar way as the ELBO:
\begin{equation*}
\widetilde{\mcl{U}}_{\text{adj}}(\lambda)\hspace{-0.7mm}=\hspace{-1.2mm}\sum_{\mbf{x} \in \mcl{X}_{\dr}}\hspace{-1mm} \int\hspace{-0.7mm} \tilde{q}_u(\mbf{f}_{\mcl{X}_\mbf{u}}|\mcl{D}_r; \lambda) p(f_{\mbf{x}}| \mbf{f}_{\mcl{X}_\mbf{u}}) \log p_{\text{adj}}(y_{\mbf{x}}| f_{\mbf{x}}; \lambda)\ \text{d}f_{\mbf{x}}\ \text{d}\mbf{f}_{\mcl{X}_\mbf{u}} 
    \hspace{-0.5mm}+ \text{KL}[\eubo(\mbf{f}_{\mcl{X}_\mbf{u}}|\mcl{D}_r; \lambda) \Vert q(\mbf{f}_{\mcl{X}_\mbf{u}}|\mcl{D})]
\end{equation*}
where $p_{\text{adj}}(y_{\mbf{x}}| f_{\mbf{x}}; \lambda) = p(y_{\mbf{x}}| f_{\mbf{x}})$ if 
$q(f_{\mbf{x}},\mbf{f}_{\mcl{X}_\mbf{u}}|\mcl{D}) > \lambda \max_{\mbf{f}_{\mcl{X}_\mbf{u}}} q(f_{\mbf{x}},\mbf{f}_{\mcl{X}_\mbf{u}}|\mcl{D})$, 
and $p_{\text{adj}}(y_{\mbf{x}}| f_{\mbf{x}}; \lambda) = 1$ otherwise.
%
EUBO can be minimized using stochastic gradient descent with random subsets (i.e., mini-batches of size $10$K) of $\mcl{D}_e$ in each iteration. 
For rKL, we use the entire $\mcl{D}_e$ in each iteration.
Since $\eubo(\mbf{f}_{\mcl{X}_\mbf{u}}| \mcl{D}_r; \lambda)$, $\elbo(\mbf{f}_{\mcl{X}_\mbf{u}}| \mcl{D}_r; \lambda)$, and $q(\mbf{f}_{\mcl{X}_\mbf{u}}| \mcl{D}_r)$  in~\eqref{eq:elbosvi}~\cite{gal2014variational} are all multivariate Gaussians,
we can directly evaluate the  performance of EUBO and rKL with varying $\lambda$ through their respective $\text{KL}[\eubo(\mbf{f}_{\mcl{X}_\mbf{u}}| \mcl{D}_r; \lambda)\ \Vert\ q(\mbf{f}_{\mcl{X}_\mbf{u}}| \mcl{D}_r)]$ and $\text{KL}[\elbo(\mbf{f}_{\mcl{X}_\mbf{u}}| \mcl{D}_r; \lambda)\ \Vert\ q(\mbf{f}_{\mcl{X}_\mbf{u}}| \mcl{D}_r)]$ which, according to Table~\ref{tbl:airlineresult}, are smaller than $\text{KL}[q(\mbf{f}_{\mcl{X}_\mbf{u}}| \mcl{D})\ \Vert\ q(\mbf{f}_{\mcl{X}_\mbf{u}}| \mcl{D}_r)]$ of value $4344.09$ (i.e., baseline representing no unlearning), hence demonstrating reasonable unlearning performance.
%
%
%
\begin{table}
\centering
\caption{KL divergence achieved by EUBO (top row) and rKL (bottom row)
with varying $\lambda$ for airline dataset.}
\begin{tabular}{ccccc}
\toprule
$\lambda$ &  $10^{-11}$ & $10^{-13}$ & $10^{-20}$ & $0$\\
\midrule
$\text{KL}[\eubo(\mbf{f}_{\mcl{X}_\mbf{u}}| \mcl{D}_r; \lambda)\ \Vert\ q(\mbf{f}_{\mcl{X}_\mbf{u}}| \mcl{D}_r)]$
& $2194.49$ & $1943.00$ & $1384.96$ & $2629.71$
\\
$\text{KL}[\elbo(\mbf{f}_{\mcl{X}_\mbf{u}}| \mcl{D}_r; \lambda)\ \Vert\ q(\mbf{f}_{\mcl{X}_\mbf{u}}| \mcl{D}_r)]$
& $418.42$ & $367.12$ & $543.45$ & $455.11$ \\
\bottomrule
\end{tabular}
\label{tbl:airlineresult}
\end{table}
%
%
\section{Conclusion}
\label{sec:conclusion}
This paper describes novel  unlearning methods for approximately unlearning a Bayesian model from a small subset of training data to be erased. Our unlearning methods are parsimonious in requiring only the approximate posterior belief of model parameters given the full data (i.e., obtained in model training with VI) and erased data to be available. This makes unlearning even more challenging due to two sources of inaccuracy in the approximate posterior belief.
We introduce novel tricks of adjusted likelihood and reverse KL to curb unlearning in the region of model parameters with low approximate posterior belief where both sources of inaccuracy primarily occur.
Empirical evaluations on synthetic and real-world datasets show that our proposed methods (especially reverse KL without adjusted likelihood) can effectively unlearn Bayesian models such as sparse GP and logistic regression from erased data.
In practice, for the approximate posterior beliefs recovered by unlearning from erased data using our proposed methods, they can be immediately used in ML applications and continue to be improved at the same time
by retraining with the remaining data at the expense of parsimony.
In our future work, we will apply our our proposed methods to unlearning more sophisticated Bayesian models like the entire family of sparse GP models~\cite{Chen13,LowTASE15,LowRSS13,LowUAI12,MinhAAAI17,HoangICML19,hoang2015unifying,HoangICML16,NghiaAAAI19,LowECML14a,low15,Ruofei18,teng20,LowAAAI14,HaibinAPP}) and deep GP models~\cite{yu19}.

\clearpage
\section*{Broader Impact}


As discussed in our introduction (Sec.~\ref{sec:intro}), a direct contribution of our work to the society in this information age is to the implementation of \emph{personal data ownership} (i.e., enforced by the General Data Protection Regulation in the European Union \cite{mantelero2013eu}) by studying the problem of machine unlearning for Bayesian models. Such an implementation can boost the confidence of users about sharing their data with an application/organization when they know that the trace of their data can be reduced/erased, as requested.
As a result, organizations/applications can gather more useful data from users to enhance their service back to the users and hence to the society.

Our unlearning work can also contribute to the defense against data poisoning attacks (i.e., injecting malicious training data). Instead of retraining the tampered machine learning model from scratch to recover the quality of a service, unlearning the model from the detected malicious data may incur much less time, which improves the user experience and reduces the cost due to the service disruption.

In contrast, the ability to unlearn machine learning models may also open the door to new adversarial activities. For example, in the context of data sharing, multiple parties share their data to train a common machine learning model. An unethical party can deliberately share a low-quality dataset instead of its high-quality one. After obtaining the model trained on datasets from all parties (including the low-quality dataset), the unethical party can unlearn the low-quality dataset and continue to train the model with its high-quality dataset.
By doing this, the unethical party achieves a better model than other parties in the collaboration. Therefore, the possibility of machine unlearning should be considered in the design of different data sharing frameworks.


\begin{ack}
This research/project is supported by the National Research Foundation, Singapore under its Strategic Capability Research Centres Funding Initiative. Any opinions, findings and conclusions or recommendations expressed in this material are those of the author(s) and do not reflect the views of National Research Foundation, Singapore.
\end{ack}

    \bibliographystyle{abbrv}
    \bibliography{unlearn}

\begin{thebibliography}{10}

\bibitem{bishop2006pattern}
C.~M. Bishop.
\newblock {\em Pattern Recognition and Machine Learning}.
\newblock Springer, 2006.

\bibitem{blei2017variational}
D.~M. Blei, A.~Kucukelbir, and J.~D. McAuliffe.
\newblock Variational inference: {A} review for statisticians.
\newblock {\em J. {American} Statistical Association}, 112(518):859--877, 2017.

\bibitem{bourtoule2019machine}
L.~Bourtoule, V.~Chandrasekaran, C.~{Choquette-Choo}, H.~Jia, A.~Travers,
  B.~Zhang, D.~Lie, and N.~Papernot.
\newblock Machine unlearning.
\newblock {arXiv:1912.03817}, 2019.

\bibitem{cao2015towards}
Y.~Cao and J.~Yang.
\newblock Towards making systems forget with machine unlearning.
\newblock In {\em Proc. {IEEE} S\&P}, pages 463--480, 2015.

\bibitem{Chen13}
J.~Chen, N.~Cao, B.~K.~H. Low, R.~Ouyang, C.~K.-Y. Tan, and P.~Jaillet.
\newblock Parallel {Gaussian} process regression with low-rank covariance
  matrix approximations.
\newblock In {\em Proc. UAI}, pages 152--161, 2013.

\bibitem{LowTASE15}
J.~Chen, B.~K.~H. Low, P.~Jaillet, and Y.~Yao.
\newblock Gaussian process decentralized data fusion and active sensing for
  spatiotemporal traffic modeling and prediction in mobility-on-demand systems.
\newblock {\em {IEEE} Trans. Autom. Sci. Eng.}, 12:901--921, 2015.

\bibitem{LowRSS13}
J.~Chen, B.~K.~H. Low, and C.~K.-Y. Tan.
\newblock {Gaussian} process-based decentralized data fusion and active sensing
  for mobility-on-demand system.
\newblock In {\em Proc. RSS}, 2013.

\bibitem{LowUAI12}
J.~Chen, B.~K.~H. Low, C.~K.-Y. Tan, A.~Oran, P.~Jaillet, J.~M. Dolan, and
  G.~S. Sukhatme.
\newblock Decentralized data fusion and active sensing with mobile sensors for
  modeling and predicting spatiotemporal traffic phenomena.
\newblock In {\em Proc. UAI}, pages 163--173, 2012.

\bibitem{du2019lifelong}
M.~Du, Z.~Chen, C.~Liu, R.~Oak, and D.~Song.
\newblock Lifelong anomaly detection through unlearning.
\newblock In {\em Proc. CCS}, pages 1283--1297, 2019.

\bibitem{Dua:2019}
D.~Dua and C.~Graff.
\newblock {UCI} machine learning repository, 2017.

\bibitem{gal2014variational}
Y.~Gal and M.~{van der Wilk}.
\newblock Variational inference in sparse {Gaussian} process regression and
  latent variable models--a gentle tutorial.
\newblock {\em arXiv preprint arXiv}, 1402, 2014.

\bibitem{ginart2019making}
A.~Ginart, M.~Guan, G.~Valiant, and J.~Y. Zou.
\newblock Making {AI} forget you: {Data} deletion in machine learning.
\newblock In {\em Proc. {NeurIPS}}, pages 3513--3526, 2019.

\bibitem{golatkar2019eternal}
A.~Golatkar, A.~Achille, and S.~Soatto.
\newblock Eternal sunshine of the spotless net: {Selective} forgetting in deep
  neural networks.
\newblock In {\em Proc. CVPR}, 2020.

\bibitem{guo2019certified}
C.~Guo, T.~Goldstein, A.~Hannun, and L.~{van der Maaten}.
\newblock Certified data removal from machine learning models.
\newblock {arXiv:1911.03030}, 2019.

\bibitem{hensman2013gaussian}
J.~Hensman, N.~Fusi, and N.~D. Lawrence.
\newblock Gaussian processes for big data.
\newblock In {\em Proc. {UAI}}, pages 282--290, 2013.

\bibitem{MinhAAAI17}
Q.~M. Hoang, T.~N. Hoang, and B.~K.~H. Low.
\newblock A generalized stochastic variational {Bayesian} hyperparameter
  learning framework for sparse spectrum {Gaussian} process regression.
\newblock In {\em Proc. {AAAI}}, pages 2007--2014, 2017.

\bibitem{HoangICML19}
Q.~M. Hoang, T.~N. Hoang, B.~K.~H. Low, and C.~Kingsford.
\newblock Collective model fusion for multiple black-box experts.
\newblock In {\em Proc. ICML}, pages 2742--2750, 2019.

\bibitem{hoang2015unifying}
T.~N. Hoang, Q.~M. Hoang, and B.~K.~H. Low.
\newblock A unifying framework of anytime sparse {Gaussian} process regression
  models with stochastic variational inference for big data.
\newblock In {\em Proc. {ICML}}, pages 569--578, 2015.

\bibitem{HoangICML16}
T.~N. Hoang, Q.~M. Hoang, and B.~K.~H. Low.
\newblock A distributed variational inference framework for unifying parallel
  sparse {Gaussian} process regression models.
\newblock In {\em Proc. ICML}, pages 382--391, 2016.

\bibitem{NghiaAAAI19}
T.~N. Hoang, Q.~M. Hoang, B.~K.~H. Low, and J.~P. How.
\newblock Collective online learning of {Gaussian} processes in massive
  multi-agent systems.
\newblock In {\em Proc. {AAAI}}, pages 7850--7857, 2019.

\bibitem{kingma2016improved}
D.~P. Kingma, T.~Salimans, R.~Jozefowicz, X.~Chen, I.~Sutskever, and
  M.~Welling.
\newblock Improved variational inference with inverse autoregressive flow.
\newblock In {\em Proc. {NeurIPS}}, pages 4743--4751, 2016.

\bibitem{LowECML14a}
B.~K.~H. Low, N.~Xu, J.~Chen, K.~K. Lim, and E.~B. {\"{O}zg\"{u}l}.
\newblock Generalized online sparse {Gaussian} processes with application to
  persistent mobile robot localization.
\newblock In {\em Proc. {ECML/PKDD Nectar Track}}, pages 499--503, 2014.

\bibitem{low15}
B.~K.~H. Low, J.~Yu, J.~Chen, and P.~Jaillet.
\newblock Parallel {Gaussian} process regression for big data: Low-rank
  representation meets {M}arkov approximation.
\newblock In {\em Proc. {AAAI}}, pages 2821--2827, 2015.

\bibitem{mantelero2013eu}
A.~Mantelero.
\newblock The {EU} proposal for a general data protection regulation and the
  roots of the `right to be forgotten'.
\newblock {\em Computer Law \& Security Review}, 29(3):229--235, 2013.

\bibitem{Ruofei18}
R.~Ouyang and B.~K.~H. Low.
\newblock Gaussian process decentralized data fusion meets transfer learning in
  large-scale distributed cooperative perception.
\newblock In {\em Proc. AAAI}, pages 3876--3883, 2018.

\bibitem{papamakarios2017masked}
G.~Papamakarios, T.~Pavlakou, and I.~Murray.
\newblock Masked autoregressive flow for density estimation.
\newblock In {\em Proc. {NeurIPS}}, pages 2338--2347, 2017.

\bibitem{quinonero2005unifying}
J.~{Qui{\~{n}}onero-Candela} and C.~E. Rasmussen.
\newblock A unifying view of sparse approximate {Gaussian} process regression.
\newblock {\em JMLR}, 6:1939--1959, 2005.

\bibitem{rasmussen06}
C.~E. Rasmussen and C.~K.~I. Williams.
\newblock {\em Gaussian Processes for Machine Learning}.
\newblock {MIT} Press, 2006.

\bibitem{rezende2015variational}
D.~J. Rezende and S.~Mohamed.
\newblock Variational inference with normalizing flows.
\newblock In {\em Proc. {ICML}}, pages 1530--1538, 2015.

\bibitem{schelteramnesia}
S.~Schelter.
\newblock ``{Amnesia}'' -- {Towards} machine learning models that can forget
  user data very fast.
\newblock In {\em Proc. International Workshop on Applied {AI} for Database
  Systems and Applications}, 2019.

\bibitem{teng20}
T.~Teng, J.~Chen, Y.~Zhang, and B.~K.~H. Low.
\newblock Scalable variational {Bayesian} kernel selection for sparse
  {Gaussian} process regression.
\newblock In {\em Proc. {AAAI}}, pages 5997--6004, 2020.

\bibitem{LowAAAI14}
N.~Xu, B.~K.~H. Low, J.~Chen, K.~K. Lim, and E.~B. {\"{O}zg\"{u}l}.
\newblock {GP-Localize}: Persistent mobile robot localization using online
  sparse {Gaussian} process observation model.
\newblock In {\em Proc. {AAAI}}, pages 2585--2592, 2014.

\bibitem{yu19}
H.~Yu, Y.~Chen, Z.~Dai, K.~H. Low, and P.~Jaillet.
\newblock Implicit posterior variational inference for deep {Gaussian}
  processes.
\newblock In {\em Proc. {NeurIPS}}, pages 14475--14486, 2019.

\bibitem{HaibinAPP}
H.~Yu, T.~N. Hoang, B.~K.~H. Low, and P.~Jaillet.
\newblock Stochastic variational inference for {Bayesian} sparse {Gaussian}
  process regression.
\newblock In {\em Proc. {IJCNN}}, 2019.

\end{thebibliography}

\clearpage
\appendix 
\section{Proof of Proposition~\ref{rmk:klmarginal}}
\label{app:klmarginal}
%
We first follow the proof of the log-sum inequality to prove the following inequality:
%
%
\begin{equation}
q_u(y|\dc)\ \log \frac{q_u(y|\dc)}{p(y|\dc)}
\le \int q_u(\bm{\theta}|\dc)\  p(y|\bm{\theta})\ \log \frac{q_u(\bm{\theta}|\dc)}{p(\bm{\theta}|\dc)}\ \text{d}\bm{\theta}
\label{eq:logsum2}
\end{equation}
where $q_u(y|\dc) \triangleq \mbb{E}_{q_u(\bm{\theta}|\dc)}[p(y|\bm{\theta})] = \int q_u(\bm{\theta}|\dc)\  p(y|\bm{\theta})\ \text{d}\bm{\theta}$ and $p(y|\dc) \triangleq \mbb{E}_{p(\bm{\theta}|\dc)}[p(y|\bm{\theta})] = \int p(\bm{\theta}|\dc)\  p(y|\bm{\theta})\ \text{d}\bm{\theta}$.
\begin{proof}
Define the function $f(t) \triangleq t\log t$ which is convex. Then, 
\begin{align*}
&\int q_u(\bm{\theta}|\dc)\  p(y|\bm{\theta})\ \log \frac{q_u(\bm{\theta}|\dc)}{p(\bm{\theta}|\dc)}\ \text{d}\bm{\theta}\\
&=\int p(\bm{\theta}|\dc) \ p(y|\bm{\theta})\  f\left(\frac{q_u(\bm{\theta}|\dc)}{p(\bm{\theta}|\dc)}\right)\ \text{d}\bm{\theta}\\
&=\mbb{E}_{p(\bm{\theta}|\dc)}[p(y|\bm{\theta})] 
    \int 
    \frac{p(\bm{\theta}|\dc) \ p(y|\bm{\theta})}{\mbb{E}_{p(\bm{\theta}|\dc)}[p(y|\bm{\theta})]}\ 
    f\left(\frac{q_u(\bm{\theta}|\dc)}{p(\bm{\theta}|\dc)}\right)\ \text{d}\bm{\theta}\\
&\ge \mbb{E}_{p(\bm{\theta}|\dc)}[p(y|\bm{\theta})] 
    \ f\left(
        \int 
        \frac{p(\bm{\theta}|\dc) \ p(y|\bm{\theta})}{\mbb{E}_{p(\bm{\theta}|\dc)}[p(y|\bm{\theta})]}
        \frac{q_u(\bm{\theta}|\dc)}{p(\bm{\theta}|\dc)}\ \text{d}\bm{\theta}
    \right)\\
&= \mbb{E}_{p(\bm{\theta}|\dc)}[p(y|\bm{\theta})] 
\ f\left(
        \int 
        \frac{p(y|\bm{\theta})\ q_u(\bm{\theta}|\dc)}{\mbb{E}_{p(\bm{\theta}|\dc)}[p(y|\bm{\theta})]}\ \text{d}\bm{\theta}
    \right)\\
&= \mbb{E}_{p(\bm{\theta}|\dc)}[p(y|\bm{\theta})] 
    \ f\left(
        \frac{ \mbb{E}_{q_u(\bm{\theta}|\dc)}[p(y|\bm{\theta})]}{\mbb{E}_{p(\bm{\theta}|\dc)}[p(y|\bm{\theta})]}
    \right)\\
&= \mbb{E}_{q_u(\bm{\theta}|\dc)}[p(y|\bm{\theta})]\ \log \frac{\mbb{E}_{q_u(\bm{\theta}|\dc)}[p(y|\bm{\theta})]}{\mbb{E}_{p(\bm{\theta}|\dc)}[p(y|\bm{\theta})]}\\
&= q_u(y|\dc)\ \log \frac{q_u(y|\dc)}{p(y|\dc)}
\end{align*}
where the inequality is due to Jensen's inequality.
\end{proof}

Then, integrating both sides of \eqref{eq:logsum2} w.r.t.~$y$, 
\begin{align*}
\int q_u(y|\dc)\ \log \frac{q_u(y|\dc)}{p(y|\dc)}\ \text{d}y
&\le 
\int \int q_u(\bm{\theta}|\dc) \ p(y|\bm{\theta})\ \log \frac{q_u(\bm{\theta}|\dc)}{p(\bm{\theta}|\dc)}\ \text{d}\bm{\theta}\ \text{d}y\\
\int q_u(y|\dc)\ \log \frac{q_u(y|\dc)}{p(y|\dc)}\ \text{d}y
&\le 
\int q_u(\bm{\theta}|\dc) \left(\int p(y|\bm{\theta})\ \text{d}y \right) \log \frac{q_u(\bm{\theta}|\dc)}{p(\bm{\theta}|\dc)}\ \text{d}\bm{\theta}\\
\int q_u(y|\dc)\ \log \frac{q_u(y|\dc)}{p(y|\dc)}\ \text{d}y
&\le 
\int q_u(\bm{\theta}|\dc) \log \frac{q_u(\bm{\theta}|\dc)}{p(\bm{\theta}|\dc)}\ \text{d}\bm{\theta}\\
\text{KL}[q_u(y|\dc)\ \Vert\ p(y|\dc)] &\le \text{KL}[q_u(\bm{\theta}|\dc)\ \Vert\ p(\bm{\theta}|\dc)]\ .
\end{align*}

\section{Proof of Proposition~\ref{theo:eubo}}
\label{app:eubo}
%
From \eqref{eq:suffdata}, 
\begin{align*}
\log p(\dr|\dc) &= \log \frac{p(\dr|\bm{\theta})\ p(\bm{\theta}|\dc)}{p(\bm{\theta}|\da)}\\
    &= \log \frac{q_u(\bm{\theta}|\dc)\ p(\dr|\bm{\theta})\  p(\bm{\theta}|\dc)}{q_u(\bm{\theta}|\dc)\  p(\bm{\theta}|\da)}\\
    &= \log p(\dr|\bm{\theta})
    + \log \frac{q_u(\bm{\theta}|\dc)}{p(\bm{\theta}|\da)}
    - \log \frac{q_u(\bm{\theta}|\dc)}{p(\bm{\theta}|\dc)}\ .
\end{align*}
Then, taking an expectation of both sides w.r.t.~$q_u(\bm{\theta}|\dc)$,
\begin{align*}
\log p(\dr|\dc) &=
    \int q_u(\bm{\theta}|\dc) \log p(\dr|\bm{\theta})\ \text{d}\bm{\theta}
    + \int q_u(\bm{\theta}|\dc) \log \frac{q_u(\bm{\theta}|\dc)}{p(\bm{\theta}|\da)}\ \text{d}\bm{\theta}
    - \int q_u(\bm{\theta}|\dc) \log \frac{q_u(\bm{\theta}|\dc)}{p(\bm{\theta}|\dc)}\ \text{d}\bm{\theta}\\
    &= \int q_u(\bm{\theta}|\dc)\  \log p(\dr|\bm{\theta})\ \text{d}\bm{\theta}
    + \text{KL}[q_u(\bm{\theta}|\dc)\ \Vert\ p(\bm{\theta}|\da)]
    - \text{KL}[q_u(\bm{\theta}|\dc)\ \Vert\ p(\bm{\theta}|\dc)]\\
    &= \mcl{U} - \text{KL}[q_u(\bm{\theta}|\dc)\ \Vert\ p(\bm{\theta}|\dc)]\ .
\end{align*}
Therefore,
\begin{align*}
\mcl{U} =  \log p(\dr|\dc) + \text{KL}[q_u(\bm{\theta}|\dc)\ \Vert\ p(\bm{\theta}|\dc)] \ge \log p(\dr|\dc)
\end{align*}
since $\text{KL}[q_u(\bm{\theta}|\dc)\ \Vert\ p(\bm{\theta}|\dc)] \ge 0$.
So, $\mcl{U}$ is an upper bound of $\log p(\dr|\dc)$.
\section{Bayesian Linear Regression}
\label{app:lr}
We perform unlearning of a simple Bayesian linear regression model: $y_x = ax^3 + bx^2 + cx + d + \epsilon$ where $a=2$, $b=-3$, $c=1$, and $d=0$ are the model parameters $\bm{\theta}$, and the noise is $\epsilon \sim \mcl{N}(0,0.05^2)$. 
Though the exact posterior belief of $\bm{\theta}$ is known to be a multivariate Gaussian, we choose to use a low-rank approximation (i.e.,  multivariate Gaussian with a diagonal covariance matrice) and represent the approximate posterior beliefs of the model parameters with independent Gaussians 
so that the approximation is not exact.

\begin{figure}
\centering
\begin{tabular}{ccc}
\includegraphics[height=0.2\textwidth]{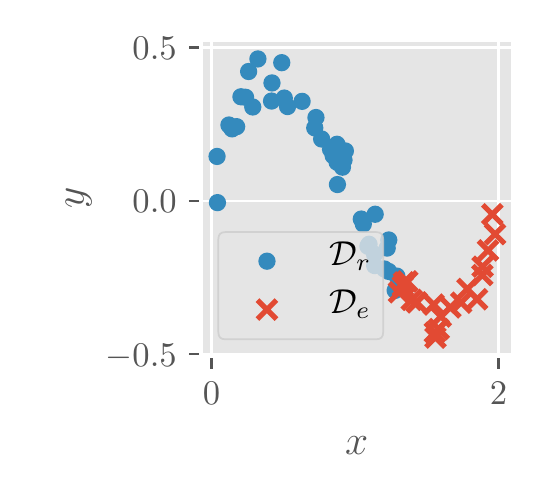}
&
\includegraphics[height=0.2\textwidth]{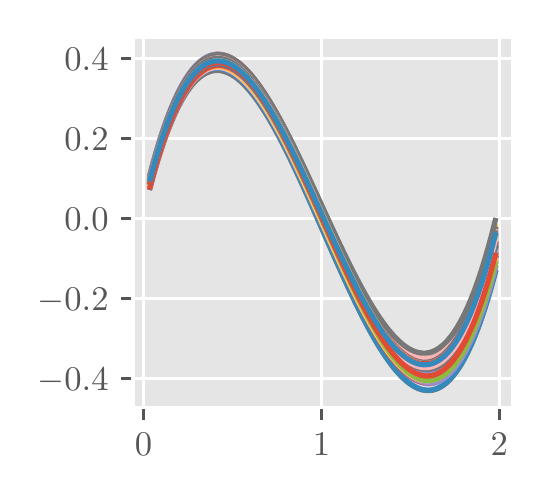}
&
\includegraphics[height=0.2\textwidth]{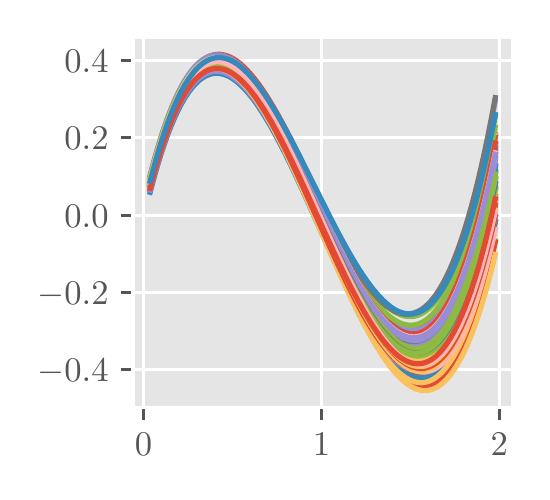}
\\
(a) Dataset
&
(b) Samples from $q(y_x| \da)$
&
(c) Samples from $q(y_x| \dc)$
\\
\includegraphics[height=0.2\textwidth]{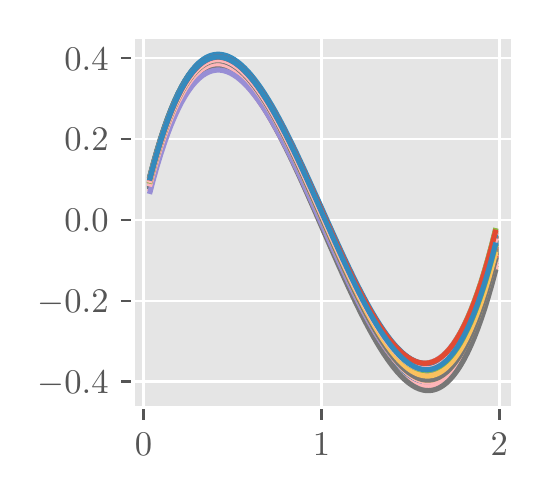}
&
\includegraphics[height=0.2\textwidth]{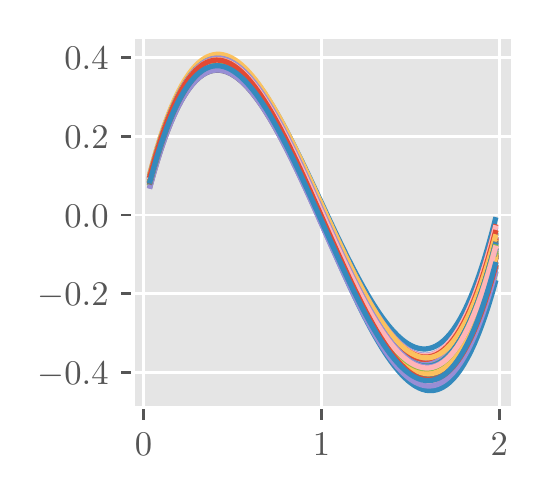}
&
\includegraphics[height=0.2\textwidth]{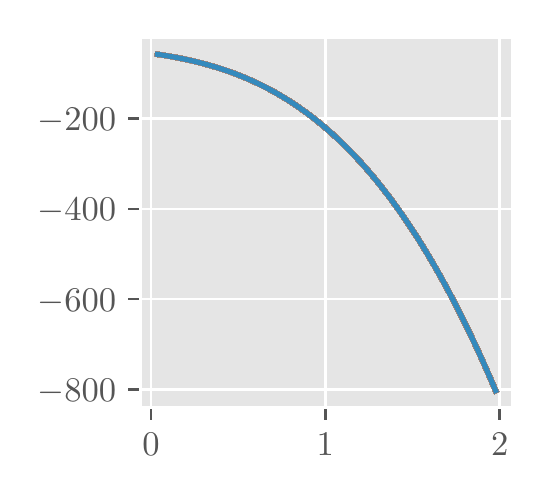}
\\
(d) EUBO with $\lambda = 0.5$
&
(e) EUBO with $\lambda = 0.1$
&
(f) EUBO with $\lambda = 0$
\\
\includegraphics[height=0.2\textwidth]{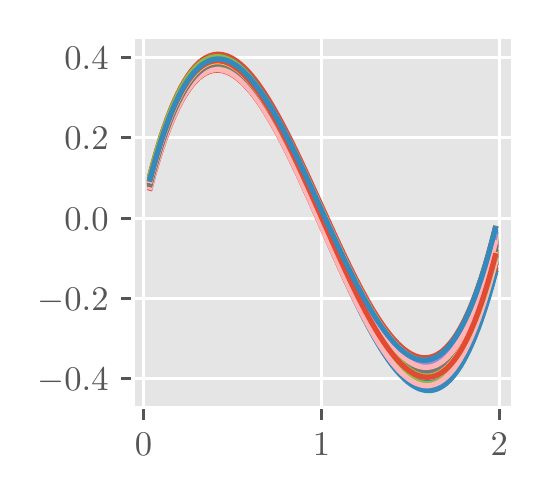}
&
\includegraphics[height=0.2\textwidth]{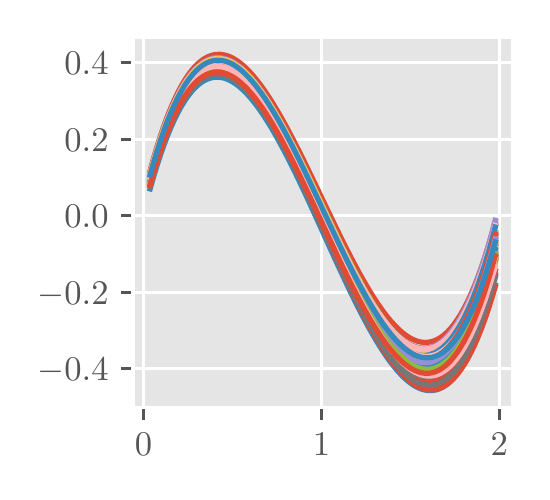}
&
\includegraphics[height=0.2\textwidth]{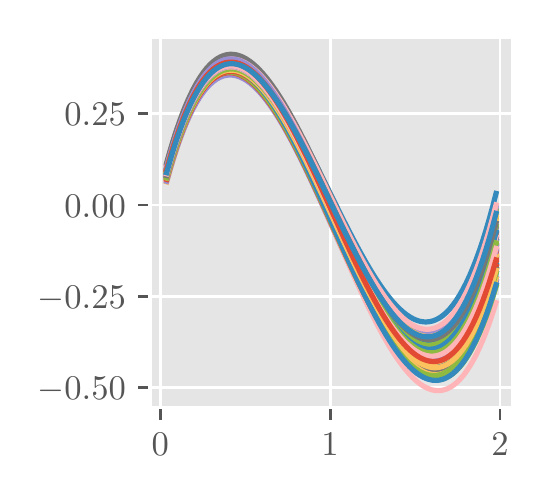}
\\
(g) rKL with $\lambda = 0.5$
&
(h) rKL with $\lambda = 0.1$
&
(i) rKL with $\lambda = 0$
\end{tabular}
\caption{Plots of (a) synthetic linear regression dataset with erased data $\dr$ (crosses) and remaining data $\dc$ (dots), and samples from predictive distributions obtained using VI from (b) training with full data $\da$ and (c) retraining with $\dc$. Plots of samples from predictive distributions (d-f) $\eubo(y_x|\dc)$ and (g-i) $\elbo(y_x|\dc)$ induced, respectively, by EUBO and rKL with varying $\lambda$.}
\label{fig:lrkl}
\end{figure}

Fig.~\ref{fig:lrkl}a shows the remaining data $\dc$ and erased data $\dr$. Note that the erased data $\dr$ is informative to the approximate posterior beliefs of the model parameters $\bm{\theta}$ as $\dr$ are clustered. So, the difference between the samples drawn from predictive distributions $q(y_x| \da)$ (Fig.~\ref{fig:lrkl}b) vs.~$q(y_x| \dc)$ (Fig.~\ref{fig:lrkl}c) is large.
\begin{table}[h]
\centering
\caption{KL divergences achieved by EUBO (left column) and rKL (right column) with varying $\lambda$ for synthetic linear regression dataset.}
\begin{tabular}{ccc}
\toprule
$\lambda$ & $\text{KL}[\eubo(\bm{\theta}|\dc;\lambda)\ \Vert\ q(\bm{\theta}|\dc)]$ & $\text{KL}[\elbo(\bm{\theta}|\dc;\lambda)\ \Vert\ q(\bm{\theta}|\dc)]$\\
\midrule
$0.5$ & $0.1143$ & $0.1012$\\
$0.1$ & $0.0899$ & $0.0600$\\
$0.0$ & $266.68$ & $0.0158$\\
\bottomrule
\end{tabular}
\label{tbl:lrkl}
\end{table}

From Table~\ref{tbl:lrkl}, the KL divergences achieved by EUBO and rKL with $\lambda =0.1, 0.5$ are smaller than $\text{KL}[q(\bm{\theta}|\da)\ \Vert\ q(\bm{\theta}|\dc)]$ of value $0.1170$ (i.e., baseline representing no unlearning), hence demonstrating reasonable unlearning performance.
When $\lambda = 0$, EUBO suffers from catastrophic unlearning, but rKL does not.
The KL divergences in Table~\ref{tbl:lrkl} also agree with the plots of samples drawn from the predictive distributions induced by EUBO and rKL in Fig.~\ref{fig:lrkl} by comparing with the samples drawn from the predictive distribution obtained using VI from retraining with $\dc$ in Fig.~\ref{fig:lrkl}c.
\section{Bimodal Posterior Belief}
\label{app:multimode}
Let the posterior belief of model parameter $\theta$ given full data $\da$ be a Gaussian mixture (i.e., a bimodal distribution): 
\begin{equation}
p(\theta|\da) \triangleq 0.5\ \phi(\theta; 0,1) + 0.5\ \phi(\theta; 2,1)
\label{eq:mmfullpost}
\end{equation}
where $\phi(\theta;\mu,\sigma^2)$ is a Gaussian p.d.f.~with mean $\mu$ and variance $\sigma^2$.
We deliberately choose the likelihood of the erased data $\dr$ to be
\begin{equation}
p(\dr|\theta) \triangleq 1 + \frac{\phi(\theta;2,1)}{\phi(\theta;0,1)}
\label{eq:mmerasedll}
\end{equation}
so that the posterior belief of $\theta$ given the remaining data $\dc$ is a Gaussian:
\begin{equation}
p(\theta|\dc) \propto \frac{p(\theta|\da)}{p(\dr|\theta)}
	= \phi(\theta; 0,1)
\label{eq:mmremainpost}
\end{equation}
where the proportionality is due to~\eqref{eq:suffdata}.

We assume to only have access to the likelihood of the erased data in~\eqref{eq:mmerasedll}; the exact posterior beliefs of $\theta$ given the full data~\eqref{eq:mmfullpost} and that given the remaining data \eqref{eq:mmremainpost} are not available. Instead, we have access to an approximate posterior belief $q(\theta|\da)$ given the full data obtained using VI by minimizing $\text{KL}[q(\theta|\da)\ \Vert\ p(\theta|\da)]$ or, equivalently, maximizing the ELBO (Section~\ref{sec:vi}):
\begin{equation}
q(\theta|\da) = \phi(\theta; 1.004, 1.390^2)\ .
\label{eq:mmfullpostvi}
\end{equation}
Given the likelihood $p(\dr|\theta)$ of the erased data in~\eqref{eq:mmerasedll} and the approximate posterior belief $q(\theta|\da)$ given the full data \eqref{eq:mmfullpostvi}, unlearning 
from $\dr$ is performed using EUBO and rKL to obtain
$$
\eubo(\theta| \dc; \lambda = 0) = \phi(\theta; 0.060, 1.000^2)\quad\text{and}\quad\elbo(\theta| \dc; \lambda = 0) = \phi(\theta; 0.062, 1.018^2)\ ,
$$
respectively. Hence, both EUBO and rKL perform reasonably well since their respective $\eubo(\theta| \dc; \lambda = 0)$ and $\elbo(\theta| \dc; \lambda = 0)$ are close to $p(\theta|\dc) = \phi(\theta;0,1)$ \eqref{eq:mmremainpost} when $p(\theta|\da)$ is a bimodal distribution.
\section{Gaussian Process (GP) Classification with Synthetic Moon Dataset: Additional Details and Experimental Results}
\label{app:moon}
This section discusses the sparse GP model that is used in the classification of the synthetic moon dataset in Sec.~\ref{subsec:expmoon}.
Let $y_{\mbf{x}} \in \{0,1\}$ be the class label of $\mbf{x} \in \mcl{X} \subset \mbb{R}^2$; $y_\mbf{x} = 1$ denotes the `blue' class plotted as blue dots in Fig.~\ref{fig:moon}a. The probability of $y_{\mbf{x}}$ is defined as follows:
\begin{equation}
\begin{aligned}
p(y_{\mbf{x}} = 1|f_{\mbf{x}}) & \triangleq \frac{1}{1 + \exp(f_{\mbf{x}})}\\
p(y_{\mbf{x}} = 0|f_{\mbf{x}}) & \triangleq  \frac{\exp(f_{\mbf{x}})}{1 + \exp(f_{\mbf{x}})}
\end{aligned}
\label{eq:moonll}
\end{equation}
where $f_{\mbf{x}}$ is modeled using a GP~\cite{rasmussen06}, that is, every finite subset of $\{f_{\mbf{x}}\}_{\mbf{x} \in \mcl{X}}$ follows a multivariate Gaussian distribution. A GP is fully specified by its \emph{prior} mean (i.e., assumed to be $0$ w.l.o.g.) and covariance $k_{\mbf{x}\mbf{x}'} \triangleq \text{cov}(\mbf{x}, \mbf{x}')$, the latter of which can be defined by the widely-used squared exponential covariance function $k_{\mbf{x}\mbf{x}'} \triangleq \sigma_f^2 \exp(-0.5 \Vert \Lambda(\mbf{x} - \mbf{x}')\Vert_2^2)$ where $\Lambda = \text{diag}[\lambda_1, \lambda_2]$ and $\sigma_f^2$ are the length-scale and signal variance hyperparameters, respectively. In this experiment, we set $\lambda_1 = 1.56$, $\lambda_2 = 1.35$, and $\sigma_f^2 = 4.74$.

We employ a sparse GP model, namely, the \emph{deterministic training conditional} (DTC) \cite{quinonero2005unifying} approximation of the GP model with a set $\mcl{X}_u$ of $20$ \emph{inducing inputs}. These inducing inputs are randomly selected from $\mcl{X}$ and remain the same (and fixed) for both model training and unlearning. Given the latent function values (i.e., also known as \emph{inducing variables}) $\mbf{f}_{\mcl{X}_u} \triangleq (f_{\mbf{x}})^{\top}_{\mbf{x} \in \mcl{X}_u}$ at these inducing inputs, the posterior belief of the latent function value $f_{\mbf{x}}$ at a new input $\mbf{x}$ is a Gaussian $p(f_{\mbf{x}}|\mbf{f}_{\mcl{X}_u}) = \mcl{N}(\mbf{k}_{\mbf{x} \mcl{X}_u} \mbf{K}_{\mcl{X}_u\mcl{X}_u}^{-1} \mbf{f}_{\mcl{X}_u}, k_{\mbf{x} \mbf{x}} - \mbf{k}_{\mbf{x} \mcl{X}_u} \mbf{K}_{\mcl{X}_u\mcl{X}_u}^{-1} \mbf{k}_{\mcl{X}_u \mbf{x}})$
%
%
where $\mbf{k}_{\mbf{x} \mcl{X}_u} \triangleq (k_{\mbf{x}\mbf{x}'})_{\mbf{x}' \in \mcl{X}_u}$, $\mbf{k}_{\mcl{X}_u \mbf{x}} = \mbf{k}_{\mbf{x} \mcl{X}_u}^\top$, and $\mbf{K}_{\mcl{X}_u\mcl{X}_u} = (k_{\mbf{x} \mbf{x}'})_{\mbf{x}, \mbf{x}' \in \mcl{X}_u}$.

Using $p(f_{\mbf{x}}|\mbf{f}_{\mcl{X}_u})$ and $q(\mbf{f}_{\mcl{X}_u}| \da) \triangleq \mcl{N}(\bm{\mu}_{\mcl{X}_u}, \bm{\Sigma}_{\mcl{X}_u})$, it can be derived that 
the approximate posterior belief $q(f_{\mbf{x}}| \da)$ of $f_{\mbf{x}}$ given full data $\da$
is also a Gaussian with the following respective \emph{posterior} mean and variance:
\begin{align}
\mu_{\mbf{x}|\da} &\triangleq \mbf{k}_{\mbf{x} \mcl{X}_u} \mbf{K}_{\mcl{X}_u\mcl{X}_u}^{-1} \bm{\mu}_{\mcl{X}_u}\ ,\label{eq:gppostmean}\\
\sigma_{\mbf{x}|\da}^2 &\triangleq k_{\mbf{x} \mbf{x}} - \mbf{k}_{\mbf{x} \mcl{X}_u} \mbf{K}_{\mcl{X}_u\mcl{X}_u}^{-1} \mbf{k}_{\mcl{X}_u \mbf{x}} + \mbf{k}_{\mbf{x} \mcl{X}_u} \mbf{K}_{\mcl{X}_u\mcl{X}_u}^{-1} \bm{\Sigma}_{\mcl{X}_u} \mbf{K}_{\mcl{X}_u\mcl{X}_u}^{-1} \mbf{k}_{\mcl{X}_u \mbf{x}}\ .\label{eq:gppostvar}
\end{align}
The approximate posterior belief $q(f_{\mbf{x}}| \dc)$ of $f_{\mbf{x}}$ from retraining with remaining data $\dc$  using VI (specifically, using $q(\mbf{f}_{\mcl{X}_u}| \dc)$) can be derived in the same way as that of $q(f_{\mbf{x}}| \da)$.

The parameters $\bm{\mu}_{\mcl{X}_u}$, $\bm{\Sigma}_{\mcl{X}_u}$ of the approximate posterior belief $q(\mbf{f}_{\mcl{X}_u}| \da)$ is optimized by maximizing the ELBO with stochastic gradient ascent (let $\bm{\theta} = \mbf{f}_{\mcl{X}_u}$ in~\eqref{eq:elbo} in Sec.~\ref{sec:vi}):
\begin{equation*}
\mbb{E}_{\mbf{f}_{\mcl{X}_u} \sim q(\mbf{f}_{\mcl{X}_u}| \da)} \left[ \log p(\da|\mbf{f}_{\mcl{X}_u})
- \log q(\mbf{f}_{\mcl{X}_u}| \da)
+ \log p(\mbf{f}_{\mcl{X}_u}) \right]
\end{equation*}
where $p(\da|\mbf{f}_{\mcl{X}_u})$ is computed using~\eqref{eq:moonll},~\eqref{eq:gppostmean} and~\eqref{eq:gppostvar}.

\begin{figure}
\begin{tabular}{cccc}
\includegraphics[trim={7mm 8mm 3mm 3mm}, clip,height=0.18\textwidth]{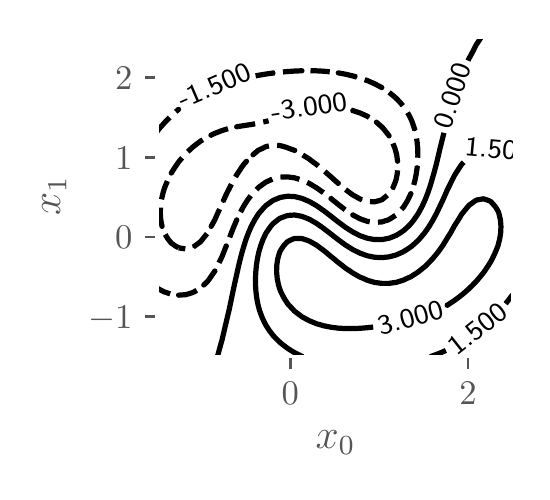}
&
\includegraphics[trim={7mm 8mm 3mm 3mm}, clip,height=0.18\textwidth]{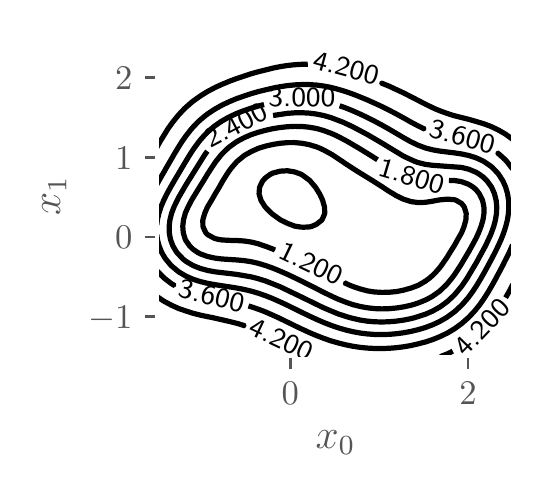}
&
\includegraphics[trim={7mm 8mm 3mm 3mm}, clip,height=0.18\textwidth]{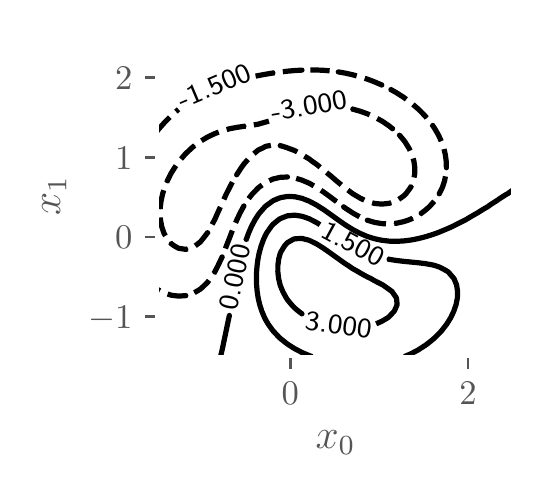}
&
\includegraphics[trim={7mm 8mm 3mm 3mm}, clip,height=0.18\textwidth]{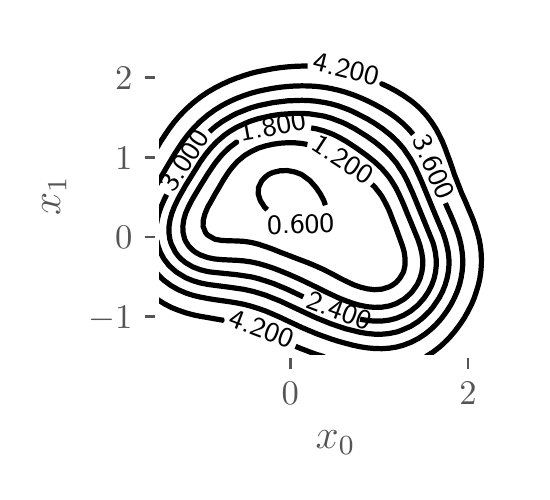}
\\
(a) $\mu_{\mbf{x}|\da}$ 
&
(b) $\sigma^2_{\mbf{x}|\da}$ 
&
(c) $\mu_{\mbf{x}|\dc}$ 
&
(d) $\sigma^2_{\mbf{x}|\dc}$ 
\end{tabular}
\caption{Plots of approximate posterior beliefs (a-b) $q(f_{\mbf{x}}|\da)$ and (c-d) $q(f_{\mbf{x}}|\dc)$.
}
\label{fig:moonfullpost}
\end{figure}

\begin{figure}
\centering
\begin{tabular}{cc}
\includegraphics[trim={7mm 8mm 3mm 3mm}, clip,height=0.18\textwidth]{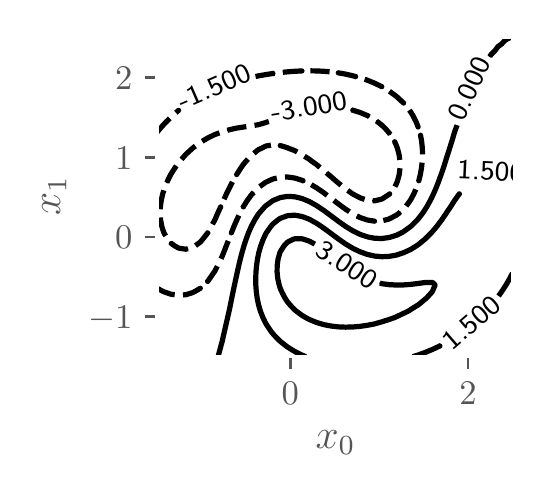}
&
\includegraphics[trim={7mm 8mm 3mm 3mm}, clip,height=0.18\textwidth]{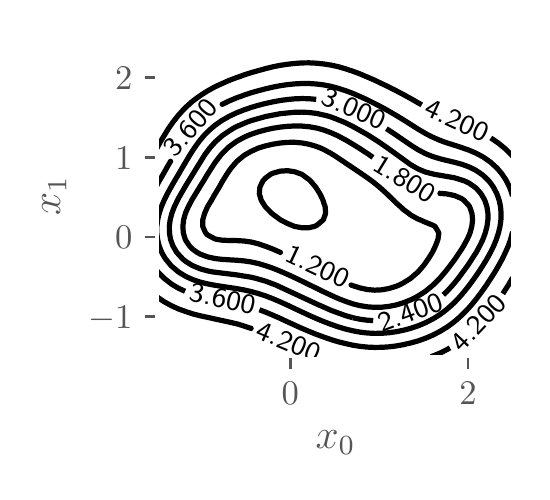}\\
(a) Mean of $\eubo(f_{\mbf{x}}|\dc; \lambda=10^{-5})$
&
(b) Variance of $\eubo(f_{\mbf{x}}|\dc; \lambda=10^{-5})$\\
\includegraphics[trim={7mm 8mm 3mm 3mm}, clip,height=0.18\textwidth]{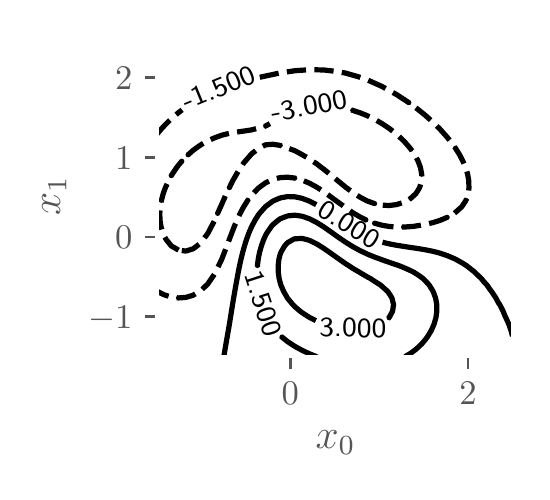}
&
\includegraphics[trim={7mm 8mm 3mm 3mm}, clip,height=0.18\textwidth]{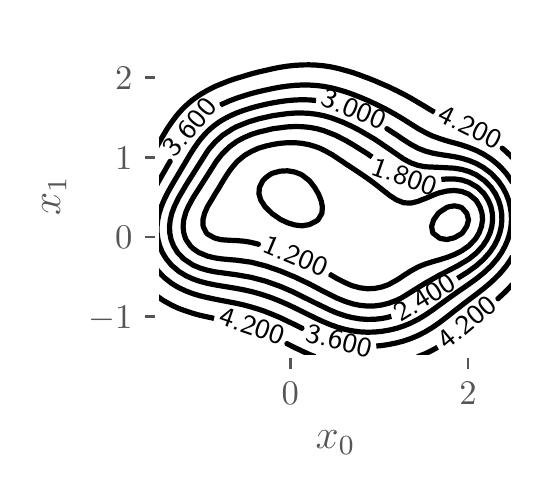}\\
(c) Mean of $\eubo(f_{\mbf{x}}|\dc; \lambda=10^{-9})$
&
(d) Variance of $\eubo(f_{\mbf{x}}|\dc; \lambda=10^{-9})$\\
\includegraphics[trim={7mm 8mm 3mm 3mm}, clip,height=0.18\textwidth]{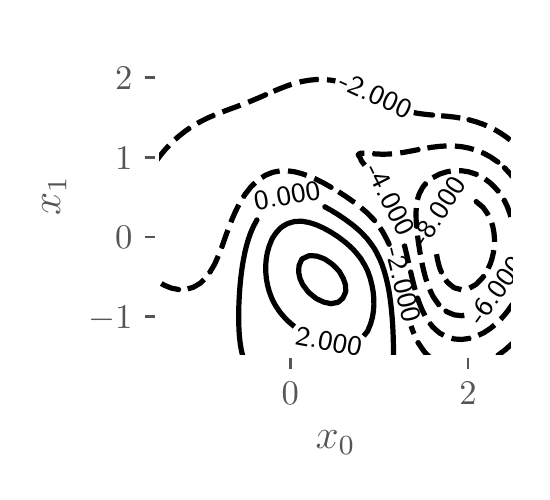}
&
\includegraphics[trim={7mm 8mm 3mm 3mm}, clip,height=0.18\textwidth]{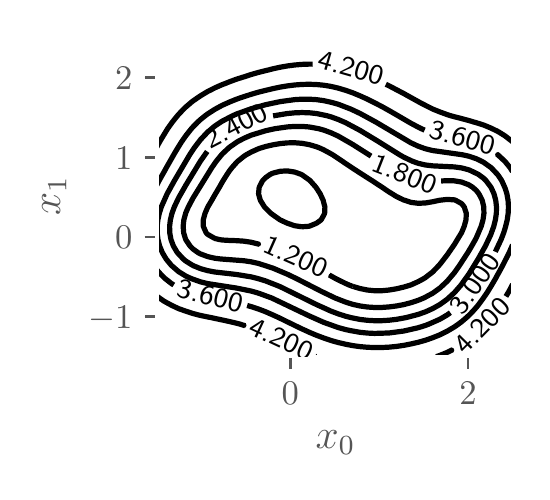}
\\
(e) Mean of $\eubo(f_{\mbf{x}}|\dc; \lambda=0)$
&
(f) Variance of $\eubo(f_{\mbf{x}}|\dc; \lambda=0)$
\end{tabular}
\caption{Plots of approximate posterior belief $\eubo(f_{\mbf{x}}|\dc;\lambda)$ induced by EUBO for varying $\lambda$.}
\label{fig:moonfullposteubo}
\end{figure}

\begin{figure}
\centering
\begin{tabular}{cc}
\includegraphics[trim={7mm 8mm 3mm 3mm}, clip,height=0.18\textwidth]{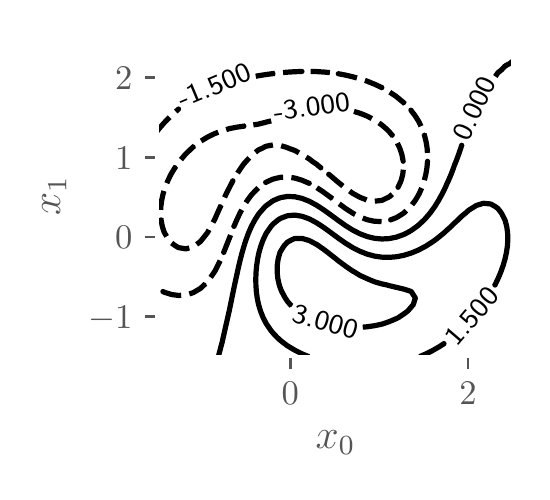}
&
\includegraphics[trim={7mm 8mm 3mm 3mm}, clip,height=0.18\textwidth]{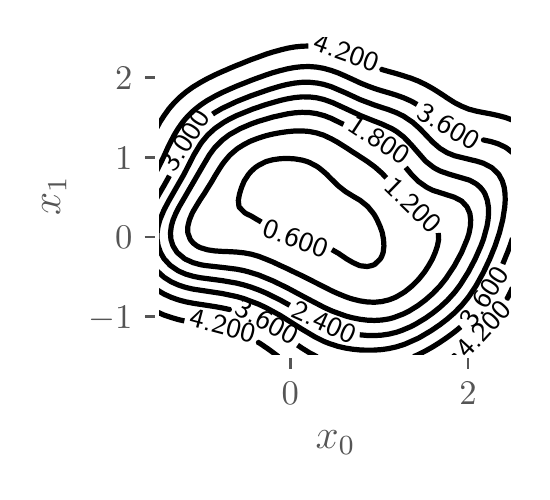}\\
(a) Mean of $\elbo(f_{\mbf{x}}|\dc; \lambda=10^{-5})$
&
(b) Variance of $\elbo(f_{\mbf{x}}|\dc; \lambda=10^{-5})$\\
\includegraphics[trim={7mm 8mm 3mm 3mm}, clip,height=0.18\textwidth]{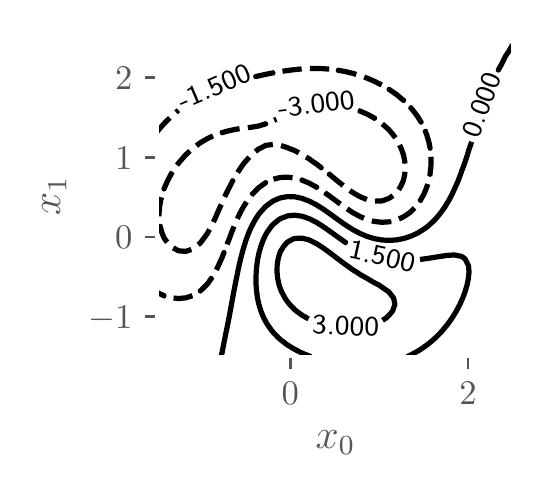}
&
\includegraphics[trim={7mm 8mm 3mm 3mm}, clip,height=0.18\textwidth]{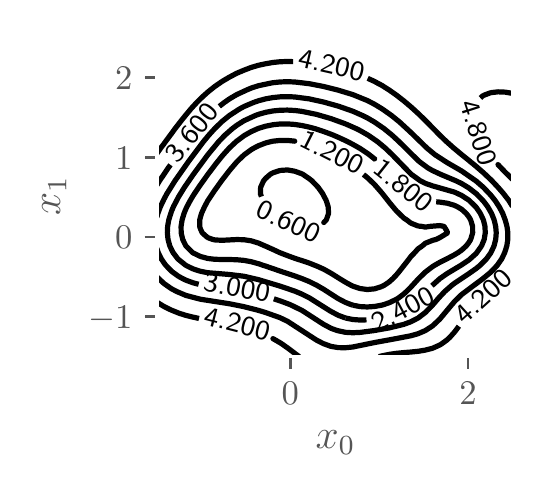}\\
(c) Mean of $\elbo(f_{\mbf{x}}|\dc; \lambda=10^{-9})$
&
(d) Variance of $\elbo(f_{\mbf{x}}|\dc; \lambda=10^{-9})$\\
\includegraphics[trim={7mm 8mm 3mm 3mm}, clip,height=0.18\textwidth]{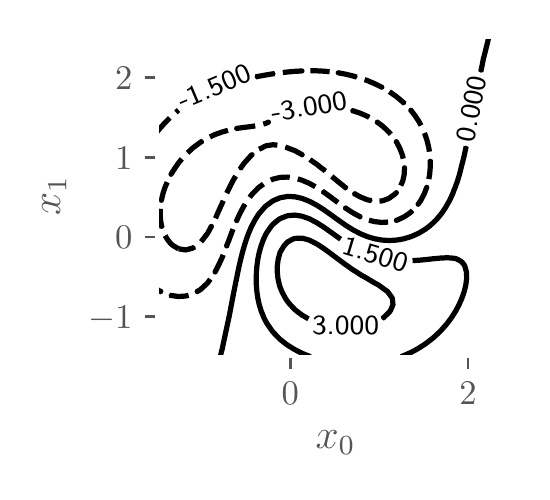}
&
\includegraphics[trim={7mm 8mm 3mm 3mm}, clip,height=0.18\textwidth]{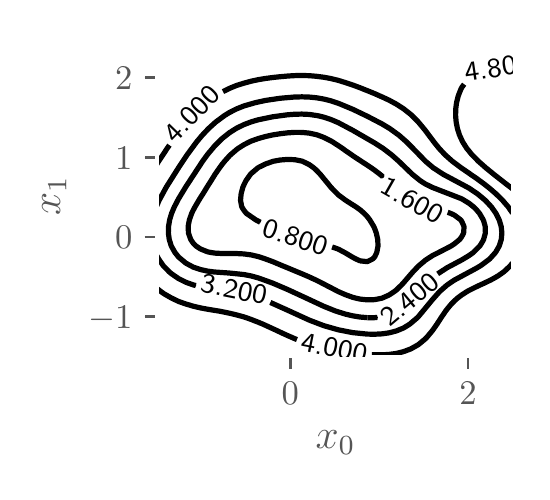}\\
(e) Mean of $\elbo(f_{\mbf{x}}|\dc; \lambda=0)$
&
(f) Variance of $\elbo(f_{\mbf{x}}|\dc; \lambda=0)$
\end{tabular}
\caption{Plots of approximate posterior belief  $\elbo(f_{\mbf{x}}|\dc;\lambda)$
induced by 
rKL for varying $\lambda$.}
\label{fig:moonfullpostelbo}
\end{figure}

Fig.~\ref{fig:moonfullpost} visualizes 
$q(f_{\mbf{x}}| \da)$ (Figs.~\ref{fig:moonfullpost}a and~\ref{fig:moonfullpost}b)
and $q(f_{\mbf{x}}| \dc)$ (Figs.~\ref{fig:moonfullpost}c and~\ref{fig:moonfullpost}d) whose corresponding predictive distributions 
$q(y_{\mbf{x}}=1| \da)$ and
$q(y_{\mbf{x}}=1| \dc)$ are shown in Figs.~\ref{fig:moon}b and~\ref{fig:moon}c, respectively.
On the other hand, Figs.~\ref{fig:moonfullposteubo} and~\ref{fig:moonfullpostelbo} visualize the approximate posterior beliefs $\eubo(f_{\mbf{x}}|\dc;\lambda)$ and $\elbo(f_{\mbf{x}}|\dc;\lambda)$
induced, respectively, by EUBO 
and rKL 
whose corresponding predictive distributions $\eubo(y_{\mbf{x}}=1|\dc)$ and $\elbo(y_{\mbf{x}}=1|\dc)$ are shown in Figs.~\ref{fig:moon}f-k.
Similar to the comparison between predictive distributions 
$\eubo(y_{\mbf{x}}=1|\dc)$ vs.~$q(y_{\mbf{x}}=1|\dc)$
in Sec.~\ref{subsec:expmoon},
it can be observed that the
approximate posterior belief $\eubo(f_{\mbf{x}}|\dc;\lambda= 10^{-9})$ induced by EUBO 
is similar to $q(f_{\mbf{x}}|\dc)$ obtained using VI from retraining with $\dc$ (compare Figs.~\ref{fig:moonfullposteubo}c vs.~\ref{fig:moonfullpost}c and Figs.~\ref{fig:moonfullposteubo}d vs.~\ref{fig:moonfullpost}d).
However, 
$\eubo(f_{\mbf{x}}|\dc;\lambda= 0)$ induced by EUBO differs
from $q(f_{\mbf{x}}|\dc)$ obtained using VI from retraining with $\dc$ (compare Figs.~\ref{fig:moonfullposteubo}e vs.~\ref{fig:moonfullpost}c and Figs.~\ref{fig:moonfullposteubo}f vs.~\ref{fig:moonfullpost}d). 
On the other hand, 
both the
approximate posterior beliefs $\elbo(f_{\mbf{x}}|\dc;\lambda= 10^{-9})$ and $\elbo(f_{\mbf{x}}|\dc;\lambda= 0)$ induced by rKL are similar to $q(f_{\mbf{x}}|\dc)$ 
obtained using VI from retraining with $\dc$ 
(compare Fig.~\ref{fig:moonfullpostelbo} vs.  Figs.~\ref{fig:moonfullpost}c-d).
\section{A Note on Erasing Informative Data}
\label{app:information}
In this section, we study the performance of our unlearning methods when  erasing a large quantity of data or with different distributions of erased data (i.e., erasing the data randomly vs.~deliberately erasing all data in a given class).
Let us consider the experiment in Sec.~\ref{subsec:expmoon} on the sparse GP model (i.e., the model parameters $\bm{\theta}$ in~\eqref{eq:elbo} in Sec.~\ref{sec:vi} are inducing variables $\mbf{f}_{\mcl{X}_u}$) in the classification of the synthetic moon dataset as it allows us to easily visualize both the approximate posterior beliefs of the latent function  $f_{\mbf{x}}$ and the predictive distributions of the output/observation $y_{\mbf{x}}$. 
A key factor influencing the performance of our unlearning methods in the above-mentioned scenarios is the difference between the approximate posterior belief of model parameters $\mbf{f}_{\mcl{X}_u}$ given remaining data $\dc$ vs.~that given full data $\da$. We quantify such a difference by how much the erased data $\dr$ reduces the entropy of model parameters/inducing variables $\mbf{f}_{\mcl{X}_u}$ given remaining data $\dc$:
%
\begin{equation}
    \mcl{I} \triangleq H(\mbf{f}_{\mcl{X}_u}| \dc) - H(\mbf{f}_{\mcl{X}_u}| \da) 
    = 
    - \int q(\mbf{f}_{\mcl{X}_u}| \dc) \log q(\mbf{f}_{\mcl{X}_u}| \dc)\ \text{d}\mbf{f}_{\mcl{X}_u}
    + \int q(\mbf{f}_{\mcl{X}_u}| \da) \log q(\mbf{f}_{\mcl{X}_u}| \da)\ \text{d}\mbf{f}_{\mcl{X}_u}\ .
    \label{eq:infomeasure}
\end{equation}
Note that $\mcl{I}$~\eqref{eq:infomeasure} is not the same as the mutual information (i.e.,  information gain) between $\mbf{f}_{\mcl{X}_u}$ and $\mbf{y}_{\dr}\triangleq (y_\mbf{x})^\top_{(\mbf{x},y_\mbf{x})\in \dr}$ 
given $\dc$, which is equal to $ H(\mbf{f}_{\mcl{X}_u}| \dc) - \mbb{E}_{p(\mbf{y}_{\dr}| \dc)} \left[ H(\mbf{f}_{\mcl{X}_u}| \dc, \mbf{y}_{\dr}) \right]$  
with an expensive-to-evaluate 
expectation term.
Furthermore, the outputs/observations $\mbf{y}_{\dr}$ are known from $\dr$. 
These therefore prompt us to choose  $\mcl{I}$~\eqref{eq:infomeasure} as the measure of how much the erased data $\dr$ reduces the entropy of model parameters/inducing variables $\mbf{f}_{\mcl{X}_u}$ given remaining data $\dc$. 

\begin{figure}
\centering
\begin{tabular}{@{}c@{}c@{}c@{}c@{}}
$\mcl{I} = 0.27$
&
\includegraphics[trim={0mm 0mm 3mm 3mm}, clip,height=0.22\textwidth]{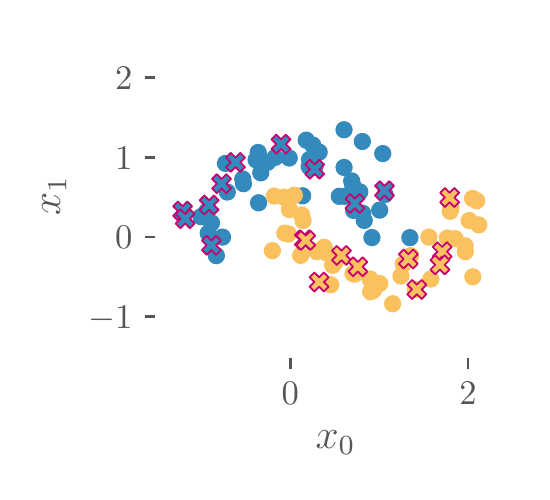}
&
\includegraphics[trim={3mm 0mm 3mm 3mm}, clip,height=0.22\textwidth]{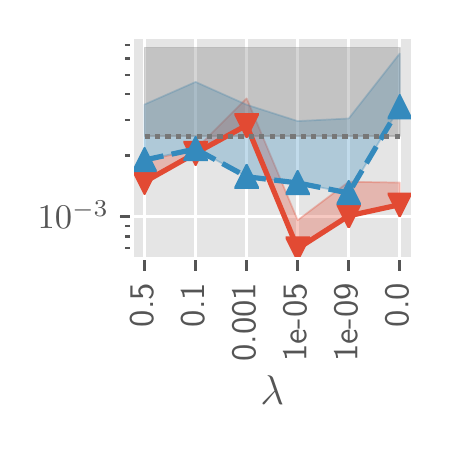}
&
\includegraphics[trim={3mm 0mm 3mm 3mm}, clip,height=0.22\textwidth]{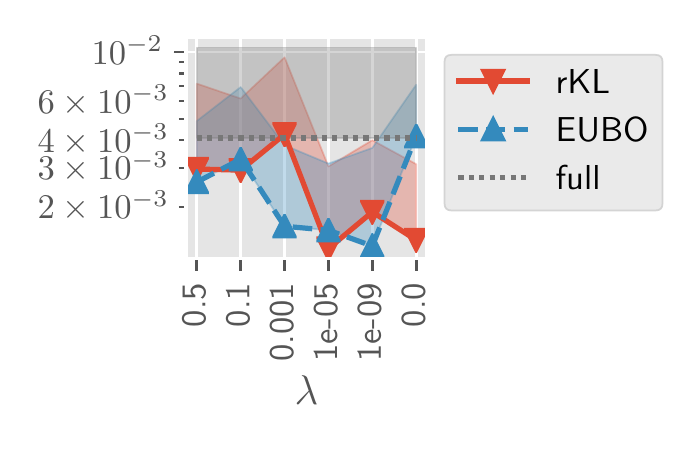}
\\
&
(a) Randomly selected $\dr$
&
(b) $\dc$
&
(c) $\dr$
\\
\\
$\mcl{I} = 1.59$
&
\includegraphics[trim={0mm 0mm 3mm 3mm}, clip,height=0.22\textwidth]{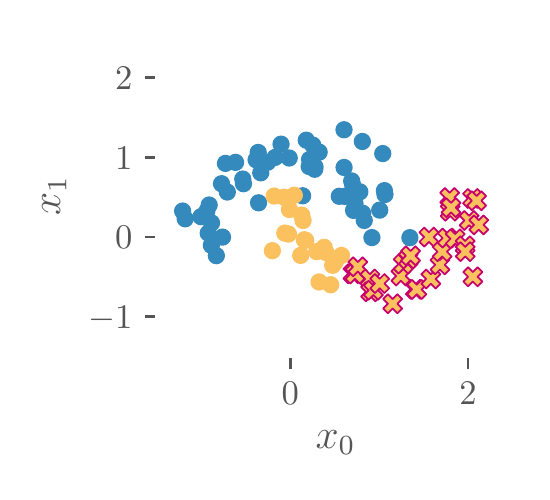}
&
\includegraphics[trim={3mm 0mm 3mm 3mm}, clip,height=0.22\textwidth]{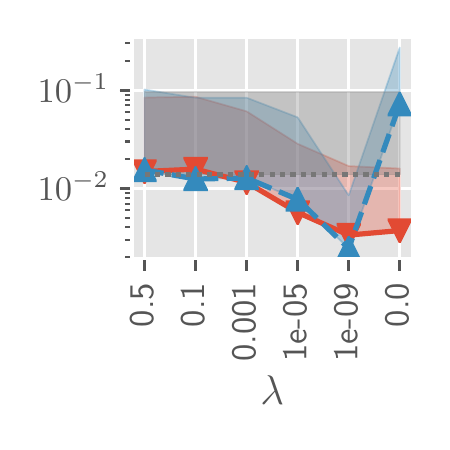}
&
\includegraphics[trim={3mm 0mm 3mm 3mm}, clip,height=0.22\textwidth]{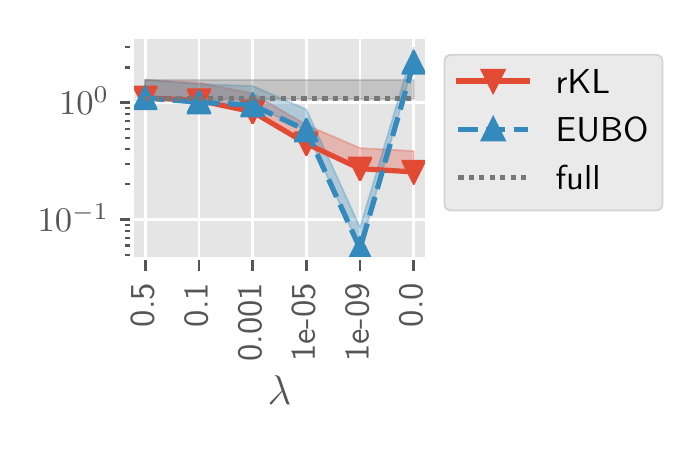}
\\
&
(d) Partially `yellow' $\dr$
&
(e) $\dc$
&
(f) $\dr$
\\
\\
$\mcl{I} = 2.06$
&
\includegraphics[trim={0mm 0mm 3mm 3mm}, clip,height=0.22\textwidth]{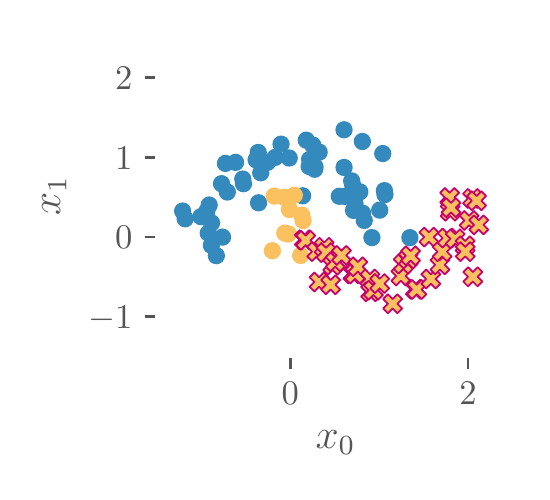}
&
\includegraphics[trim={3mm 0mm 3mm 3mm}, clip,height=0.22\textwidth]{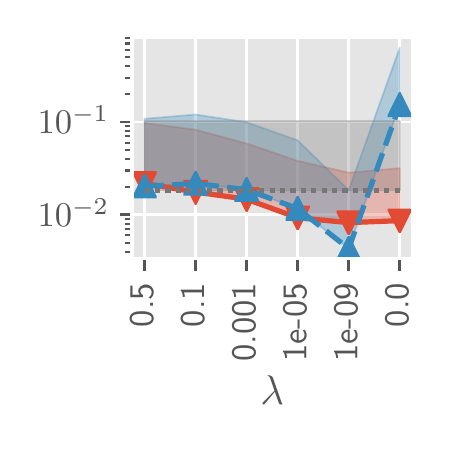}
&
\includegraphics[trim={3mm 0mm 3mm 3mm}, clip,height=0.22\textwidth]{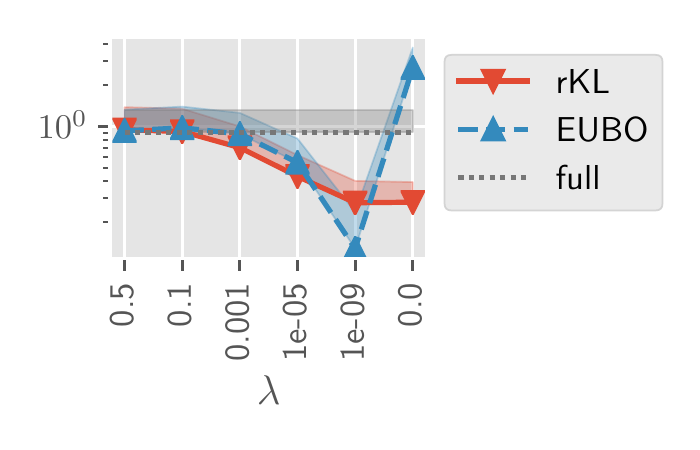}
\\
&
(g) Largely  `yellow' $\dr$
&
(h) $\dc$
&
(i) $\dr$
\\
\\
$\mcl{I} = 3.86$
&
\includegraphics[trim={0mm 0mm 3mm 3mm}, clip,height=0.22\textwidth]{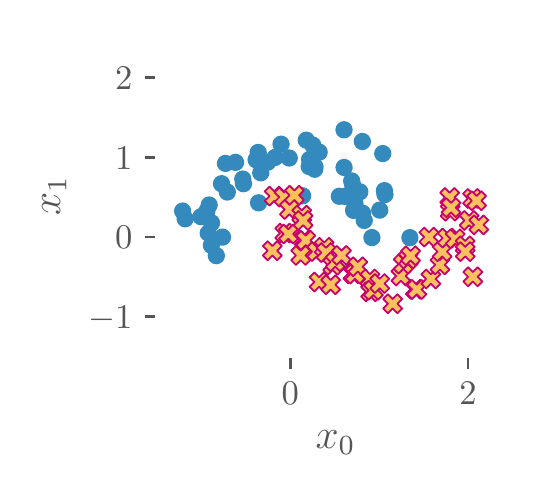}
&
\includegraphics[trim={3mm 0mm 3mm 3mm}, clip,height=0.22\textwidth]{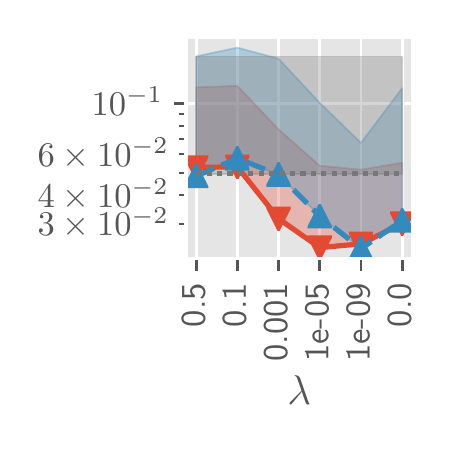}
&
\includegraphics[trim={3mm 0mm 3mm 3mm}, clip,height=0.22\textwidth]{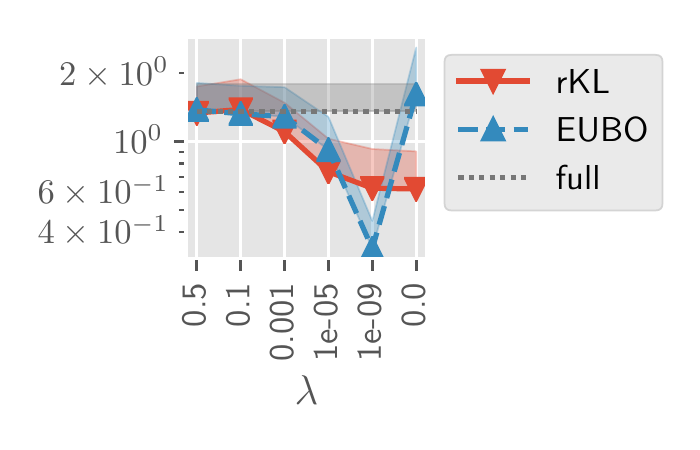}
\\
&
(j) Fully `yellow' $\dr$
&
(k) $\dc$
&
(l) $\dr$
\end{tabular}
\caption{Plots of (a,d,g,j) synthetic moon dataset with erased data $\dr$ (crosses) and remaining data $\dc$ (dots) in $4$ different scenarios. Graphs of averaged KL divergence vs.~$\lambda$ achieved by EUBO, \emph{reverse KL} (rKL), and $q(\bm{\theta}|\da)$ (i.e., baseline labeled as \emph{full}) over $\dc$ and $\dr$ in the following $4$ scenarios: (b-c) randomly selected $\dr$, (e-f) partially  `yellow' $\dr$, (h-i) largely `yellow' $\dr$, and (k-l) fully `yellow' $\dr$.}
\label{fig:moon4casemae}
\end{figure}

We investigate $4$ different scenarios in the order of increasing $\mcl{I}$:
\begin{enumerate}
    \item Randomly selected $\dr$ ($\mcl{I} = 0.27$): The erased data of size $|\dr|=20$ are randomly selected from $\da$. Hence, they are not necessarily near the decision boundary, i.e., $\dr$ does not reduce the entropy of  model parameters/inducing variables $\mbf{f}_{\mcl{X}_u}$ given $\dc$ much;
    \item Partially `yellow' $\dr$ ($\mcl{I} = 1.59$): The erased data of size $|\dr|=30$ are labeled with the `yellow' class and comprise inputs $\mbf{x}$ with the largest possible first component $x_0$. Such a choice ensures that the erased data group together to cover a part of the decision boundary, as shown in Fig.~\ref{fig:moon4casemae}d;
    \item Largely `yellow' $\dr$ ($\mcl{I} = 2.06$): The erased data of size $|\dr|=40$ are labeled with the yellow class and comprise inputs $\mbf{x}$ with the largest possible first component $x_0$. As the quantity of the erased data $\dr$ increases from $30$ (i.e., partially `yellow' $\dr$) to $40$, $\dr$ covers a larger part of the decision boundary (compare Figs.~\ref{fig:moon4casemae}g vs.~\ref{fig:moon4casemae}d); and
    \item Fully `yellow' $\dr$ ($\mcl{I} = 3.86$): The erased data of size $|\dr| = 50$ comprise all data in the yellow class. In this case, $\dr$ reduces the entropy of the model parameters/inducing variables $\mbf{f}_{\mcl{X}_u}$ given $\dc$ the most when compared to the above $3$ scenarios.
\end{enumerate}
As $\mcl{I}$ increases, the difference between the approximate posterior belief of $\mbf{f}_{\mcl{X}_u}$ given remaining data $\dc$ vs.~that given full data $\da$ increases. Though it is difficult to visualize such a difference directly, Proposition~\ref{rmk:klmarginal} tells us that this difference can be alternatively understood by comparing the predictive distributions $q(y_{\mbf{x}}=1| \dc)$ in Table~\ref{tbl:moon4casemean} vs.~$q(y_{\mbf{x}}=1|\da)$ in Fig.~\ref{fig:moon}b.

Fig.~\ref{fig:moon4casemae} shows results of averaged KL divergences (i.e., performance metric described in Sec.~\ref{sec:experiment}) achieved by EUBO, rKL, and $q(\mbf{f}_{\mcl{X}_u}|\da)$ over $\dc$ and $\dr$ for the $4$ scenarios above. 
Table~\ref{tbl:moon4casemean} also analyzes the performance of our unlearning methods qualitatively by plotting the means of the approximate posterior beliefs $\eubo(f_{\mbf{x}}|\dc; \lambda)$ and  $\elbo(f_{\mbf{x}}|\dc; \lambda)$ induced, respectively, by EUBO and rKL with the corresponding predictive distributions
$\eubo(y_{\mbf{x}}=1|\dc)$ and $\elbo(y_{\mbf{x}}=1|\dc)$, together with the mean of the approximate posterior belief $q(f_{\mbf{x}}|\dc)$ with the corresponding predictive distribution
$q(y_{\mbf{x}}=1|\dc)$ obtained using VI from retraining with remaining data $\dc$.
The following observations result:
\begin{itemize}
    \item Fig.~\ref{fig:moon4casemae} shows that as $\mcl{I}$ increases across the $4$ scenarios, the averaged KL divergence between  $q(y_{\mbf{x}}|\da)$ vs.~$q(y_{\mbf{x}}|\dc)$ over $\dc$ and $\dr$ 
    (i.e., baseline labeled as \emph{full}) generally increases.
    
    \item In the scenario of randomly selected $\dr$ (i.e., $\mcl{I}$ is small), we expect the difference between the predictive distributions $q(y_{\mbf{x}}|\da)$ vs.~$q(y_{\mbf{x}}|\dc)$ over $\dc$ and $\dr$
    to be small, which is reflected in the very small averaged KL divergences of about $0.002$ and $0.004$  achieved by $q(\mbf{f}_{\mcl{X}_u}|\da)$ (i.e., baseline labeled as \emph{full}) in Figs.~\ref{fig:moon4casemae}b and~\ref{fig:moon4casemae}c, respectively. 
    It can also be observed that though  EUBO and rKL with $\lambda \in \{10^{-5}, 10^{-9}\}$ achieve smaller averaged KL divergences than that of $q(\mbf{f}_{\mcl{X}_u}|\da)$ (i.e., baseline), 
    EUBO's averaged KL divergence increases beyond than that of the baseline when $\lambda = 0$, but remains very small. 
    As a result, the first row in Table~\ref{tbl:moon4casemean} shows that when $\lambda = 10^{-9}$ or $\lambda = 0$, the predictive distributions $\eubo(y_{\mbf{x}}=1|\dc)$ and $\elbo(y_{\mbf{x}}=1|\dc)$ induced, respectively, by EUBO and rKL are similar to $q(y_{\mbf{x}}=1|\dc)$ obtained using VI from retraining with $\dc$. Hence, we can conclude that both EUBO and rKL perform reasonably well in this scenario, even when $\lambda = 0$.
    \item In the scenarios of partially and largely `yellow' $\dr$, $\mcl{I}$ is much larger than that in the scenario of randomly selected $\dr$. So, we expect an increase in the difference between the predictive distributions $q(y_{\mbf{x}}|\da)$ vs.~$q(y_{\mbf{x}}|\dc)$ over $\dc$ and $\dr$.
    It can be observed from Figs.~\ref{fig:moon4casemae}e-f and~\ref{fig:moon4casemae}h-i that when $\lambda = 0$, EUBO performs poorly as its averaged KL divergence is larger than that of $q(\mbf{f}_{\mcl{X}_u}|\da)$ (i.e., baseline labeled as \emph{full}), while rKL performs  well 
    as its averaged KL divergence is much smaller than that of the baseline. 
    On the other hand, when $\lambda = 10^{-9}$, both EUBO and rKL perform well,
    which can also be observed from the second and third rows of Table~\ref{tbl:moon4casemean}.
    These plots also show that while the predictive distributions $\elbo(y_{\mbf{x}}=1|\dc)$ induced by rKL with $\lambda=10^{-9}$ are not as similar to $q(y_{\mbf{x}}=1|\dc)$ as $\eubo(y_{\mbf{x}}=1|\dc)$ induced by EUBO with $\lambda=10^{-9}$, the  performance of rKL with $\lambda = 0$ is more robust.
    \item In the scenario of fully `yellow' $\dr$ (i.e., $\mcl{I}$ is largest), the difference between the predictive distributions $q(y_{\mbf{x}}|\da)$ vs.~$q(y_{\mbf{x}}|\dc)$ over $\dc$ and $\dr$ is larger than that in the above $3$ scenarios. 
    Except for EUBO with $\lambda=0$, 
    the predictive distributions $\eubo(y_{\mbf{x}}|\dc)$ and $\elbo(y_{\mbf{x}}|\dc)$ induced, respectively, by EUBO and rKL are closer to $q(y_{\mbf{x}}|\dc)$ than $q(y_{\mbf{x}}|\da)$
    as they achieve smaller averaged KL divergences than that of $q(\mbf{f}_{\mcl{X}_u}|\da)$, as shown in Figs.~\ref{fig:moon4casemae}k-l. However, the fourth row of Table~\ref{tbl:moon4casemean} shows that both EUBO and rKL do not perform that well. Nevertheless, it can be observed that when $\lambda = 0$, the predictive distribution $\elbo(y_{\mbf{x}}=1|\dc)$ induced by rKL  is still usable while $\eubo(y_{\mbf{x}}=1|\dc)$ induced by EUBO is useless.
%
\end{itemize}
%
%
To summarize, when only an approximate posterior belief $q(\bm{\theta}|\da)$ of model parameters $\bm{\theta}=\mbf{f}_{\mcl{X}_u}$ given full data $\da$ (i.e., obtained in model training with VI) is available, both EUBO and rKL can perform well if the difference between the approximate posterior belief of model parameters given remaining data $\dc$ vs.~that given full data $\da$ is sufficiently small. 
In practice, this is expected
due to the small quantity of erased data and redundancy in real-world datasets. In the case where the erased data is highly informative, the approximate posterior belief $\elbo(\bm{\theta}|\dc; \lambda=0)$ induced by rKL remains usable by being close to $q(\bm{\theta}|\da)$ and hence sacrificing its unlearning performance.
On the other hand, EUBO may suffer from poor unlearning performance when $\lambda$ is too small.

The above remark highlights the limitation of our unlearning methods when the erased data $\dr$ is informative and only the approximate posterior belief $q(\bm{\theta}|\da)$ is available.
Such a limitation is due to the lack of  information about the difference between the exact posterior belief $p(\bm{\theta}|\da)$ vs.~the approximate one $q(\bm{\theta}|\da)$ (Sec.~\ref{subsec:apprfull}), which 
motivates future investigation into maintaining additional information about this difference during the model training with VI to improve the unlearning performance.
In practice, an ML application may require an unlearning method to be time-efficient in order to satisfy the constraint on the response time to a user's request for her data to be erased while not rendering the model useless (e.g., due to catastrophic unlearning). 
After processing the user's request, the ML application can continue to improve the
approximate posterior belief recovered by unlearning from erased data (i.e., using our proposed EUBO or rKL) by retraining with the remaining data at the expense of parsimony (i.e., in terms of time and space costs).

One may wonder how our unlearning methods can handle multiple users' request arriving sequentially over time.
To avoid approximation errors from accumulating, we can adopt the approach of \emph{lazy} unlearning by aggregating all the (past and new) users' erased data into $\dr$ and performing unlearning (i.e., using only $q(\bm{\theta}|\da)$ and $\dr$) as and when necessary. As expected, our unlearning methods can perform well, provided that the aggregated erased data $\dr$ remains sufficiently small or contains enough redundancy.
%
%
\begin{table}
\caption{Plots of the mean 
of 
approximate posterior belief $q(f_{\mbf{x}}|\dc)$ with the corresponding predictive distribution 
$q(y_{\mbf{x}}=1|\dc)$ obtained using VI from retraining with remaining data $\dc$, and also the means of approximate posterior beliefs
$\eubo(f_{\mbf{x}}|\dc;\lambda)$ and $\elbo(f_{\mbf{x}}|\dc;\lambda)$ induced, respectively, by EUBO and rKL with the corresponding
predictive distributions
$\eubo(y_{\mbf{x}}=1|\dc)$ and $\elbo(y_{\mbf{x}}=1|\dc)$ for $\lambda \in [10^{-9}, 0]$.
The $1$-st, $2$-nd, $3$-rd, and $4$-th rows correspond to the following $4$ respective scenarios: randomly selected $\dr$, partially `yellow' $\dr$, largely `yellow' $\dr$, and fully `yellow' $\dr$.}
\begin{tabular}{@{}c@{}c@{}c@{}c@{}c@{}c@{}c@{}}
\toprule
\multirow[t]{2}{*}{Dataset}
&
\multicolumn{2}{c}{Retrained}
&
\multicolumn{2}{c}{EUBO}
&
\multicolumn{2}{c}{rKL}
\\
\cmidrule(l){2-3} \cmidrule(l){4-5} \cmidrule(l){6-7}
&
Mean $\mu_{\mbf{x}|\dc}$
&
$q(y_{\mbf{x}}=1|\dc)$
&
Mean
&
$\eubo(y_{\mbf{x}}=1|\dc)$
&
Mean
&
$\elbo(y_{\mbf{x}}=1|\dc)$
\\
\midrule
\\
\multirow[t]{4}{*}{
    \includegraphics[trim={0mm 0mm 3mm 3mm}, clip,height=0.13\textwidth]{img/moon/moon_random_data.pdf}
}
&
\multirow[t]{4}{*}{
    \includegraphics[trim={7mm 8mm 3mm 3mm}, clip,height=0.13\textwidth]{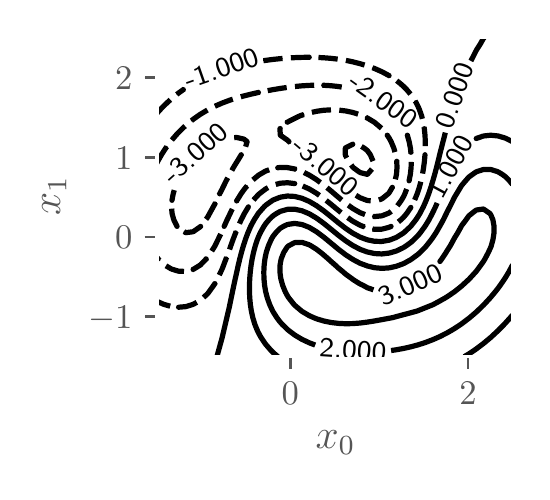}
}
&
\multirow[t]{4}{*}{
    \includegraphics[height=0.13\textwidth]{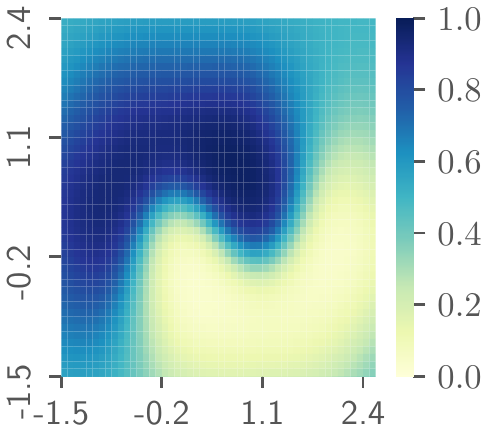}
}
&
\includegraphics[trim={7mm 8mm 3mm 3mm}, clip,height=0.13\textwidth]{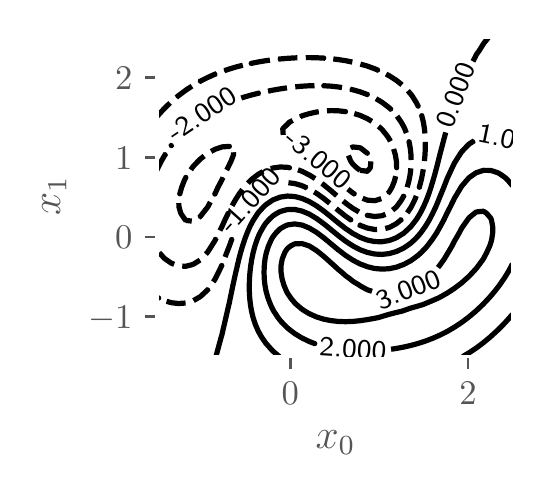}
&
\includegraphics[height=0.13\textwidth]{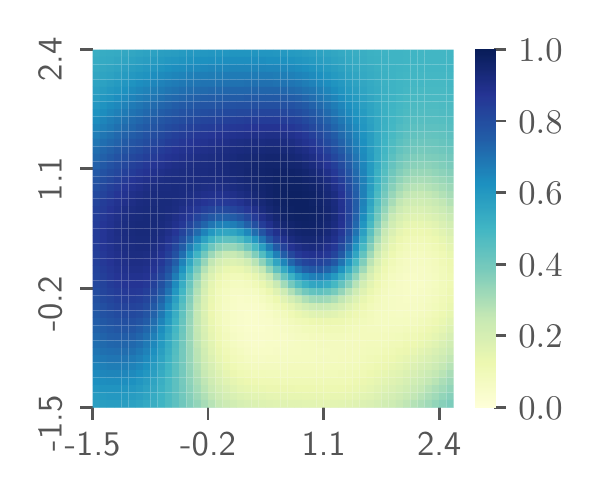}
&
\includegraphics[trim={7mm 8mm 3mm 3mm}, clip,height=0.13\textwidth]{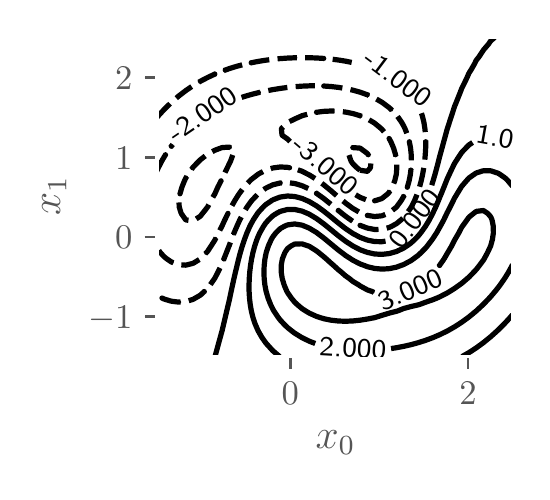}
&
\includegraphics[height=0.13\textwidth]{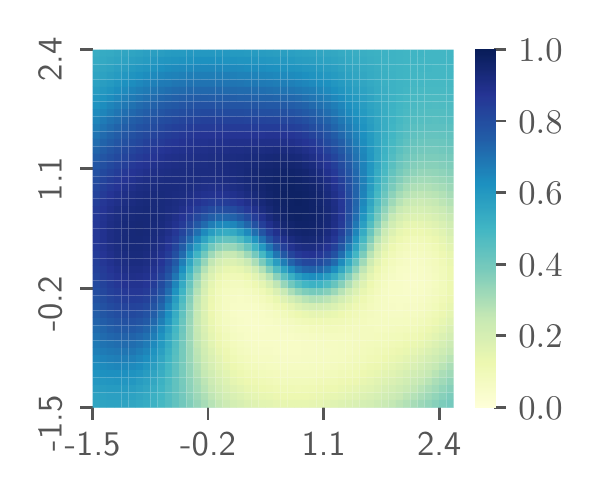}
\\
& & &
\multicolumn{4}{c}{$\lambda = 10^{-9}$}
\\
& & &
\includegraphics[trim={7mm 8mm 3mm 3mm}, clip,height=0.13\textwidth]{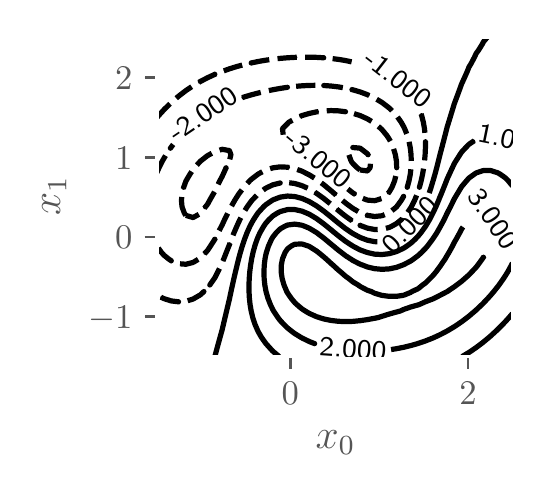}
&
\includegraphics[height=0.13\textwidth]{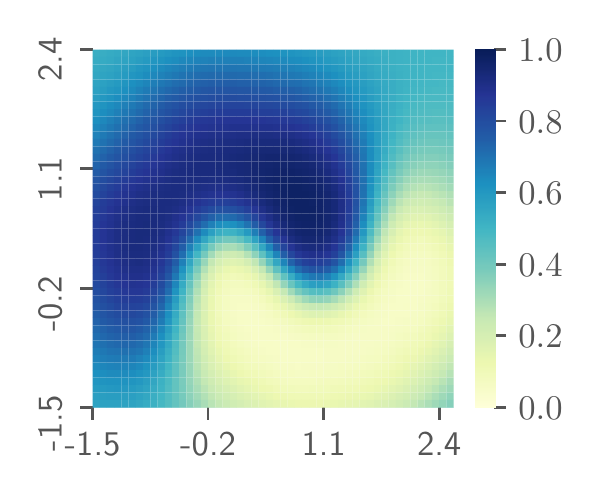}
&
\includegraphics[trim={7mm 8mm 3mm 3mm}, clip,height=0.13\textwidth]{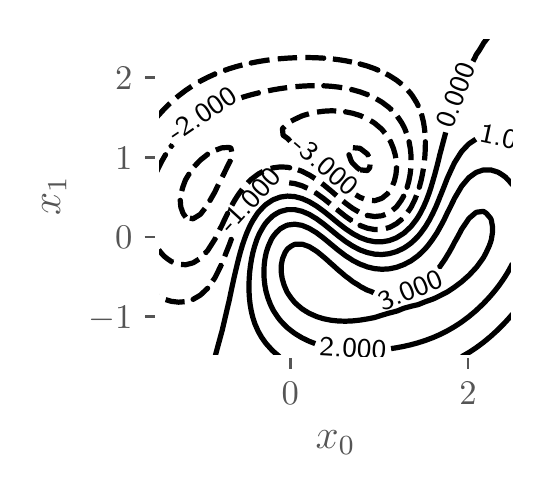}
&
\includegraphics[height=0.13\textwidth]{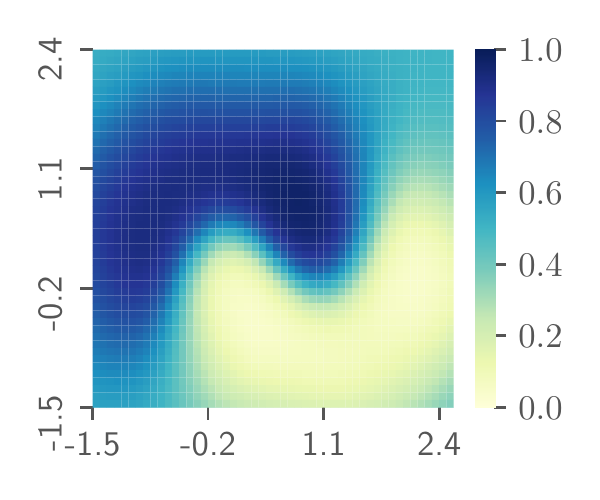}
\\
& & & 
\multicolumn{4}{c}{$\lambda = 0$}\\
\midrule
\multirow[t]{4}{*}{
    \includegraphics[trim={0mm 0mm 3mm 3mm}, clip,height=0.13\textwidth]{img/moon/moon_rm_30_data.pdf}
}
&
\multirow[t]{4}{*}{
    \includegraphics[trim={7mm 8mm 3mm 3mm}, clip,height=0.13\textwidth]{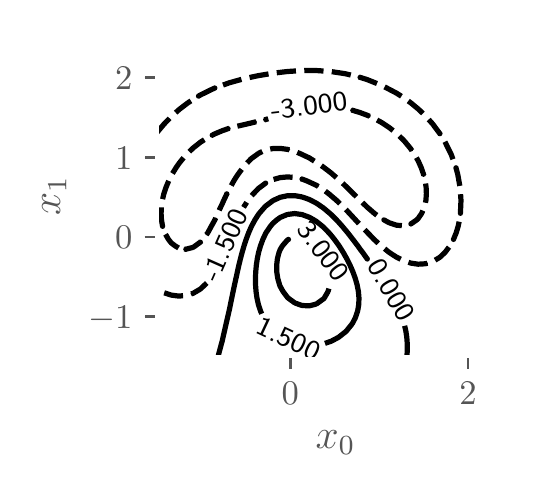}
}
&
\multirow[t]{4}{*}{
    \includegraphics[height=0.13\textwidth]{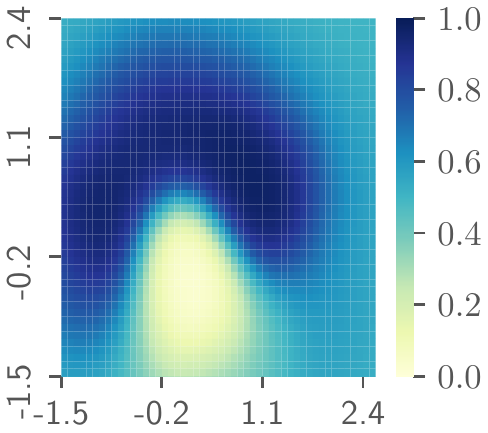}
}
&
\includegraphics[trim={7mm 8mm 3mm 3mm}, clip,height=0.13\textwidth]{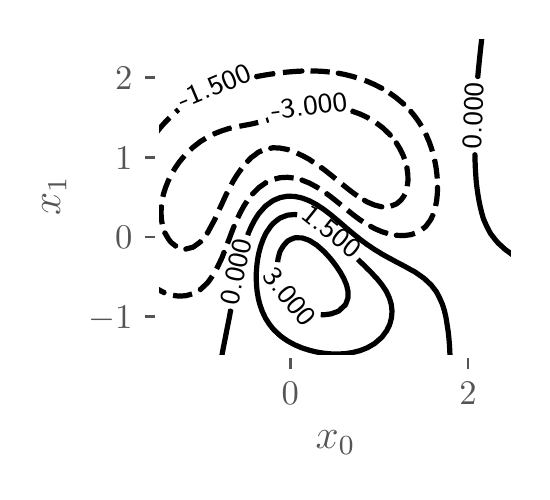}
&
\includegraphics[height=0.13\textwidth]{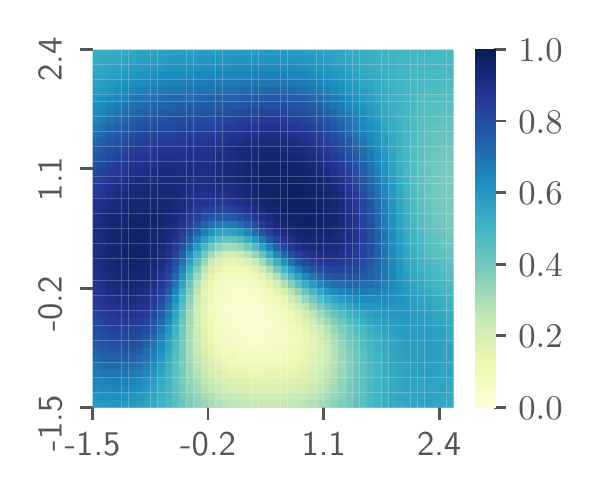}
&
\includegraphics[trim={7mm 8mm 3mm 3mm}, clip,height=0.13\textwidth]{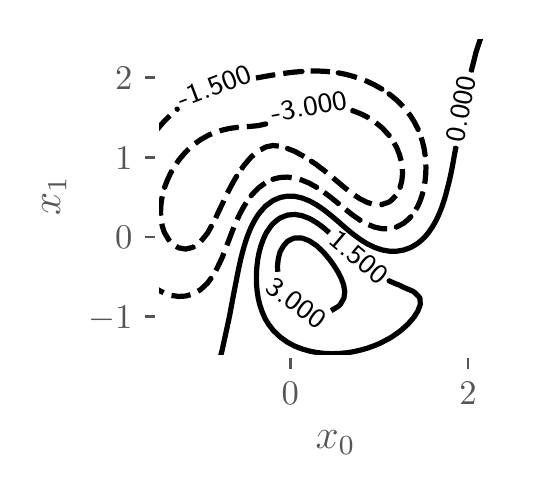}
&
\includegraphics[height=0.13\textwidth]{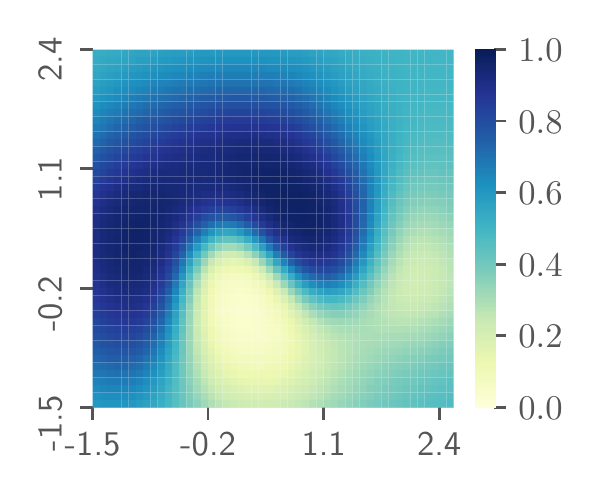}
\\
& & &
\multicolumn{4}{c}{$\lambda = 10^{-9}$}
\\
& & &
\includegraphics[trim={7mm 8mm 3mm 3mm}, clip,height=0.13\textwidth]{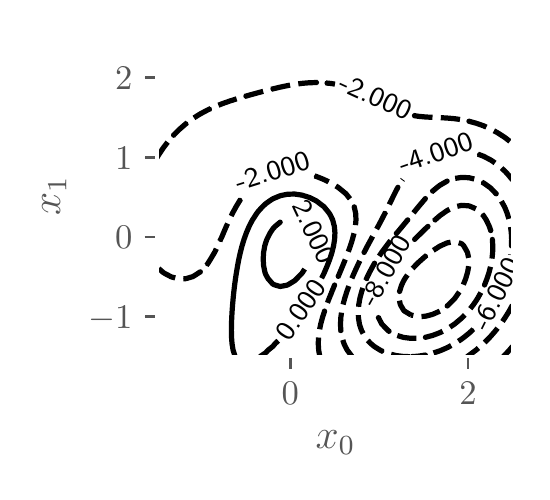}
&
\includegraphics[height=0.13\textwidth]{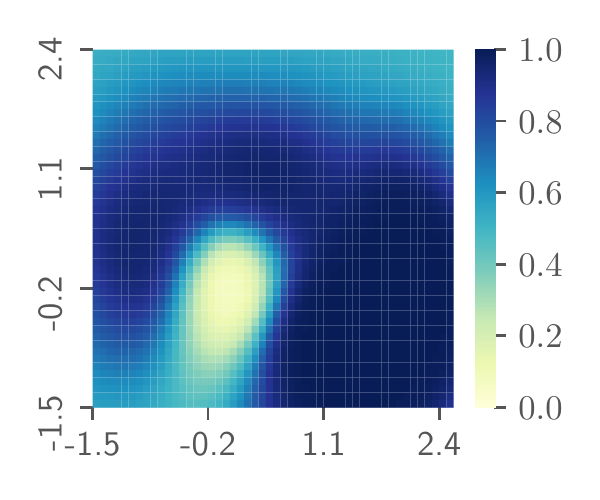}
&
\includegraphics[trim={7mm 8mm 3mm 3mm}, clip,height=0.13\textwidth]{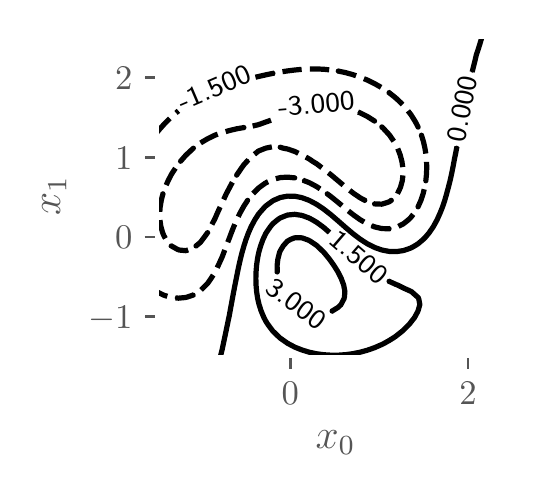}
&
\includegraphics[height=0.13\textwidth]{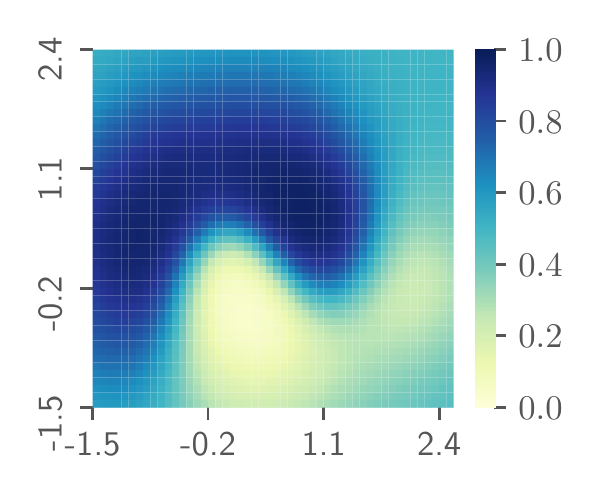}
\\
& & & 
\multicolumn{4}{c}{$\lambda = 0$}\\
\midrule
\multirow[t]{4}{*}{
    \includegraphics[trim={0mm 0mm 3mm 3mm}, clip,height=0.13\textwidth]{img/moon/moon_rm_40_data.pdf}
}
&
\multirow[t]{4}{*}{
    \includegraphics[trim={7mm 8mm 3mm 3mm}, clip,height=0.13\textwidth]{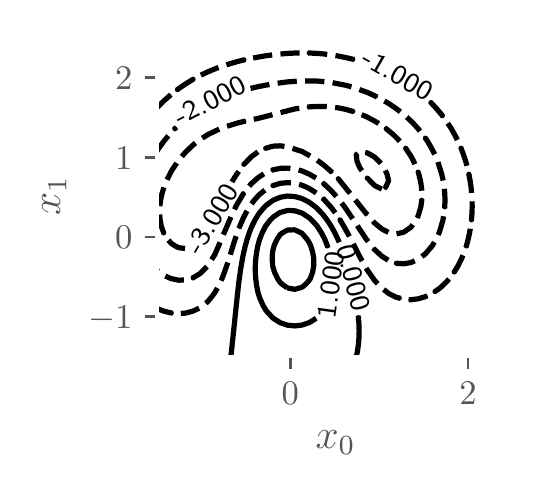}
}
&
\multirow[t]{4}{*}{
    \includegraphics[height=0.13\textwidth]{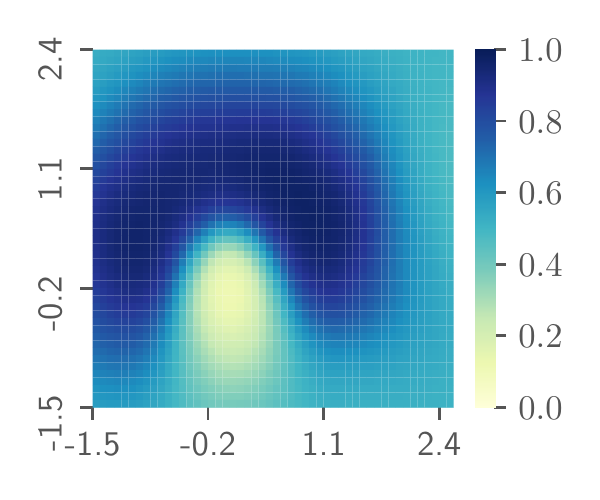}
}
&
\includegraphics[trim={7mm 8mm 3mm 3mm}, clip,height=0.13\textwidth]{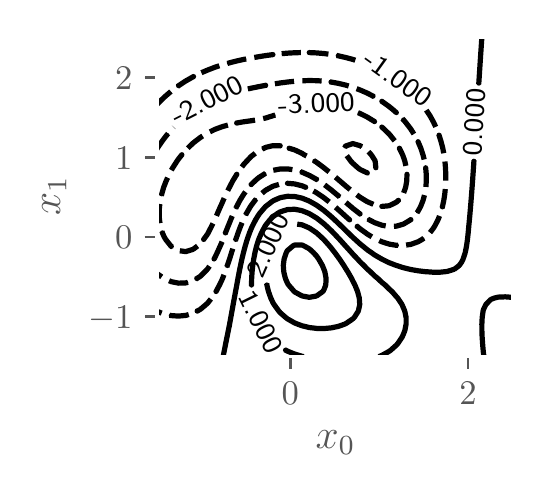}
&
\includegraphics[height=0.13\textwidth]{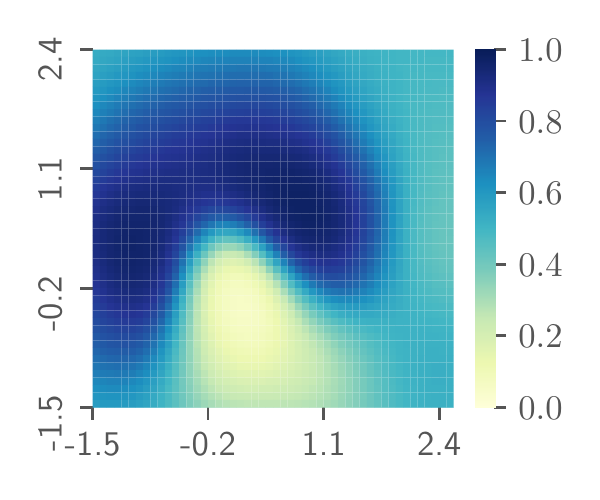}
&
\includegraphics[trim={7mm 8mm 3mm 3mm}, clip,height=0.13\textwidth]{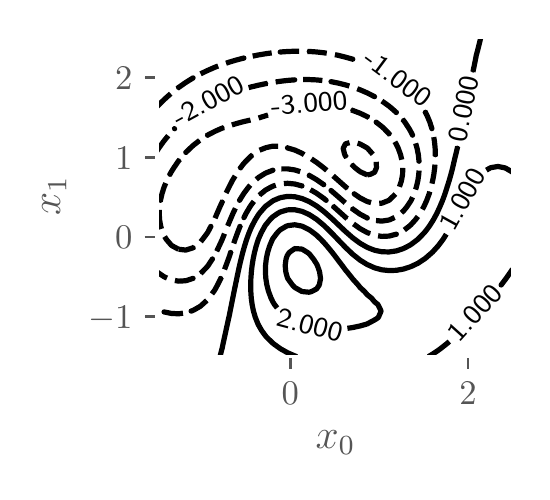}
&
\includegraphics[height=0.13\textwidth]{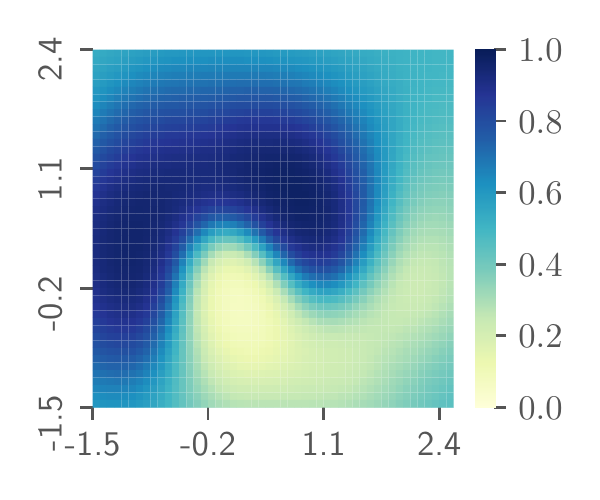}
\\
& & &
\multicolumn{4}{c}{$\lambda = 10^{-9}$}
\\
& & &
\includegraphics[trim={7mm 8mm 3mm 3mm}, clip,height=0.13\textwidth]{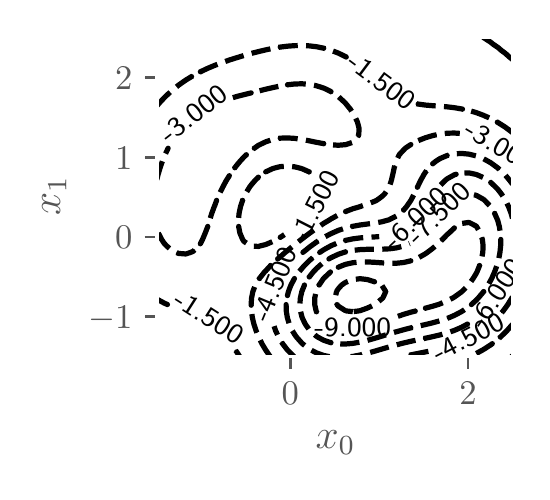}
&
\includegraphics[height=0.13\textwidth]{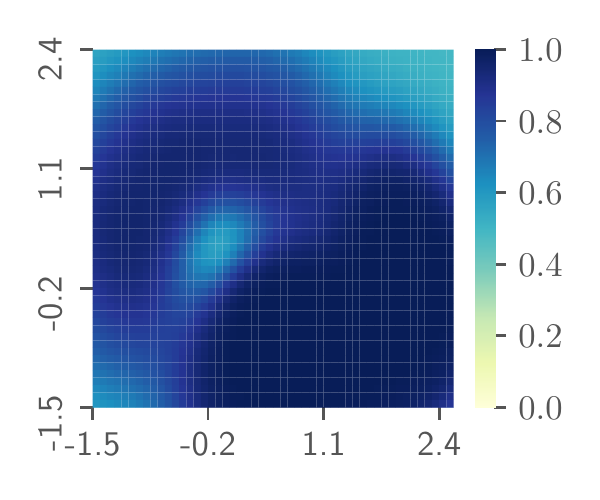}
&
\includegraphics[trim={7mm 8mm 3mm 3mm}, clip,height=0.13\textwidth]{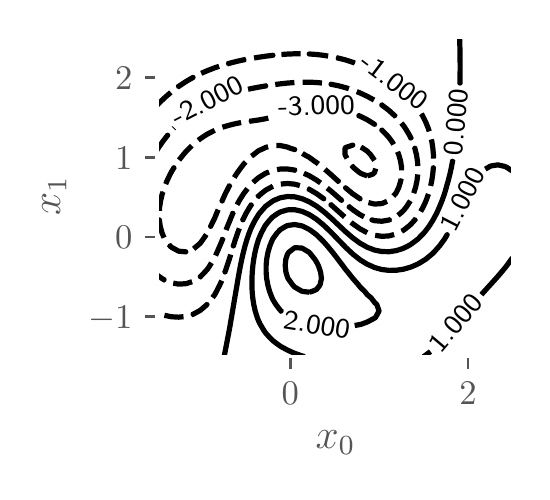}
&
\includegraphics[height=0.13\textwidth]{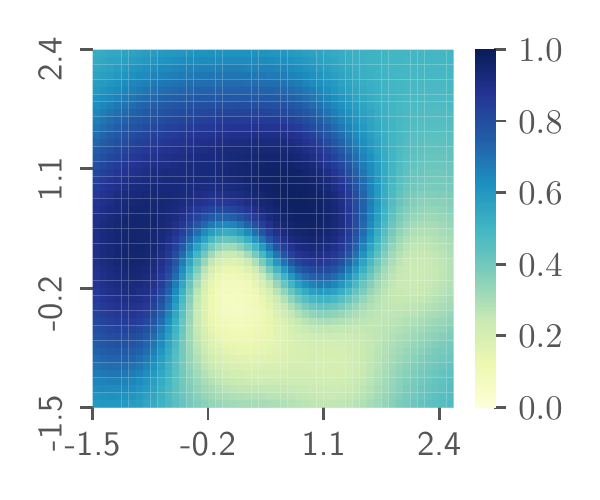}
\\
& & & 
\multicolumn{4}{c}{$\lambda = 0$}\\
\midrule
\multirow[t]{4}{*}{
    \includegraphics[trim={0mm 0mm 3mm 3mm}, clip,height=0.13\textwidth]{img/moon/moon_rm_50_data.pdf}
}
&
\multirow[t]{4}{*}{
    \includegraphics[trim={7mm 8mm 3mm 3mm}, clip,height=0.13\textwidth]{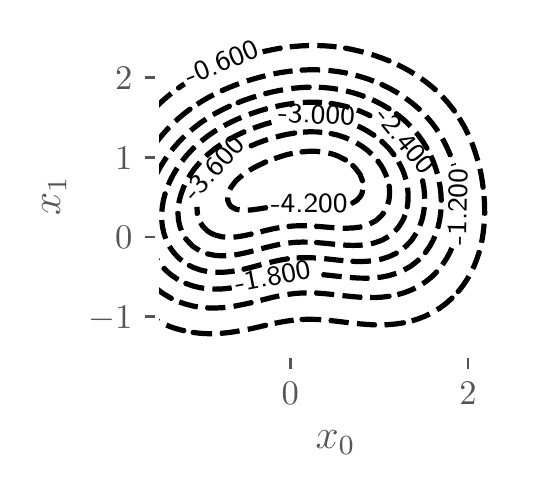}
}
&
\multirow[t]{4}{*}{
    \includegraphics[height=0.13\textwidth]{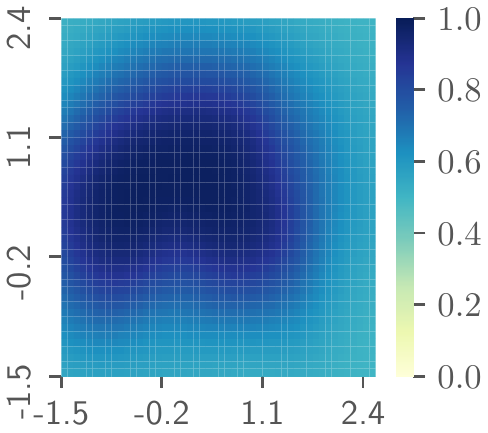}
}
&
\includegraphics[trim={7mm 8mm 3mm 3mm}, clip,height=0.13\textwidth]{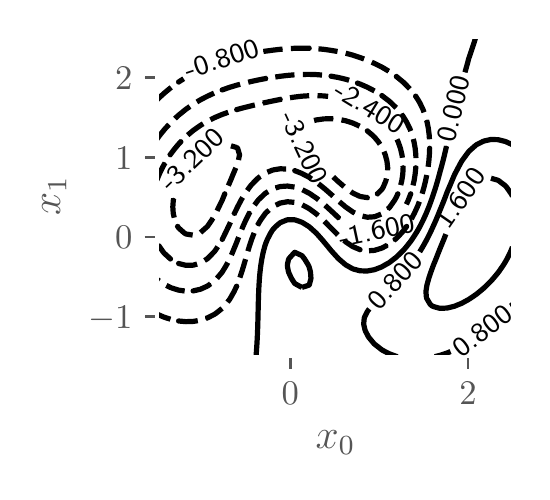}
&
\includegraphics[height=0.13\textwidth]{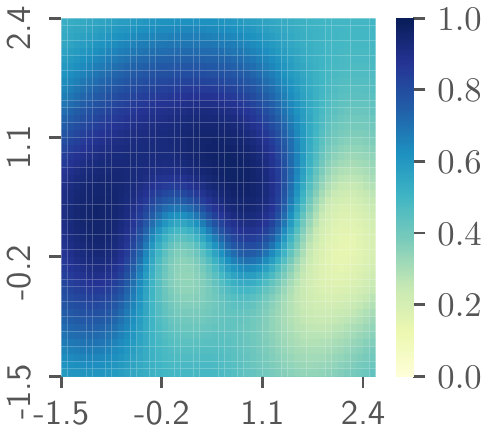}
&
\includegraphics[trim={7mm 8mm 3mm 3mm}, clip,height=0.13\textwidth]{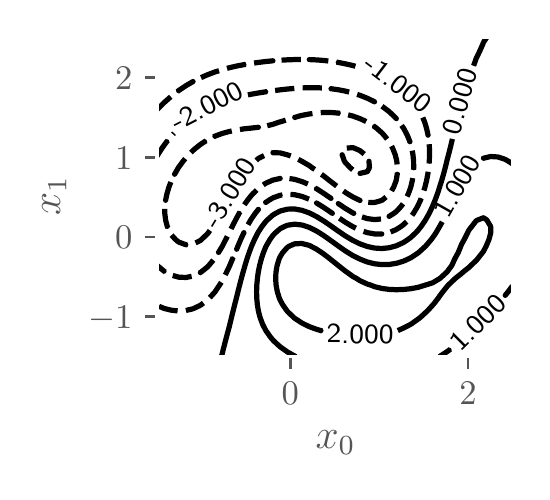}
&
\includegraphics[height=0.13\textwidth]{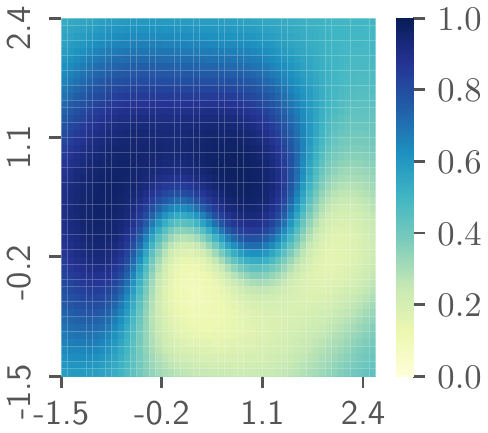}
\\
& & &
\multicolumn{4}{c}{$\lambda = 10^{-9}$}
\\
& & &
\includegraphics[trim={7mm 8mm 3mm 3mm}, clip,height=0.13\textwidth]{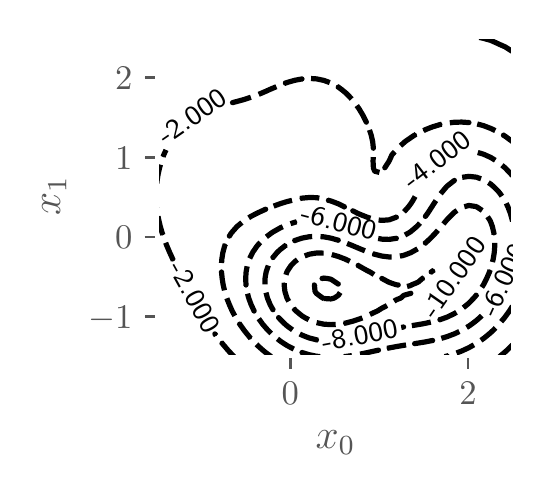}
&
\includegraphics[height=0.13\textwidth]{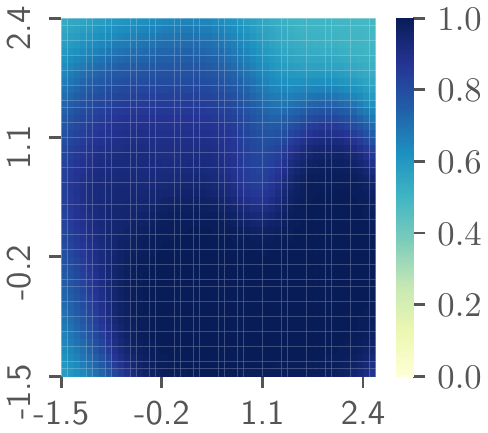}
&
\includegraphics[trim={7mm 8mm 3mm 3mm}, clip,height=0.13\textwidth]{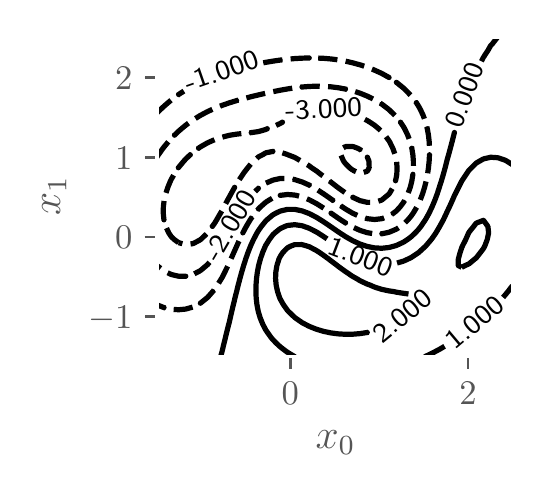}
&
\includegraphics[height=0.13\textwidth]{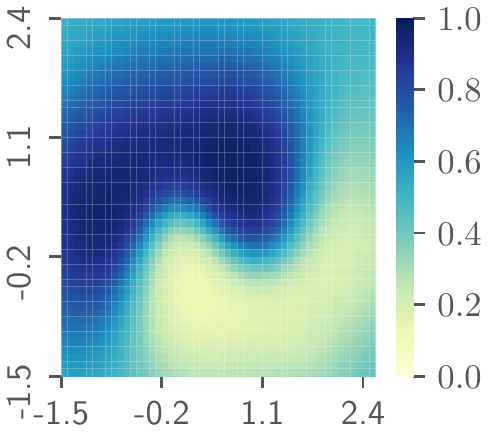}
\\
& & & 
\multicolumn{4}{c}{$\lambda = 0$}\\
\bottomrule
\end{tabular}
\label{tbl:moon4casemean}
\end{table}

\section{Logistic Regression with Fashion MNIST Dataset: Additional Experimental Results
}
\label{app:fashionmnist}
\begin{figure}
\centering
\begin{tabular}{@{}c@{}cc@{}c@{}}
(a)
&
\includegraphics[height=0.4\textwidth]{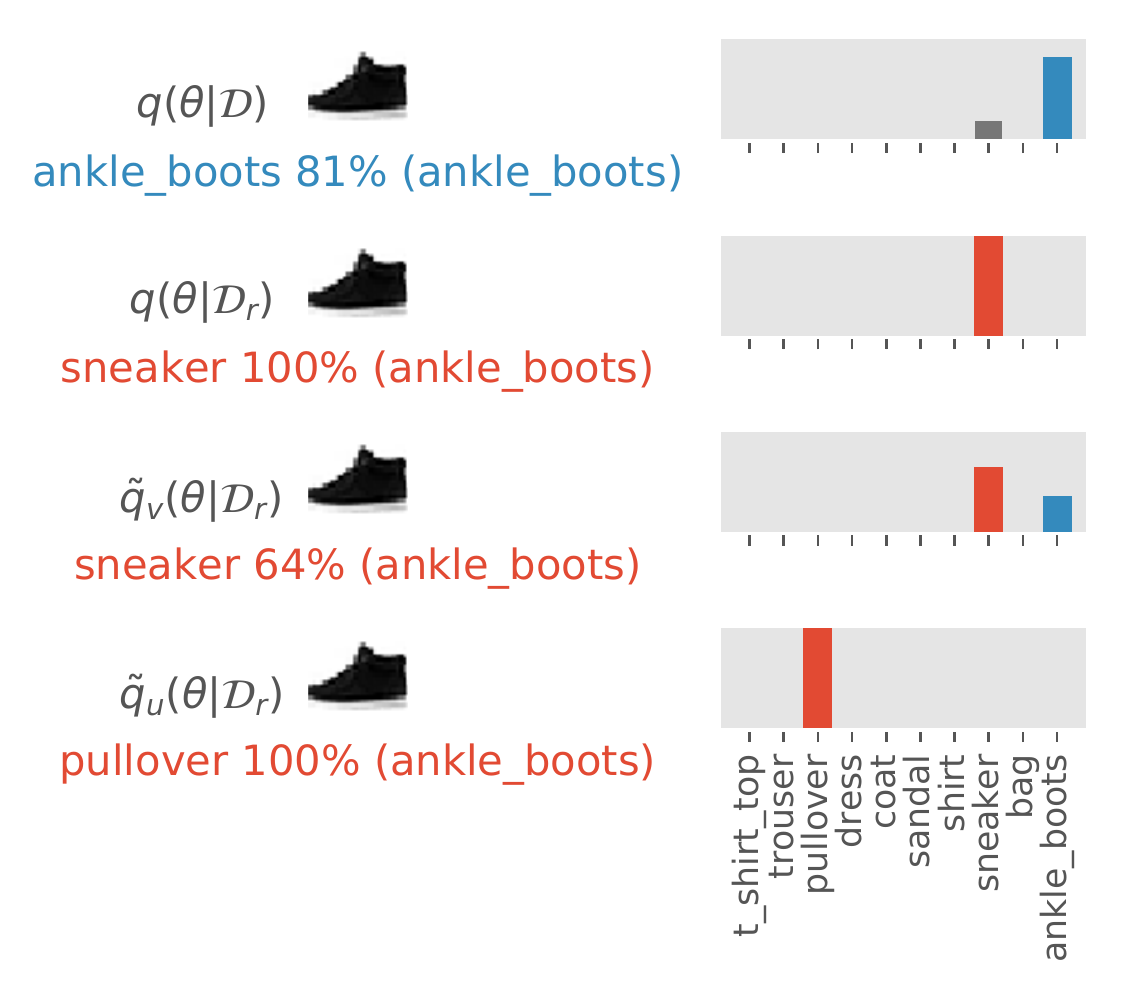}
&
(b)
&
\includegraphics[height=0.4\textwidth]{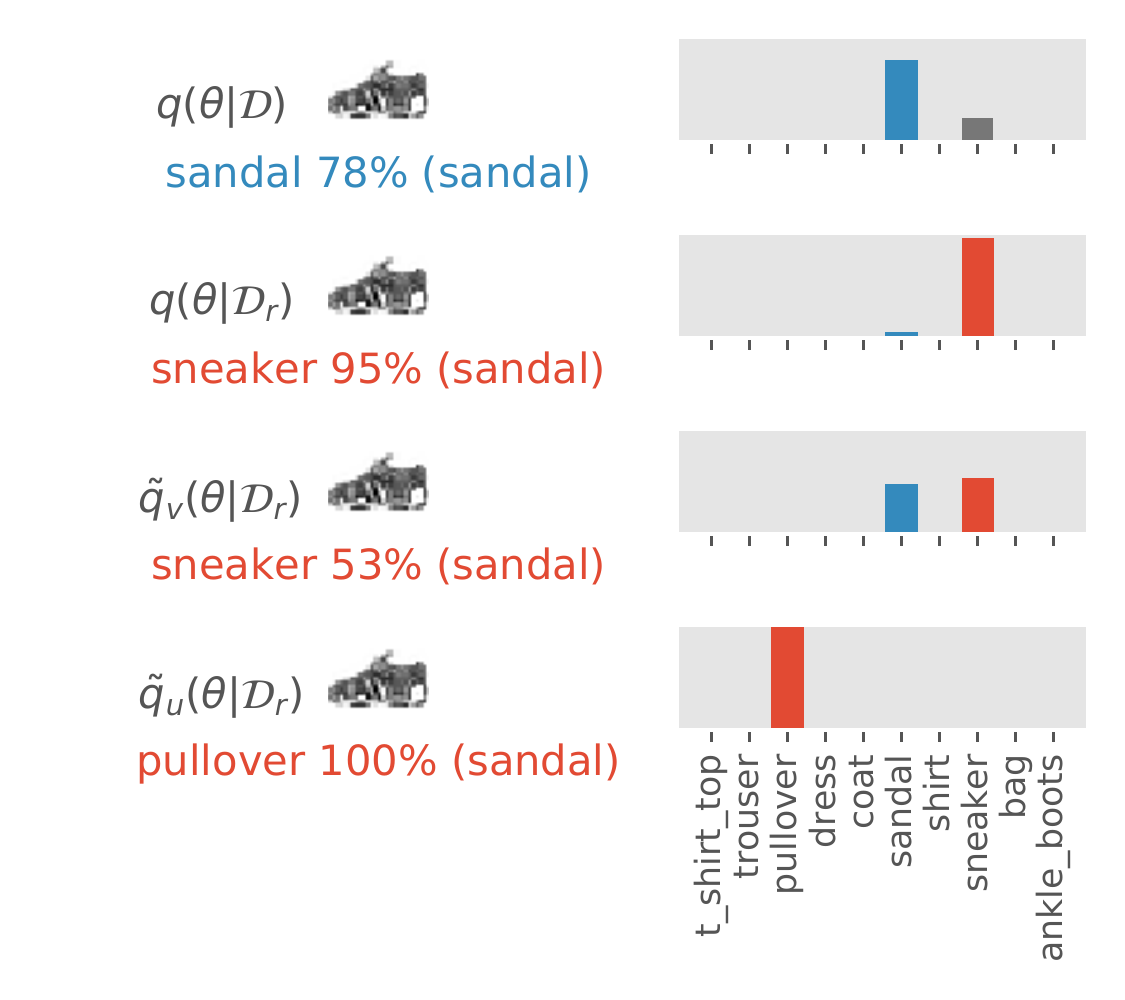}
\\
(c)
&
\includegraphics[height=0.4\textwidth]{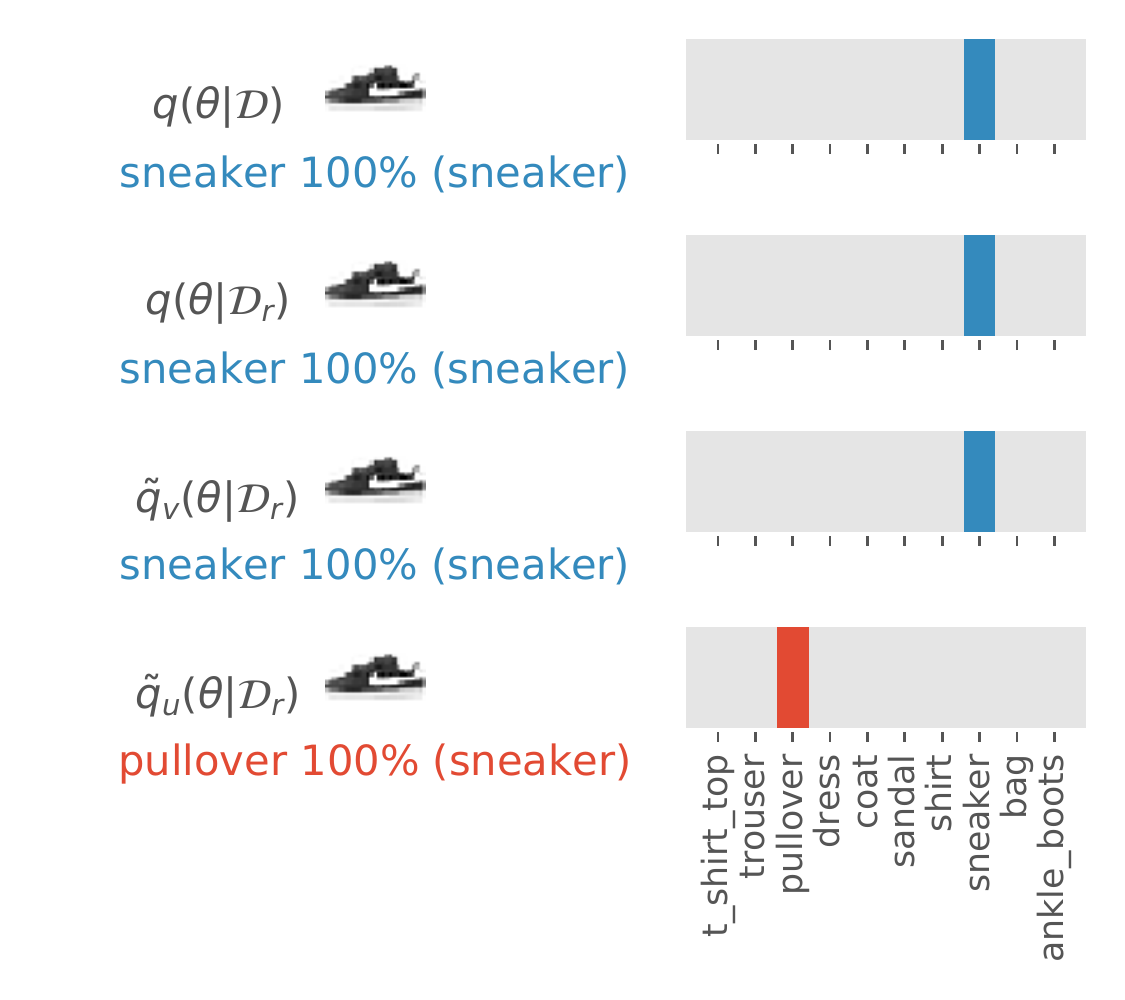}
&
(d)
&
\includegraphics[height=0.4\textwidth]{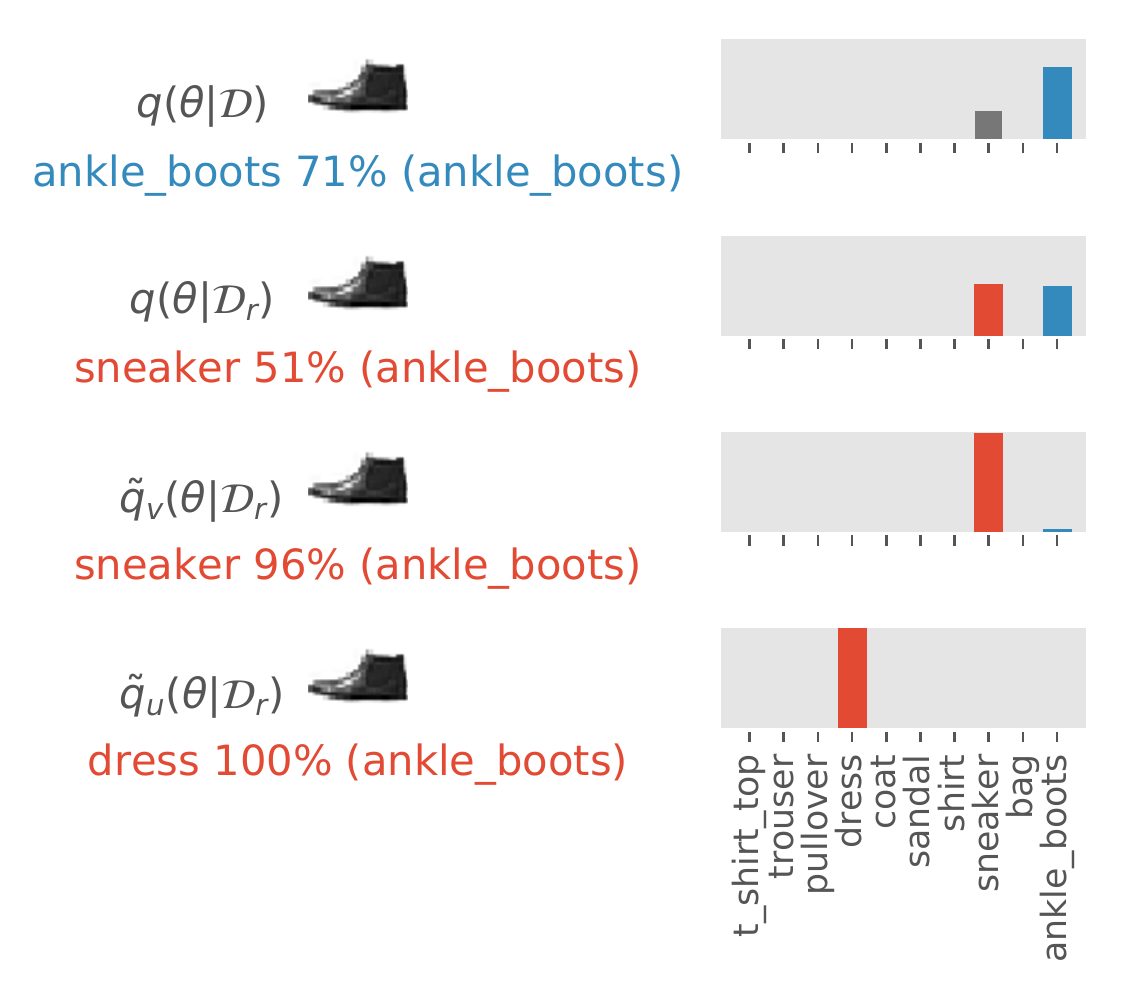}
\\
(e)
&
\includegraphics[height=0.4\textwidth]{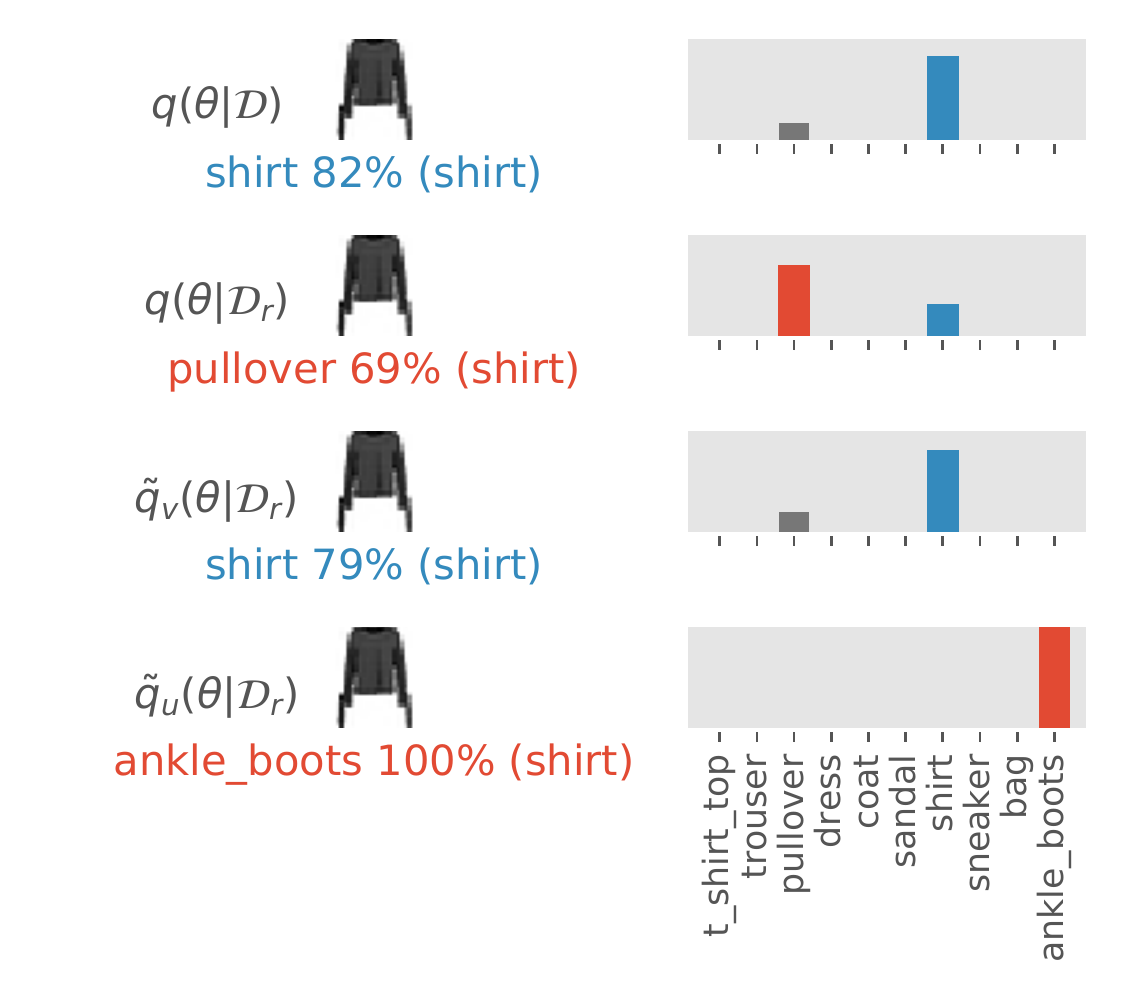}
&
(f)
&
\includegraphics[height=0.4\textwidth]{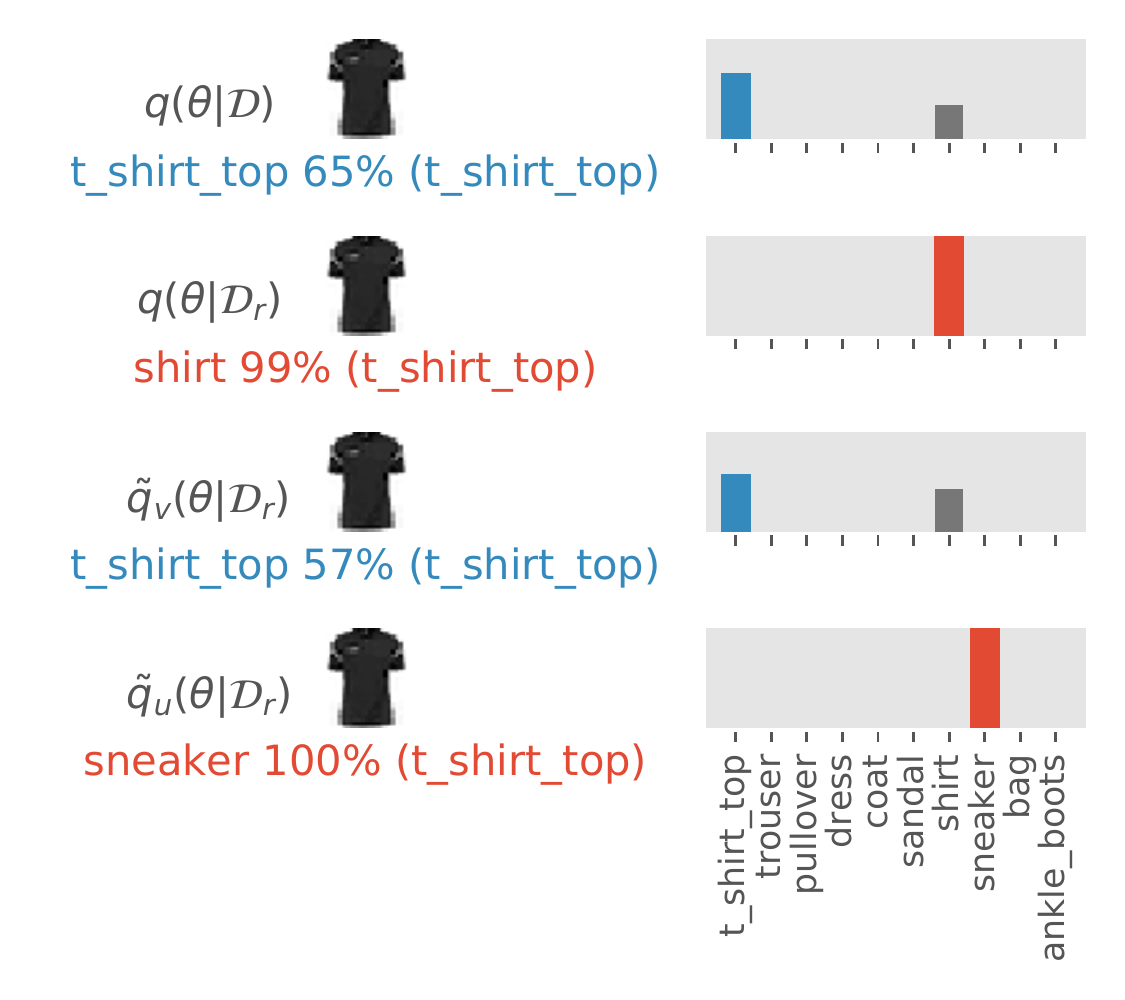}
\\
(g)
&
\includegraphics[height=0.4\textwidth]{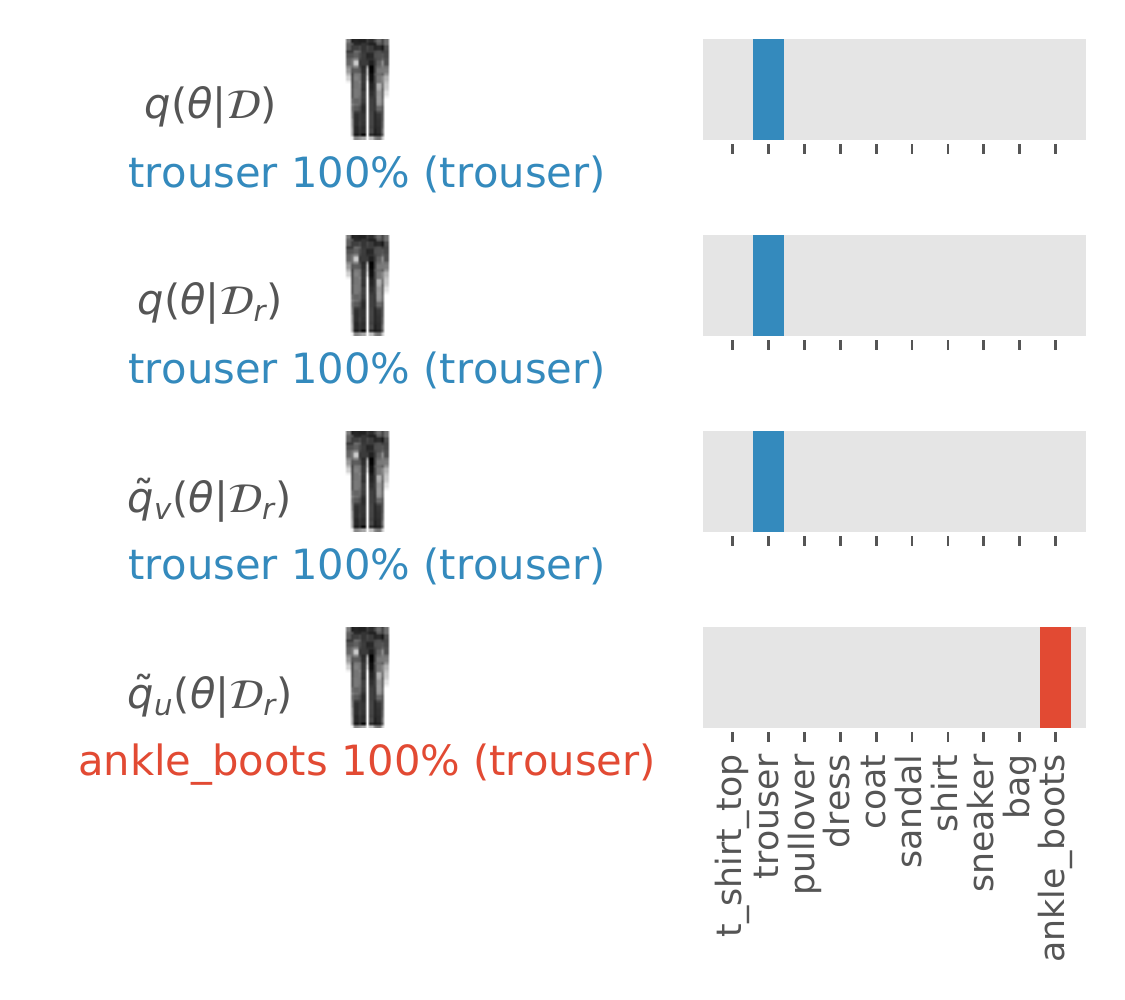} 
&
(h)
&
\includegraphics[height=0.4\textwidth]{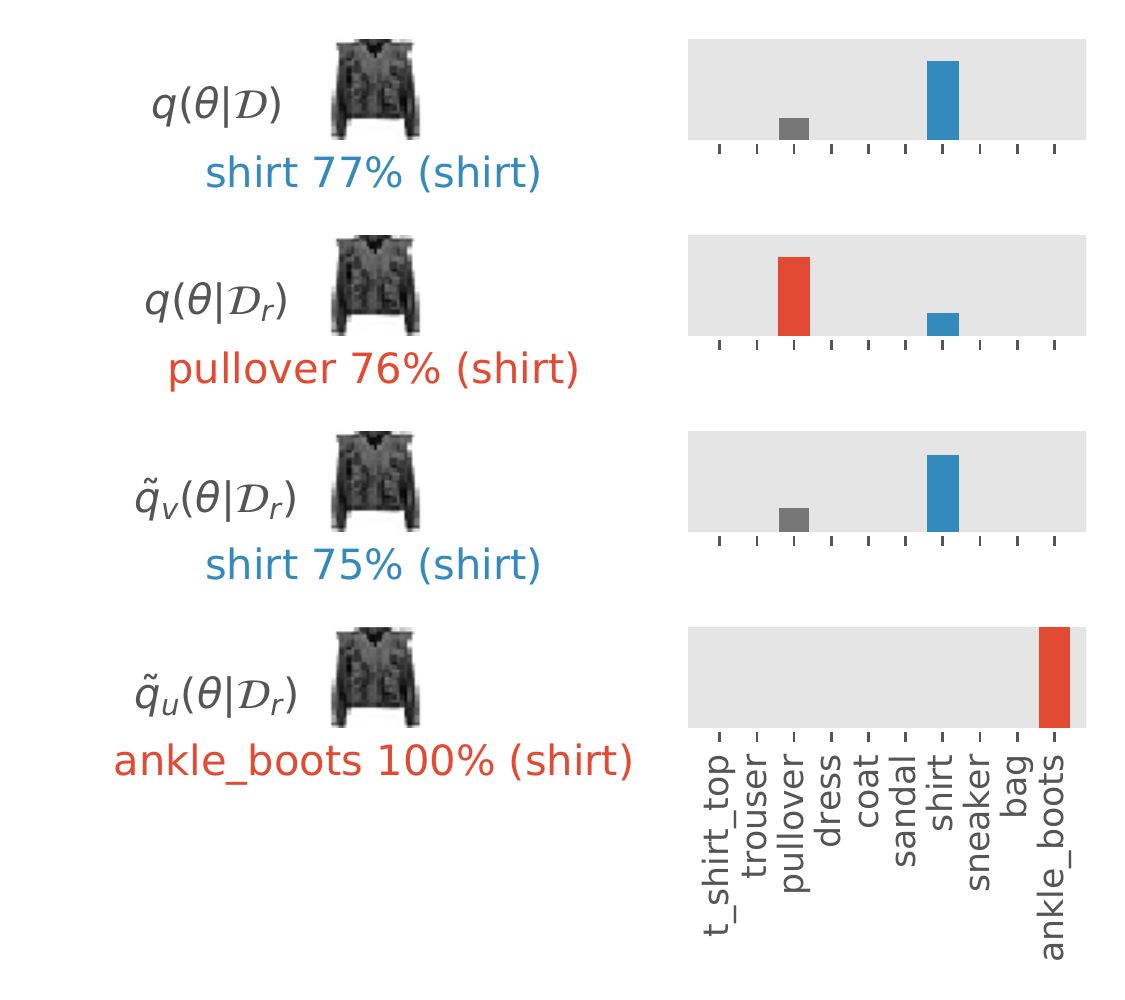}
\end{tabular}
\caption{Plots of class probabilities for images in $\dc$ obtained using $q(\bm{\theta}|\da)$, $q(\bm{\theta}|\dc)$, optimized $\elbo(\bm{\theta}|\dc;\lambda=0)$ and $\eubo(\bm{\theta}|\dc;\lambda=0)$.}
\label{fig:mnistmore}
\end{figure}

In this section, we will present the following:
\begin{itemize}
    \item Additional visualizations of the class probabilities for images in $\dc$ evaluated at the mean of the approximate posterior beliefs obtained using EUBO and rKL with $\lambda = 0$ in Fig.~\ref{fig:mnistmore}, and
    \item Comparison of the unlearning performance obtained using approximate posterior beliefs modeled with independent Gaussians (i.e., diagonal covariance matrices) vs.~that modeled with multivariate Gaussians (i.e., full covariance matrices).
\end{itemize}
Fig.~\ref{fig:mnistmore} shows the class probabilities for the images in $\dc$ evaluated at the mean of the approximate posterior beliefs with $\lambda = 0$. 
Figs.~\ref{fig:mnistmore}a-d and~\ref{fig:mnistmore}g show that rKL induces the highest class probability for the same class as that of
$q(\bm{\theta}|\dc)$.
In Figs.~\ref{fig:mnistmore}e-f and~\ref{fig:mnistmore}h, the class probabilities obtained using optimized $\elbo(\bm{\theta}|\dc;\lambda=0)$ resemble that obtained using $q(\bm{\theta}|\da)$, though the probability of the correct class is reduced due to unlearning.

\begin{figure}
\centering
\begin{tabular}[t]{@{}c@{}c@{}c@{}c@{}c@{}}
\includegraphics[height=0.4\textwidth]{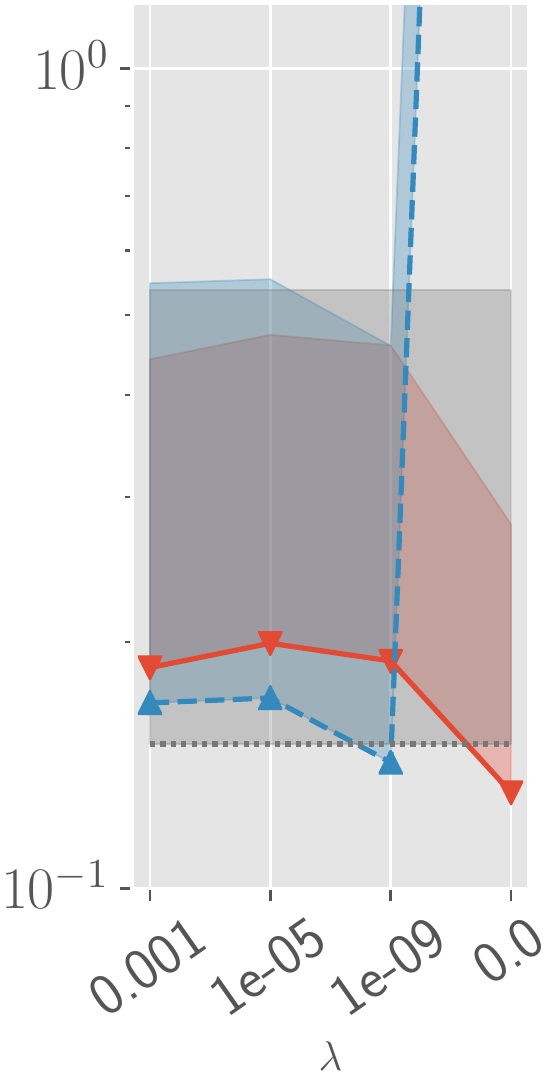}
&
\includegraphics[ height=0.4\textwidth]{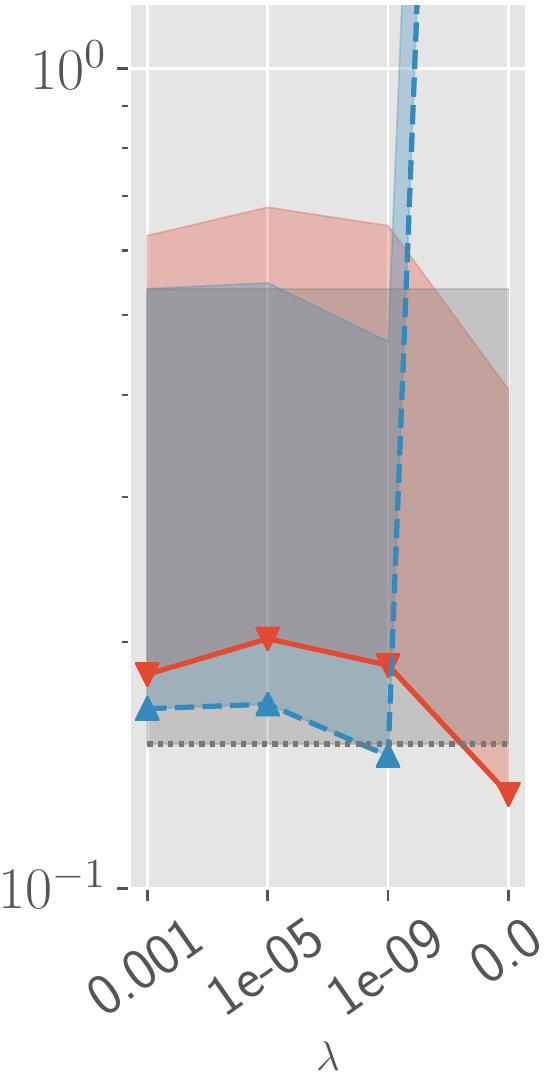}
&
\includegraphics[ height=0.4\textwidth]{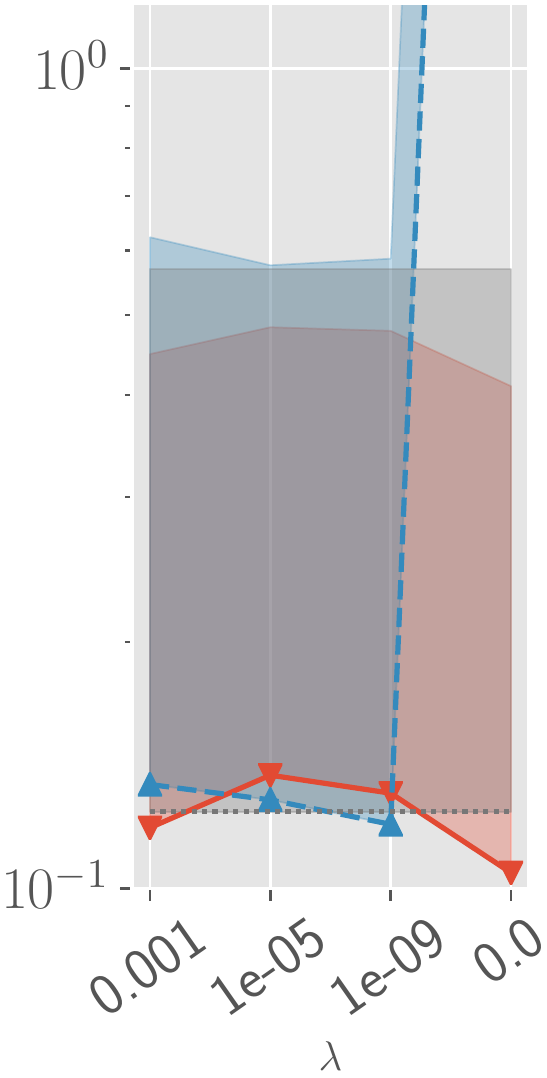}
&
\includegraphics[ height=0.4\textwidth]{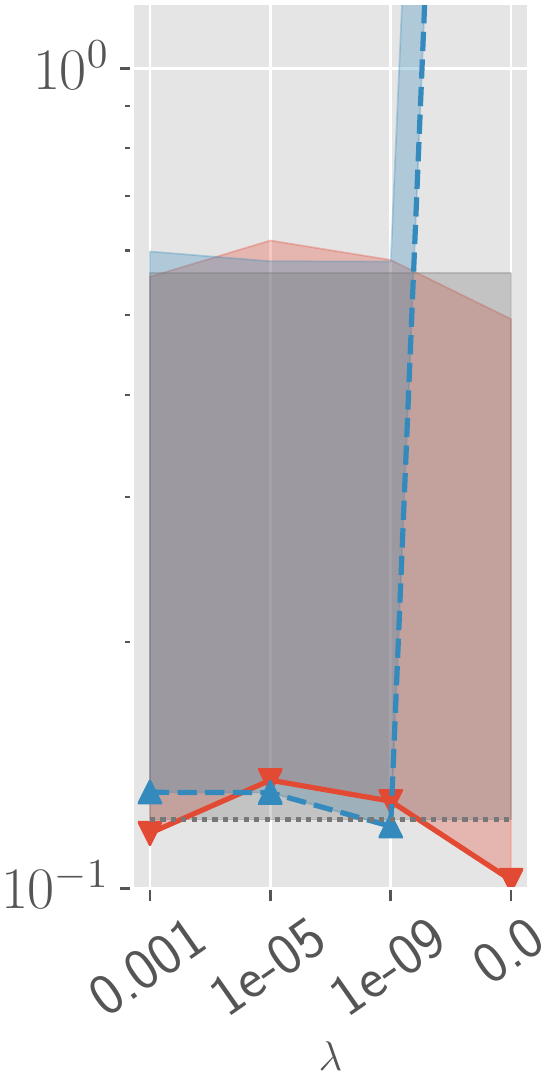}
&
\includegraphics[trim={0mm 20mm 0 0}, clip, height=0.35\textwidth]{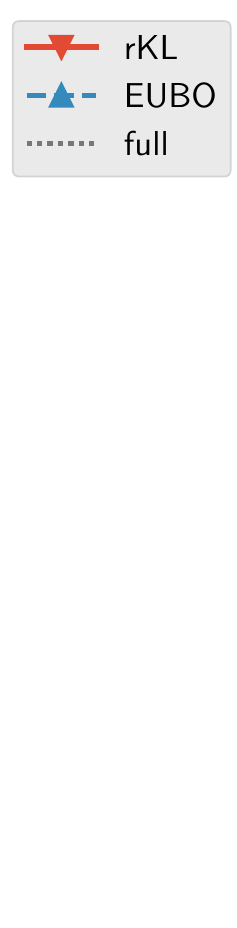}\\
(a) $\dc$
&
(b) $\dr$
&
(c) $\dc$
&
(d) $\dr$
&
\end{tabular}
\caption{Graphs of averaged KL divergence vs.~$\lambda$ achieved by EUBO, rKL, and $q(\bm{\theta}|\da)$ (i.e., baseline labeled as \emph{full}) over $\dc$ and $\dr$ for the fashion MNIST dataset. The approximate posterior beliefs of the model parameters/weights are represented by (a-b) independent Gaussians (i.e., diagonal covariance matrices) and (c-d) multivariate Gaussians (i.e., full covariance matrices).}
\label{fig:mnistdiagfull}
\end{figure}

Fig.~\ref{fig:mnistdiagfull} shows the averaged KL divergences of EUBO, rKL, and $q(\bm{\theta}|\da)$ where the approximate posterior beliefs are modeled with independent Gaussians (i.e., diagonal covariance matrices) in Figs.~\ref{fig:mnistdiagfull}a-b and multivariate Gaussians (i.e., full covariance matrices) in Figs.~\ref{fig:mnistdiagfull}c-d. It can be observed that the averaged KL divergences between  $q(y_{\mbf{x}}|\da)$ vs.~$q(y_{\mbf{x}}|\dc)$ over $\dc$ and $\dr$ 
    (i.e., baselines labeled as \emph{full})
decrease when multivariate Gaussians with full covariance matrices are used to model the approximate posterior beliefs instead (compare the baselines labeled as \emph{full} in Figs.~\ref{fig:mnistdiagfull}c-d vs.~that in Figs.~\ref{fig:mnistdiagfull}a-b).
Furthermore, in such a case, the unlearning performance of both EUBO and rKL improve as their averaged KL divergences are not as large (relative to the baselines) as that using independent Gaussians.

\end{document}